\PassOptionsToPackage{table}{xcolor}
\documentclass[twoside]{article}

\usepackage[accepted]{aistats2025}

\usepackage[round]{natbib}

\usepackage[utf8]{inputenc} 
\usepackage[T1]{fontenc}    
\usepackage{hyperref}       
\usepackage{url}            
\usepackage{booktabs}       
\usepackage{amsfonts}       
\usepackage{nicefrac}       
\usepackage{microtype}      
\usepackage{xcolor}         
\usepackage{graphicx} 
\usepackage{amsmath}
\usepackage{amssymb}
\usepackage[outercaption]{sidecap} 
\usepackage{mathtools}
\usepackage{multirow}
\usepackage{amssymb}

\usepackage{float}

\newcommand{\thismodel}{\textsc{ChronosX}\xspace}
\newcommand{\momentx}{\textsc{MOMENTX}\xspace}
\newcommand{\timesfmx}{\textsc{TimesFMX}\xspace}
\newcommand{\thatmodel}{\textsc{Chronos}\xspace}
\newcommand{\moment}{\textsc{MOMENT}\xspace}
\newcommand{\timesfm}{\textsc{TimesFM}\xspace}
\newcommand{\rulesep}{\unskip\ \vrule\ }
\usepackage{inconsolata}
\usepackage{graphicx}
\usepackage{mathrsfs}
\usepackage{multirow}
\usepackage{booktabs}
\usepackage{enumitem}
\usepackage{hyperref}
\usepackage{soul}
\usepackage{subcaption}
\usepackage[font=small]{caption}
\usepackage[bottom]{footmisc}

\newcommand{\bftable}{\fontseries{b}\selectfont}

\usepackage{algorithm}
\usepackage{algpseudocode}

\usepackage{xkcdcolors}
\hypersetup{
    colorlinks=true,
    linkcolor=xkcdOcean,
    urlcolor=xkcdBluish,
    linktoc=all,
    citecolor=xkcdBrown
}
\usepackage{wrapfig}
\usepackage{makecell}

\makeatletter
\newcommand{\LeftComment}[1]{%
  \Statex \hspace*{\ALG@thistlm}\(\triangleright\) #1}
\makeatother

\usepackage{amsmath,amsfonts,bm}

\def\eqref#1{Eq.~(\ref{#1})}

\def\1{\bm{1}}

\def\rvq{{\mathbf{q}}}

\def\rvx{{\mathbf{x}}}

\DeclareMathAlphabet{\mathsfit}{\encodingdefault}{\sfdefault}{m}{sl}
\SetMathAlphabet{\mathsfit}{bold}{\encodingdefault}{\sfdefault}{bx}{n}

\newcommand{\R}{\mathbb{R}}

\usepackage{cleveref}

\usepackage{thmtools,thm-restate}

\usepackage{xspace}

\begin{document}

\runningtitle{ChronosX: Adapting Pretrained Time Series Models with Exogenous Variables}
\runningauthor{
Pineda, 
Mercado, 
Kapoor, 
Ansari, 
Stella, 
Shen, 
Turkmen, 
Shchur, 
Maddix, 
Bohlke, 
Wang, 
Rangapuram
}
\twocolumn[

\aistatstitle{ChronosX: Adapting Pretrained Time Series Models \\ with Exogenous Variables }

\aistatsauthor{
Sebastian Pineda Arango\textnormal{\textsuperscript{*2\textdagger}},
Pedro Mercado\textnormal{\textsuperscript{*1}}, 
Shubham Kapoor\textnormal{\textsuperscript{1}},
Abdul Fatir Ansari\textnormal{\textsuperscript{1}}, \\\bf
Lorenzo Stella\textnormal{\textsuperscript{1}},
Huibin Shen\textnormal{\textsuperscript{1}},
Hugo Senetaire\textnormal{\textsuperscript{3\textdagger}}, Caner Turkmen\textnormal{\textsuperscript{1}},
Oleksandr Shchur\textnormal{\textsuperscript{1}},\\\bf
Danielle C. Maddix\textnormal{\textsuperscript{1}},
Michael Bohlke-Schneider\textnormal{\textsuperscript{1}},
Yuyang Wang\textnormal{\textsuperscript{1}},
Syama Rangapuram\textnormal{\textsuperscript{1}}.
}

\aistatsaddress{ 
\textsuperscript{1}Amazon Web Services \textsuperscript{2}University of Freiburg 
\textsuperscript{3}Technical University of Denmark \\
}
]
\vspace{100pt}
\begin{abstract}
Covariates provide valuable information on external factors that influence time series and are critical in many real-world time series forecasting tasks. For example, in retail, covariates may indicate promotions or peak dates such as holiday seasons that heavily influence demand forecasts. Recent advances in pretraining large language model architectures for time series forecasting have led to highly accurate forecasters. However, the majority of these models do not readily use covariates as they are often specific to a certain task or domain. This paper introduces a new method to incorporate covariates into pretrained time series forecasting models. Our proposed approach incorporates covariate information into pretrained forecasting models through modular blocks that inject past and future covariate information, without necessarily modifying the pretrained model in consideration. In order to evaluate our approach, we introduce a benchmark composed of 32 different synthetic datasets with varying dynamics to evaluate the effectivity of forecasting models with covariates. Extensive evaluations on both synthetic and real datasets show that our approach effectively incorporates covariate information into pretrained models, outperforming existing baselines.
\end{abstract}
\section{Introduction}
One important component for time series forecasting --regardless of the forecasting model class-- is exogenous variables, also called covariates. Covariates provide external information to the forecasting model and allow the forecasting model to adjust the forecast based on this additional information. For example, the predicted energy output of a wind farm can be informed by additional weather information like the forecasted wind speed. While it is straightforward to incorporate covariate information into local statistical models and global deep learning (or other machine learning) models, it is less clear how this can be achieved for pretrained forecasting models. Pretrained forecasting models are trained across different datasets from different domains and every dataset might have a different number of variables with different properties, making it difficult to pretrain these models with covariates. Consequently, the majority of pretrained time series models do not support covariate data~\citep{dooley2023forecastpfn, goswami2024moment, das2023decoder, ansari2024Chronos}. One notable exception is Moirai, which introduces one possible way to address this problem with a mechanism called ``any-variate attention'' that can pretrain on datasets with a varying number of covariates~\citep{woo2024unified}. Nevertheless, how to include covariates into pretrained models that do not natively support covariates remains an open question.

In this paper, we present an approach called \thismodel to include covariate data into the pretrained forecasting model Chronos~\citep{ansari2024Chronos}. \thatmodel is a pretrained model that is trained on a time series dataset corpus without covariates. Drawing inspiration from modular deep learning~\citep{pfeiffer2023modular}, our proposed approach consists of attaching two modules. The first module updates the pretrained token embeddings with past covariates and the second module uses future covariates to adjust the output distribution. These light-weight adapter modules can be quickly trained for a downstream forecasting task even when the underlying pretrained model is frozen. Moreover, we show that the same framework can be extended to other pretrained models to incorporate covariates and showcase this with state-of-the-art pretrained models such as \timesfm~\citep{das2023decoder} and \moment~\citep{goswami2024moment}.

To validate our approach, we introduce a novel collection of 32 datasets that emulates different kinds of covariates together with several time dynamics.
We also demonstrate that \thismodel achieves low forecasting error across 18 real-world datasets with covariates. Furthermore, we show that the adapter approach can be used to finetune a pretrained time series model on a target dataset, which is much faster than full finetuning of the entire pretrained model. In summary, we make the following contributions:  
1) We introduce \thismodel, an adapter approach for incorporating covariates into pretrained models for time series forecasting, 
2) we show that the same framework can be extended to other pretrained models to ingest covariates and introduce \timesfmx and \momentx,
3) we introduce a novel benchmark of 32 synthetic datasets that allows us to evaluate how effective are forecasting models in leveraging covariate information under different dynamics,
4) we demonstrate that \thismodel achieves low forecasting error both for synthetic and real-world datasets with covariates.

\section{Background and Related Work}

\textbf{Time series forecasting} is the task of extrapolating a time series into the future based on its historical values and, optionally, a set of covariates.
These covariates are external variables that potentially influence the primary time series and may enhance the forecasting model's accuracy by providing additional context (e.g., weather may influence the sales of fans). 
In its general form, time series forecasting can be cast as the problem of modeling the conditional distribution, 
\begin{equation}
    P(\mathbf{z}_{C+1:H} | \mathbf{z}_{1:C}, \mathbf{X}_{1:H}; \Phi),
    \label{eq:tss-conditionalDistribution}
\end{equation}
where $\mathbf{z}_{1:C} = [z_1,\dots,z_C]$ is the historical context of the primary time series, $\mathbf{z}_{C+1:H} = [z_{C+1},\dots,z_H]$ is the future target until horizon $H$, $\mathbf{X}_{1:H} = [\mathbf{x}_1,\dots,\mathbf{x}_H]$, are covariates from the historical context until horizon $H$, and $\Phi$ denotes a set of learnable parameters.
In this paper, we focus on the case where the primary time series is univariate, $z_t \in \R$, while the covariates may have multiple dimensions, $\mathbf{x}_t \in \R^c$. 

A wide variety of time series forecasting models have been proposed in the literature.
Based on how they leverage time series data for training, they may be categorized into: \emph{local} and \emph{global} models.
Local models such as ARIMA~\citep{box2015time} and ETS~\citep{hyndman2008forecasting} are fit individually for each time series.
On the other hand, global models such as DeepAR~\citep{salinas2020deepar}, TFT~\citep{lim2021temporal}, PatchTST~\citep{Nie2023PatchTST}, and NHiTS~\citep{challu2023nhits} are trained across multiple time series from a given dataset, and are able to leverage global patterns present within a dataset.

\begin{figure*}[tp]
\centering
    \centering
    \includegraphics[width=\linewidth]{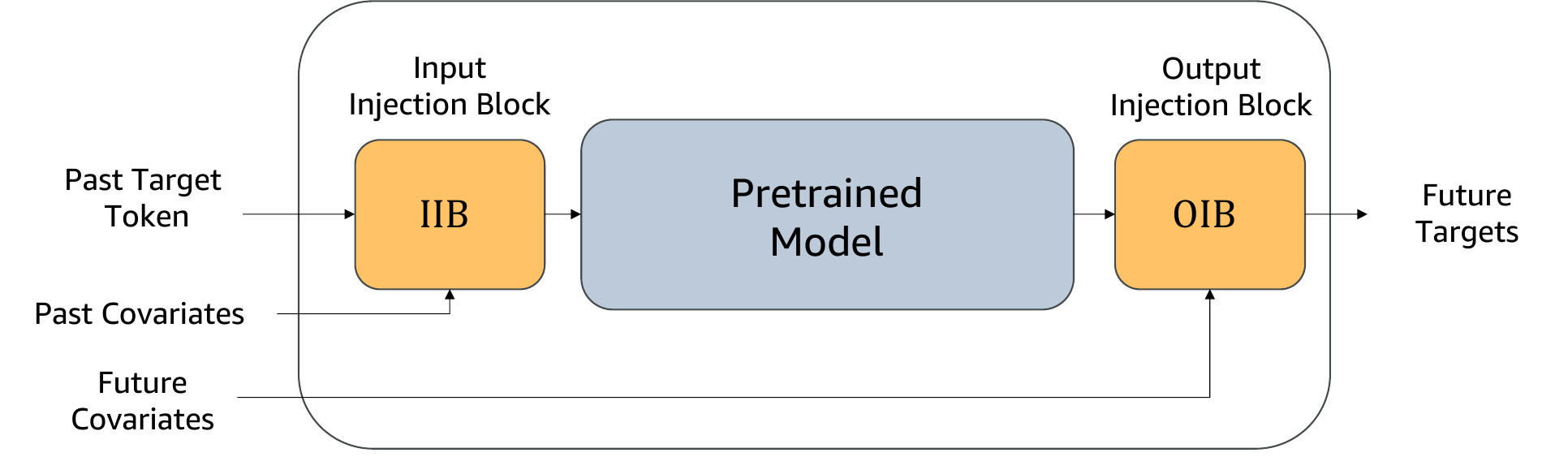}
    \caption{Architecture of \thismodel. It consists of two blocks: the first block adds covariates of the past and updates token embeddings. The second block adds future covariates updating the estimated distribution of the pretrained model.}\label{fig:architecture}
\end{figure*}

\textbf{Pretrained time series models.}
Recent work~\citep{ansari2024Chronos, woo2024unified, das2023decoder, goswami2024moment} on large-scale training of time series models has given rise to a category of global models called \emph{pretrained models} (also referred to as ``foundation'' models).
Pretrained models are trained on a large corpus of time series data and can be used for accurate zero-shot forecasting or finetuned on downstream datasets and tasks.
Moirai~\citep{woo2024unified}, \moment~\citep{goswami2024moment}, and \timesfm~\citep{das2023decoder} are pretrained models based on patching~\citep{Nie2023PatchTST}.
Specifically, they convert time series into patches before processing these patches with transformer-based models.
Further, TTM~\citep{ttm_2024} is a pretrained model built on top of model blocks of TSMixer~\citep{chen2023tsmixer}, which is an architecture designed by stacking multiple MLPs that allow to extract information efficiently from both time and feature dimensions.
On the other hand, \thatmodel~\citep{ansari2024Chronos} maps time series values into tokens from a fixed vocabulary via scaling and quantization, and trains existing language models architectures on such \emph{time series tokens}.
\citet{ansari2024Chronos} show that this simple tokenization scheme is very effective and \thatmodel\ models achieve zero-shot performance comparable to deep learning models with access to training data. 

\textbf{Forecasting with covariates.}
Covariates are essential to account for external events in forecasting models, such as peak dates in retail, planned strikes that might disrupt traffic, or sports events that generate massive online attendance.
Forecasting models incorporate covariates in different ways. For instance, in DeepAR~\citep{salinas2020deepar} covariates are part of the input to the RNN block, whereas TFT~\citep{lim2021temporal} explicitly encodes past and future covariates with different encoding networks whose outputs are concatenated.
ARIMA uses extra coefficients to account for covariates, and NBEATSx~\citep{olivares2023neural} and N-HiTS~\citep{challu2023nhits} concatenate the primary time series with past and future covariates and compute a fixed input size.
Moreover, \cite{ttm_2024} introduce TTM-CM, a fine-tuning strategy for TTM based on channel mixing that allows to integrate covariates. To the best of our knowledge, the only pretrained model that natively consumes covariates is Moirai~\citep{woo2024unified}, which flattens time series and covariates into a single sequence and assigns a variate ID to distinguish between target time series and covariates.
In this paper, we introduce \thismodel, an adapter approach inspired by modular deep learning~\citep{pfeiffer2023modular}, for incorporating covariates into the univariate \thatmodel~\citep{ansari2024Chronos} model which does not support covariates natively.

\section{\thismodel: Covariate Integration in Pretrained Models}\label{sec_proposed_method}

We now introduce the proposed model to add covariates to pretrained models. 
We denote by \textbf{\thismodel} the resulting model of adding covariates to the pretrained model \thatmodel~\citep{ansari2024Chronos}. See~\Cref{fig:architecture} for an overview of the model architecture.

\thatmodel is a pretrained model for univariate probabilistic time series forecasting based on the encoder-decoder T5 transformer models. \thatmodel applies a preprocessing scheme where time series are first mean-scaled and then tokenized through a suitable bin-quantization approach. The resulting tokenization is then passed to generate the corresponding pretrained embedding and fed into the encoder. \thatmodel uses the categorical distribution over the vocabulary of tokens as output distribution, and is trained to minimize the cross entropy between the distribution of the quantized ground truth label and the predicted distribution. For more details, we refer the reader to~\citep{ansari2024Chronos}. 

Our novel approach \thismodel adds the information of covariates in two different stages: 1) we add covariates from the past to update the corresponding token embeddings, and 2) we add covariates of the future to adjust the logits. Observe that, depending on the user context, one can choose to either work with covariates of the past, the future, or both.
Following a modular deep learning paradigm~\citep{pfeiffer2023modular}, we propose modules to achieve this while minimizing the modifications to the original model. 

\begin{figure*}[h]
    \begin{subfigure}{.45\linewidth}
    \includegraphics[width=\linewidth]{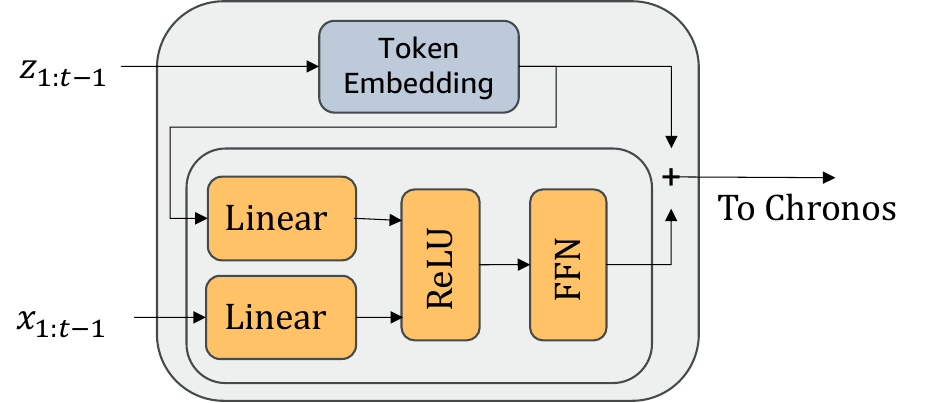}
    \caption{Input Injection Block (IIB)}\label{fig:IIB}
    \end{subfigure}
    \rulesep
    \begin{subfigure}{.45\linewidth}
    \includegraphics[width=\linewidth]{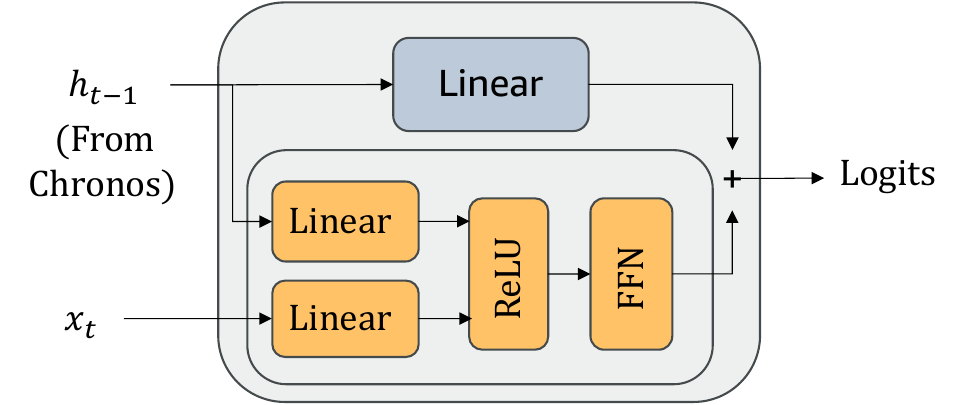}
    \caption{Output Injection Block (OIB)}\label{fig:OIB}
    \end{subfigure}
\caption{Input and Output injection blocks used in \thismodel. Observe that IIB and OIB take past and future covariates, respectively.
\Cref{fig:IIB} shows that the input injection block takes a pretrained tokenized embedding together with covariates of the past, whereas \Cref{fig:OIB} takes the pretrained logits (expressed as the final hidden state multiplied by a pretrained matrix) together with covariates of the future. The blue color indicates that the module is taken from the pretrained model.
}\label{fig:injetion_blocks}
\end{figure*}

\thismodel allows us to freeze any arbitrary portion of the original architecture and just update the proposed modules. In the following sections, we refer as Input Injection Block (IIB) and Output Injection Block (OIB) to the adapters injecting covariates from the past and the future, respectively.
See \Cref{fig:injetion_blocks} for a depiction of these.
Additionally, we will overload the notation and use $\mathbf{z}$ for tokenized time series. In what follows we will use fully connected feed-forward networks consisting of two linear transformations with a ReLU activation in between, $\text{FFN}(x)=\text{max}(0, xW_1+b_1)W_2 + b_2$.

\paragraph{Input Injection Block.}
To add information of covariates from the past we update the token embeddings. Specifically, for every time step $\mathbf{z}_{t-1}$ we pass the corresponding token embeddings and covariates through independent linear layers and apply an FFN to the ReLU of the concatenated mappings. Let 
$h_{\mathrm{emb}}(\mathbf{z}_{t-1})\in\mathbb{R}^{d_{\mathrm{model}}}$ 
be the pretrained token embedding, the updated embeddings $f_{\mathrm{IIB}}$ are defined as
\begin{equation}
f_{\mathrm{IIB}}\left(
\mathbf{z}_{t-1},
\mathbf{x}_{t-1}
\right) 
= 
h_{\mathrm{emb}}(\mathbf{z}_{t-1})
+
g_{\mathrm{IIB}}(h_{\mathrm{\text{emb}}}(\mathbf{z}_{t-1}),\mathbf{x}_{t-1})
\end{equation}
and $g_{\mathrm{IIB}}$ is defined as
\begin{equation}
    \begin{aligned}
     g_{\mathrm{IIB}}(h_{\mathrm{\text{emb}}}(\mathbf{z}_{t-1}),\mathbf{x}_{t-1}) = \qquad\qquad\\
     \mathrm{FFN}\left(\mathrm{ReLU}\left(h_{\mathrm{\text{emb}}}(\mathbf{z}_{t-1})W_{{\mathrm{IIB}}}^{({\mathrm{emb}})}   \oplus \mathbf{x}_{t-1}W_{{\mathrm{IIB}}}^{({\mathrm{cov}})} \right)\right)
    \end{aligned}
    \label{eq:g_iib}
\end{equation}
with linear mappings
$W_{{\mathrm{IIB}}}^{({\mathrm{emb}})}$ and 
$W_{{\mathrm{IIB}}}^{(\mathrm{cov})}$ and $\oplus$ represents the concatenation operator.

The function $g_{\mathrm{IIB}}$ adjusts the embeddings of the tokenized time series value $\mathbf{z}_{t-1}$ by merging information of covariates from the past 
$\mathbf{x}_{t-1}$ 
and the corresponding time series embedding $h_{\mathrm{\text{emb}}}(\mathbf{z}_{t-1})$.
\newpage
\paragraph{Output Injection Block.}
We leverage information from future covariates $\mathbf{x}_{t}$ by adjusting the logits of the pretrained model.
These are generated through a matrix multiplication between the last hidden state and a pretrained matrix~\citep{vaswani2017attention}.
The adjusted logits are defined as follows:
let 
$h_{\mathrm{out}}(\mathbf{z}_{t-1})$ 
be the final hidden state corresponding to $\mathbf{z}_{t-1}$, and 
$W_{\mathrm{out}}$ 
the pretrained matrix producing the logits in the model, then the adjusted logits $f_{\mathrm{OIB}}$ are defined as 
\begin{equation}
f_{\mathrm{OIB}}\left(
\mathbf{z}_{t-1},
\mathbf{x}_{t}
\right)
= 
h_{\mathrm{out}}(\mathbf{z}_{t-1}) W_{\mathrm{out}}
+ 
g_{\mathrm{OIB}}(h_{\mathrm{\mathrm{out}}}(\mathbf{z}_{t-1}),\mathbf{x}_{t})
\end{equation}
where $g_{\mathrm{OIB}}$
\begin{equation}
\begin{aligned}
    g_{\mathrm{OIB}}(h_{\mathrm{\text{out}}}(\mathbf{z}_{t-1}),\mathbf{x}_{t}) = \qquad\qquad\\ 
    \mathrm{FFN} \left(\mathrm{ReLU} \left(h_{\mathrm{\text{out}}}(\mathbf{z}_{t-1})W_{{\mathrm{OIB}}}^{({\mathrm{out}})}   \oplus \mathbf{x}_{t}W_{{\mathrm{OIB}}}^{({\mathrm{cov}})} \right)\right)
    \label{eq:g_oib}
\end{aligned}
\end{equation}
with parameter matrices
$W_{{\mathrm{OIB}}}^{({\mathrm{out}})}$ and $W_{{\mathrm{OIB}}}^{(\mathrm{cov})}$, and $\oplus$ represents the concatenation operator.
Observe that the inputs considered for this update are covariates from the future and the last hidden state evaluated on time series values from the past.

\textbf{Modular variants of~\thismodel.} Due to its modular nature, we can adapt \thismodel depending on the kind of covariates that we want to incorporate. In what follows we will denote by \thismodel the case where both past and future covariates are incorporated, i.e. IIB and OIB are used.
Yet, we can extend our model to the case where only covariates of the past are available by only considering the input injection block. Similarly, for the case where we only want to consume future covariates, we can consider only the output injection block. When training \thismodel, we train from scratch the parameters of the Feed-Forward Networks (FFN) and new matrices $W^{(\mathrm{emb})}_{\mathrm{IIB}}, W^{(\mathrm{cov})}_{\mathrm{IIB}}, W^{(\mathrm{cov})}_{\mathrm{OIB}}, W^{(\mathrm{out})}_{\mathrm{OIB}}$. Additionally, we consider a full-finetuned variant \textsc{ChronosX(FF)} where all the parameters, including the pretrained ones, are updated. 
Whereas in the main paper we will only report results for \thismodel, in the appendix we describe and present evaluations of additional variants. For instance, we study the case where either only past or future covariates are considered.

\begin{SCfigure*}[0.5]
    \centering
    \begin{subfigure}{1.4\linewidth}
        \newcolumntype{C}{ >{\centering\arraybackslash} m{0.25\linewidth} }
        \newcolumntype{D}{ >{\centering\arraybackslash} m{0.05 \linewidth} }
        \begin{tabular}{C D C D C}
            \includegraphics[trim={0.3cm 0.3cm 0.3cm 0.3cm}, width=\linewidth]{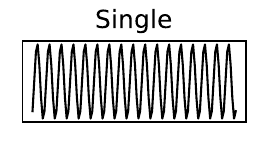} &
            $+$ &       \includegraphics[trim={0.3cm 0.3cm 0.3cm 0.3cm}, width=\linewidth]{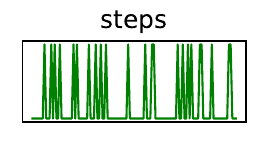} & $=$  &        \includegraphics[trim={0.3cm 0.3cm 0.3cm 0.3cm}, width=\linewidth]{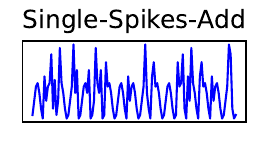} \\
            \includegraphics[trim={0.3cm 0.3cm 0.3cm 0.3cm}, width=\linewidth]{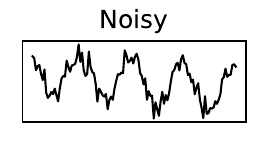} &
            $\times$ &       \includegraphics[trim={0.3cm 0.3cm 0.3cm 0.3cm}, width=\linewidth]{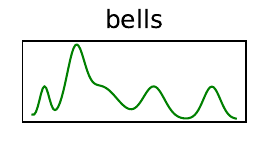} & $=$  &        \includegraphics[trim={0.3cm 0.3cm 0.3cm 0.3cm}, width=\linewidth]{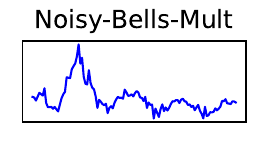} \\
        \end{tabular}
    \end{subfigure}
    \label{fig:metrics_vs_efficiency}
    \caption{\label{fig:synthetic_gen} 
    Illustration of the generation of synthetic time series. Different combinations of the $4$ possible main signals modified by one of the $4$ possible external covariates through a chosen operator ($+$ or $\times$), result in realistic synthetic time series with covariates.
    }
    \vspace{-15pt}
\end{SCfigure*}
\section{Extending Injection Blocks to Other Pretrained Models}
In the previous section, we introduced \thismodel, an extension for \thatmodel to incorporate covariates. 
Given the modular and minimally-invasive nature of our design, our approach can be extended to other pretrained models. We will showcase this with two pretrained models: \moment and \timesfm. Contrary to \thatmodel which is an encoder-decoder transformer, \moment and \timesfm are encoder-only and decoder-only models, respectively.
The main differences to consider with respect to \thatmodel are: \textit{i)} input patching and \textit{ii)} point forecasts, which imply minor modifications to our framework. 
The resulting extensions of pretrained models with covariates will be referred to as \textbf{\timesfmx} and \textbf{\momentx}.

\textbf{Extending on Input Patching. } 
Patching has received a relevant amount of attention in recent years, becoming a standard operation in recent forecasting models~\citep{Nie2023PatchTST,goswami2024moment, das2023decoder}. Patching is the operation of breaking a time series into potentially disjoint fixed-length subsequences.

To extend our framework to models that rely on a patched input $\tilde{\mathbf{z}}_{t-1} \in \mathbb{R}^{P_n\times P_d }$, we also create patches for the covariates $\tilde{\mathbf{x}}_{t-1}=\mathrm{Patch}(\mathbf{x}_{t-1}) \in \mathbb{R}^{P_n \times (P_d\cdot c) }$, where $P_d,P_n, c$ denote the patch dimension, the number of patches and the number of covariates. In this manner, we guarantee that both the primary time series input patches and the 
covariate patches have the same length. Subsequently, we feed these patched inputs into the modules as in Equation \ref{eq:g_iib}.

\textbf{Extending on Point Forecast.} 
Our framework can be extended to the case of point forecasts as follows. Instead of considering a categorical distribution, we can take a given point forecast generated by the current model $\hat{\mathbf{z}}_t=h_{\mathrm{out}}(\mathbf{z}_{t-1}) W_{\mathrm{out}}$ and 
apply the following update rule:
\begin{equation}
   \begin{aligned}
        f_{\mathrm{OIB}}(h_{\mathrm{\text{out}}}(\mathbf{z}_{t-1}),\mathbf{x}_{t}, \hat{\mathbf{z}}_t) =\hat {\mathbf{z}}_t+\qquad\qquad\qquad\\ 
        \mathrm{FFN} \left(\mathrm{ReLU} \left(h_{\mathrm{\text{out}}}(\mathbf{z}_{t-1})W_{{\mathrm{OIB}}}^{({\mathrm{out}})}   \oplus \mathbf{x}_{t}W_{{\mathrm{OIB}}}^{({\mathrm{cov}})}   \oplus \hat{\mathbf{z}}_tW_{{\mathrm{OIB}}}^{({\mathrm{p}})}\right)\right)
        \label{eq:g_oib}
    \end{aligned} 
\end{equation}

\section{Synthetic Datasets with Covariates}\label{sec:synthetic-data-explanation}
Although the task of forecasting time series with covariates is a highly relevant task, there is a limited amount of freely available time series datasets with covariates where forecasting models can be evaluated. To overcome this limitation in this paper we introduce a collection of 32 different synthetic datasets of time series with covariates carefully handcrafted to emulate real-life applications where external factors greatly impact the time series. 

Each of the datasets contains 100 times series of daily frequency extending over 1827 days\footnote{This corresponds to the number of days in the fictional time series between 01/01/2025 and 31/12/2029.}.  For each dataset, we consider 3 basic elements that allow for the generation of 100 times series with covariates: 1)~the \textbf{main signal} ($\hat{z}_t$), 2)~an \textbf{external covariate} ($x_t$), and 3) a \textbf{combination operator} $\odot\in\{+, \times\}$. The generation of time series with these three ingredients is defined by $z_t = \hat{z}_t \odot x_t$. In what follows we present several variations of these three basic elements with the goal of providing a comprehensive evaluation of suitable forecasting models for time series with covariates.

The time series representing the main signal $\hat{z}_t$ are divided in two types: \textbf{Simple} and \textbf{Complex}. 
The simple synthetic type includes single sinusoids and simple sinusoids variants, whereas the complex synthetic type includes diverse sinusoids and noisy sinusoids variants. We now describe each of the four different variants of the main signal $\hat{z}_t$ with increasing complexity with the form of~\eqref{equation:main_signal_shape_main}, mainly:
1) \textit{Single sinusoids} considers a standard sine curve with fixed zero phase, unit amplitude and period of 7 days corresponding to one week for the entire dataset; 
2) \textit{Simple sinusoids} is the sum of three sine curves each with a weekly, monthly and yearly period, together with a randomly sampled amplitude and zero phase;
3) \textit{Diverse sinusoids} is the result of combining three sine curves with randomly sampled phases and amplitudes and with a global trend, and
4) \textit{Noisy sinusoids} which results from adding gaussian noise ($\epsilon$) to \textit{diverse sinusoids} and allows us to assess model robustness.
Please see the Appendix for more details on the process to generate the dataset.
\vspace{-4pt}
\begin{equation}\label{equation:main_signal_shape_main}
\hat z_t = \sum_{i=1}^3 a_i \sin \left(f_i t + \phi_i \right)
                                + b_1 t  + b_2
                                + \epsilon
\vspace{-8pt}
\end{equation}

We design four different kinds of \textbf{external covariates}: 1) \textit{spikes} models short time events such as strikes or power failure, 2) \textit{steps} model longer events with sudden and sustained changes such as discounts in the retail industry, 3) \textit{bells} represent smoother changes such as festive season, and 4) \textit{autoregressive process} with randomly generated parameters that model cases where the value of covariates is determined by previous observations. Finally, the \textbf{combination operator} $\odot$ is defined either as the addition $+$ or the multiplication $\times$ between the main signal ($\hat{z}_t$) and the covariates ($x_t$). An illustration of the generative process for this synthetic datasets along with a few examples of the resulting time series is depicted in \cref{fig:synthetic_gen}. 


\section{Experiments and Results}\label{sec:experiments_and_results}

\textbf{Setup.} In all experiments for our models we consider learning rates $\{10^{-2}, 10^{-3}, 10^{-4}\}$ and select the one with the best validation error with a validation window of length equal to the prediction length of the dataset in consideration. A detailed description of the hyperparameters is presented in the appendix.

\begin{figure*}[h]
    \centering
    \begin{subfigure}{.48\linewidth}
    \includegraphics[width=1\linewidth]{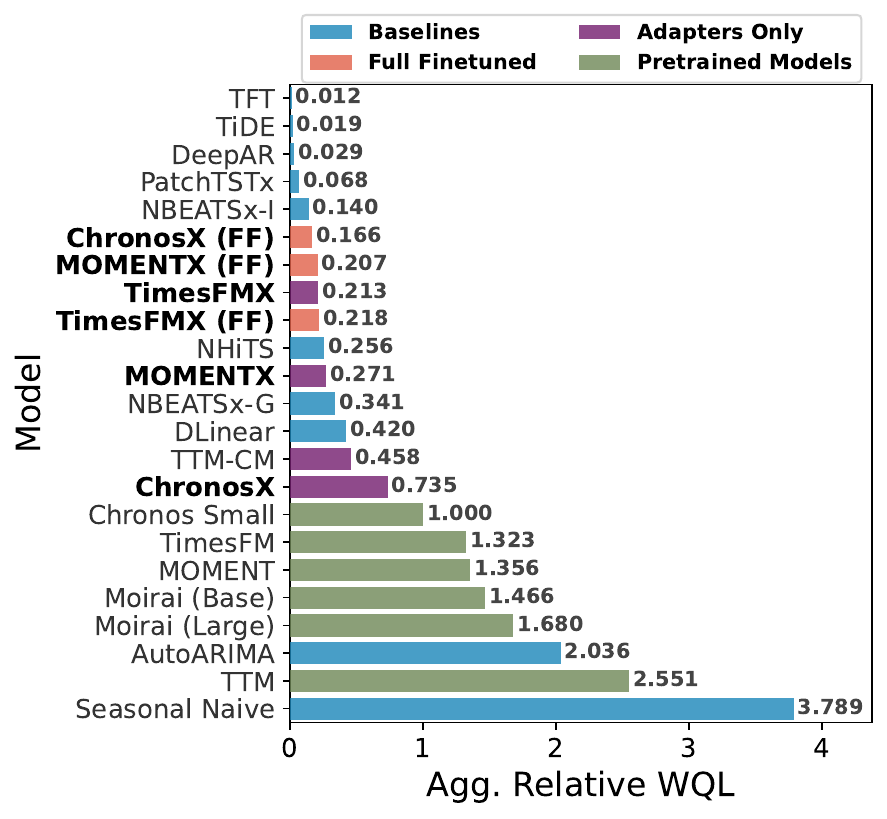}
    \caption{Agg. Rel (WQL) on \textbf{Simple} synthetic datasets.}
    \label{fig:datasets_synthetic_simple_wql_main}
    \end{subfigure}
    \quad
    \begin{subfigure}{.48\linewidth}
    \includegraphics[width=1\linewidth]{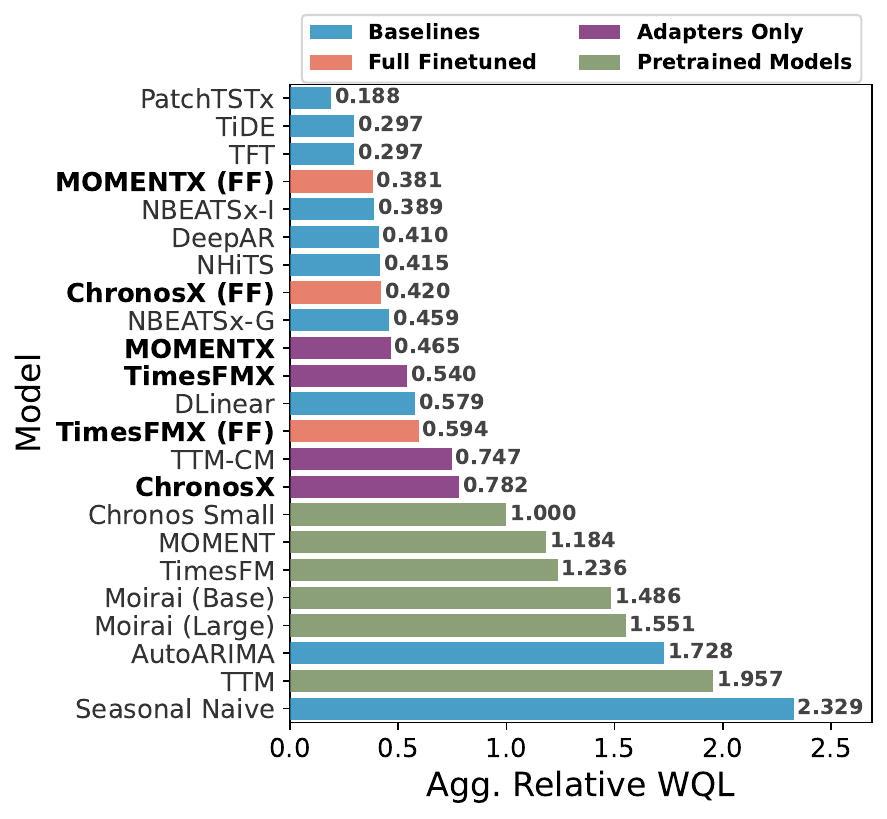}
    \caption{Agg. Rel (WQL) on \textbf{Complex} synthetic datasets.}
    \label{fig:datasets_synthetic_complex_wql_main}
    \end{subfigure}
    \caption{Evaluations on Simple and Complex datasets as introduced in~\Cref{sec:synthetic-data-explanation}. Scores are normalized by \textsc{Chronos Small}. We can see that our proposed models \textsc{ChronosX} and \textsc{ChronosX(FF)} effectively incorporate covariates. 
    }
    \label{fig:datasets_synthetic_main}
\end{figure*}

\textbf{Metrics}. We consider the Weighted Quantile Loss (WQL) to evaluate probabilistic forecasts, which measure the alignment of quantile levels of the predictive distribution with respect to the ground truth~\citep{gneiting2007strictly, pmlr-v89-gasthaus19a,shchur2023autogluon}. In all experiments the WQL is computed on quantile levels \{0.1, 0.2, \ldots, 0.9\}. For methods generating sample forecasts we compute the quantiles based on 100 samples, whereas quantile forecasting methods are trained on the same quantile levels we use for evaluation. We report the Mean Absolute Scaled Error (MASE), which measures deviations from the median forecast with the ground-truth values~\citep{hyndman2006another}. Finally, we follow~\citep{fleming1986not,ansari2024Chronos,woo2024unified} and report the aggregated WQL and MASE computed as the geometric mean of normalized scores by the corresponding baseline. 
\begin{figure*}[h]
    \vspace{-7pt}
    \centering
    \begin{subfigure}{.48\linewidth}
    \includegraphics[width=1\linewidth]{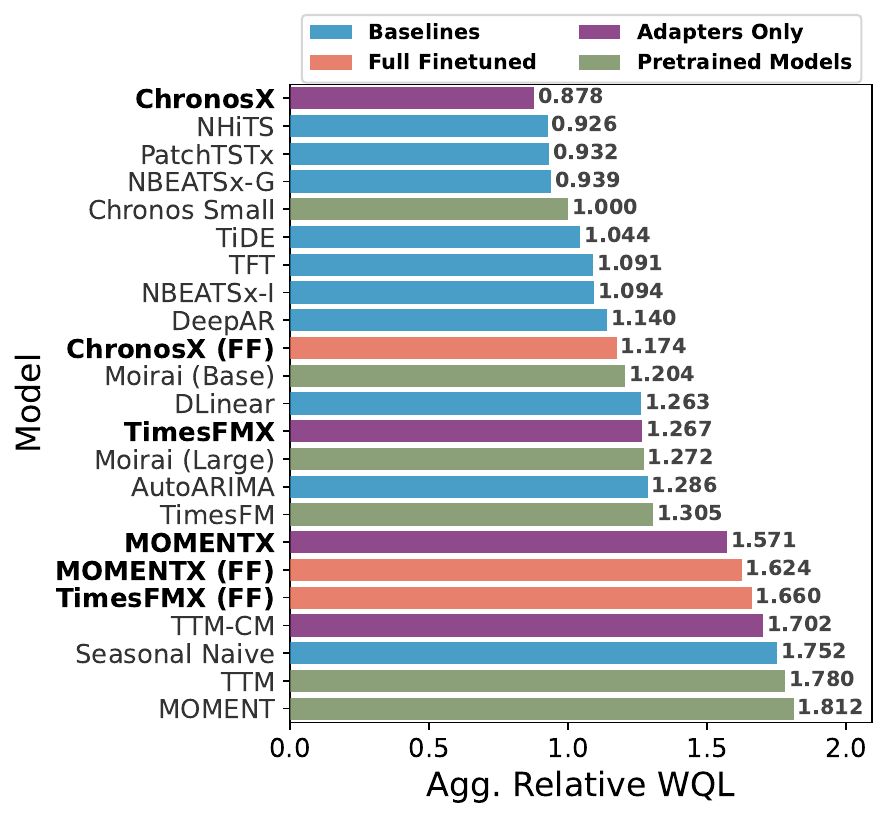}
    \caption{Agg. Rel. WQL on real datasets with covariates.}
    \label{fig:ciptfinal_real_with_covs_wql}
    \end{subfigure}
    \quad
    \begin{subfigure}{.475\linewidth}
    \includegraphics[width=1\linewidth]{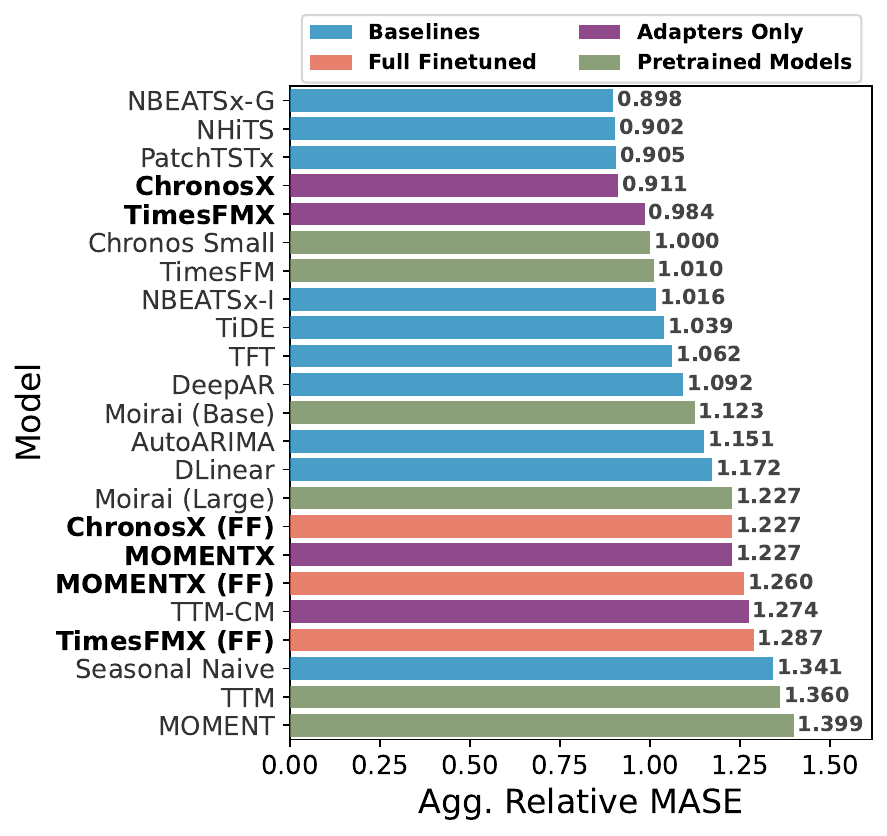}
    \caption{Agg. Rel. MASE on real datasets with covariates.}
    \label{fig:ciptfinal_real_with_covs_mase}
    \end{subfigure}
    \caption{Evaluations on real datasets with covariates. We report the Aggregated Relative WQL and MASE. 
    }
    \label{fig:ciptfinal_real_with_covs_grouped_main}
    \vspace{-0.5em}
\end{figure*}

\subsection{Finetuning and Adapters on Synthetic Benchmark Datasets with Covariates}
\label{sec:exps_datasets_synthetic_datasets}
In this section we evaluate how effective different forecasting models are in incorporating external information through covariates under different dynamics. 

We take the 32 different datasets as introduced in~\Cref{sec:synthetic-data-explanation} and follow the suggested splits Simple and Complex dataset collections each consisting of 16 datasets, and for each dataset we generate 100 time series of daily frequency with length 1827 and prediction length of 30. Recall that our adapter-based approach \thismodel, and its extensions \momentx and \timesfmx, incorporates both past and future covariates. Further, we consider a full finetuning variant per model and denote it by~\textsc{ChronosX(FF)}, \textsc{MOMENTX(FF)}, \textsc{TimesFMX(FF)}, which updates all parameters including those of the pretrained model. 
We also consider baselines that include covariates like DeepAR, PatchTSTx (an extension of PatchTST that supports covariates; see the supplementary appendix for details), NHiTS, and NBEATSx with its interpretable and general variants, together with AutoArima, which incorporates covariates through residuals. For all methods we standardize the covariates by the mean of absolute values. All scores are normalized by the scores of \textsc{Chronos Small}, which is a variant of \thatmodel with 46M parameters.

Figures~\ref{fig:datasets_synthetic_simple_wql_main}~and~\ref{fig:datasets_synthetic_complex_wql_main} present the aggregated relative WQL on \textit{Simple} and \textit{Complex} synthetic datasets. For both types of datasets, our proposed adapter-only approaches \thismodel, \timesfmx, and \momentx, together with their full fine-tuning variants (\thismodel(FF), \timesfmx(FF), and \momentx(FF)) incorporate covariates, as they clearly outperform the zero-shot performance of their pretrained models, i.e. \textsc{Chronos Small}, \timesfm, and \moment, respectively. In particular, we can see that \thismodel outperforms \thatmodel by 22\% in both WQL and MASE.
Further, we observe that baseline models like TFT, DeepAR and PatchTSTx perform particularly well in \textit{Simple} datasets, verifying that they effectively incorporate covariates in simple cases, whereas for \textit{Complex} datasets the errors are larger, yet most of baselines outperform the zero-shot performance of \textsc{Chronos Small}.
Moreover, we can see that in general full fine-tuning further improves the performance, which can be observed with \thismodel(FF) and \momentx(FF), whereas for \timesfmx(FF) in general it seems that there is a slight decrease in performance with complex synthetic datasets.

Overall, we can see that our proposed benchmark of synthetic datasets is useful to discern the effectivity of different models to incorporate covariates in time series. In particular, we have seen that most of the baselines here considered incorporate covariates that yield better scores than methods that do not take covariates, and that our covariate extensions for pretrained models are effectively incorporating covariate information

\subsection{Finetuning and Adapters on Real Datasets with Covariates.}
\label{sec:exps_datasets_real_datasets_with_covariates}
In this section, we evaluate our proposed models on 18 real datasets with covariates taken from diverse sources~\citep{haoyietal-informer-2021, makridakis2022m5, godahewa2021monash, woo2024unified, wang2023benchmarks}.
These datasets encompass data from fields such as retail, nature, transport, and mostly energy. Frequencies include hourly, daily, and 15-minute intervals, with forecast horizons up to 30 steps ahead. Please see the appendix for further details.

In~\Cref{fig:ciptfinal_real_with_covs_grouped_main} we can see that our adapter-based approach \thismodel consistently performs best among pretrained models adapted to include covariates \timesfmx and \momentx. 
Moreover, we can see that adapter-only variants \thismodel, \timesfmx, and \momentx perform better than their zero-shot counterparts \thatmodel, \timesfm, and \moment, respectively, further verifying that our models incorporate covariate information.
Additionally, we can see that full fine-tuning variants \thismodel(FF), \timesfmx(FF), and \momentx(FF) underperform with respect to several of the baselines here considered. This is likely due to the fact that several datasets consist of only one time series, and hence having a model with less trainable parameters provides an advantage in data-sparse regimes. 
Further, we can see that our model \thismodel performs best in WQL, and in terms of MASE it performs in the top-5 models together with \timesfmx. On the other side, despite the fact that \moment underperforms other models in general, we can see that our covariate extensions \momentx and \momentx(FF) significantly improve its performance, verifying that our adaption of pretrained models to incorporate covariates is effectively consuming this kind of information.

Overall we have shown that our proposed covariate extensions \thismodel, \timesfmx, and \momentx effectively incorporate covariate information in real datasets and \thismodel performs best in WQL and is comparable to other models in MASE.

\begin{figure*}[h]
    \vspace{-5pt}
    \centering
    \begin{subfigure}{.325\linewidth}
        \includegraphics[width=1\linewidth]{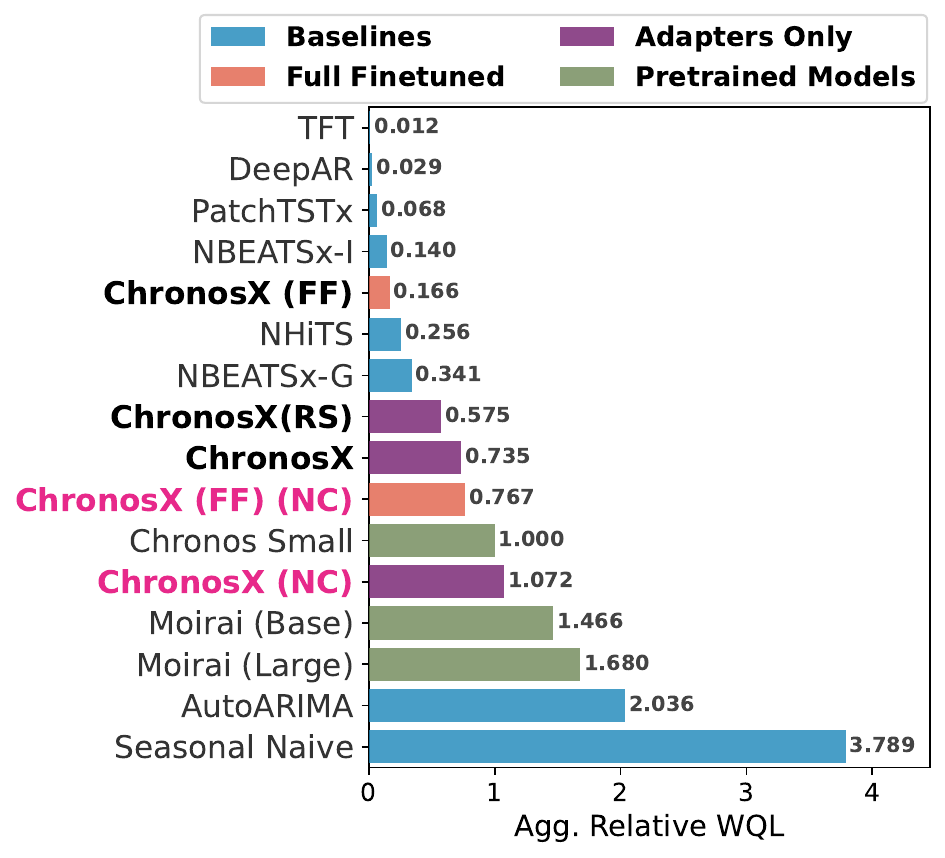}
        \caption{WQL on \textbf{Simple} Synthetic Data.}
        \label{fig:datasets_synthetic_simple_wql_nocovs_main}
    \end{subfigure}
    \begin{subfigure}{.325\linewidth}
        \includegraphics[width=1\linewidth]{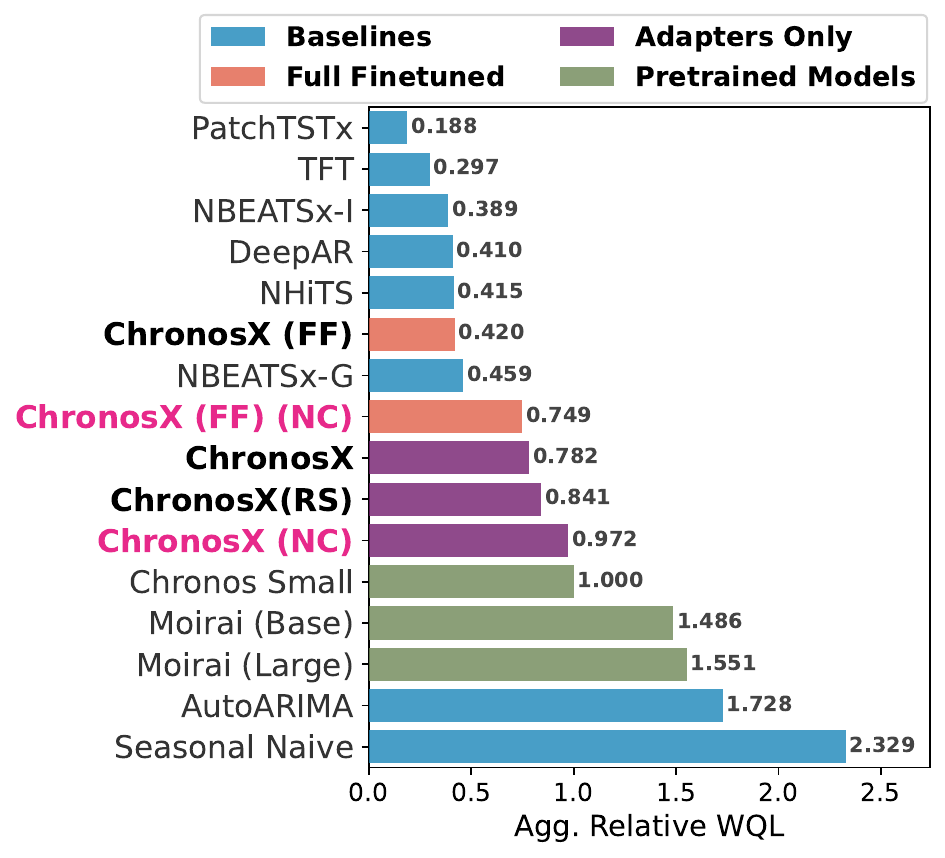}
        \caption{WQL on \textbf{Complex} Synthetic Data.}
        \label{fig:datasets_synthetic_complex_wql_nocovs_main}
    \end{subfigure}
\begin{subfigure}{.325\linewidth}
    \includegraphics[width=1\linewidth]{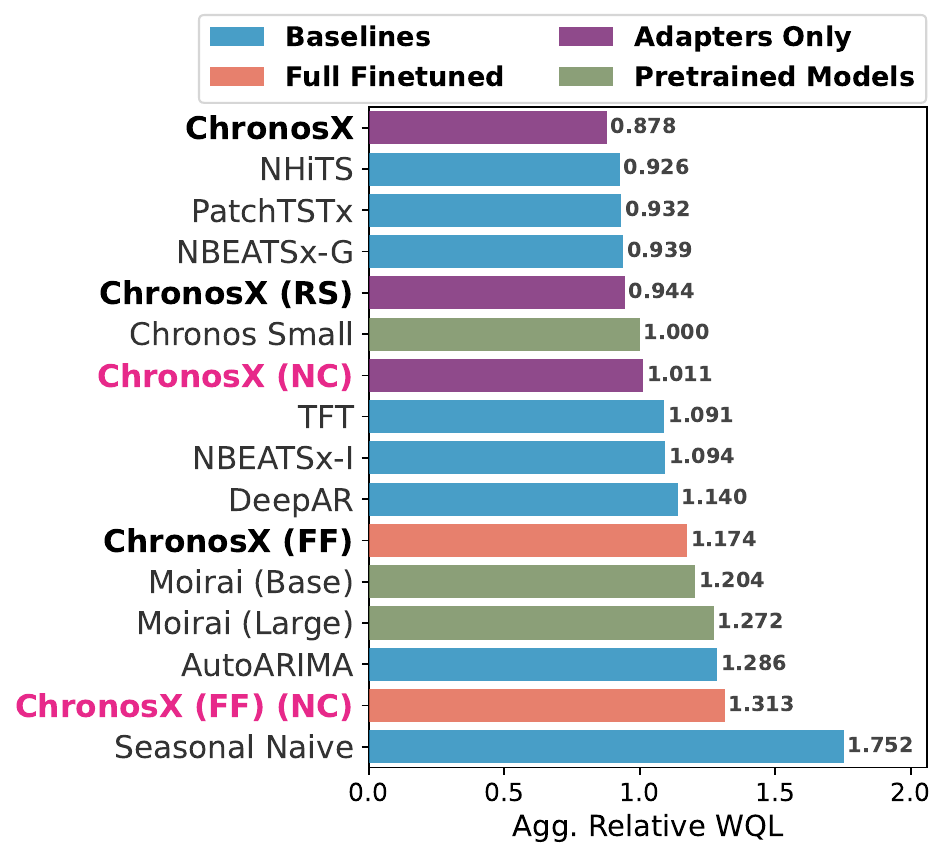}
    \caption{WQL  \textbf{Real}  Datasets.}
    \label{fig:datasets_real_wql_nocovs_main}
\end{subfigure}
\caption{
    Covariate ablation study. \thismodel (NC) and \thismodel (FF) (NC) are variants with no-covariates of our models \thismodel and \thismodel (FF), respectively. Overall model ablations with no-covariates perform worse.
}
\label{fig:wql_nocovs_main}
\end{figure*}

\begin{figure*}[h]
    \centering
    \begin{subfigure}{.33\linewidth}
        \includegraphics[width=1\linewidth]{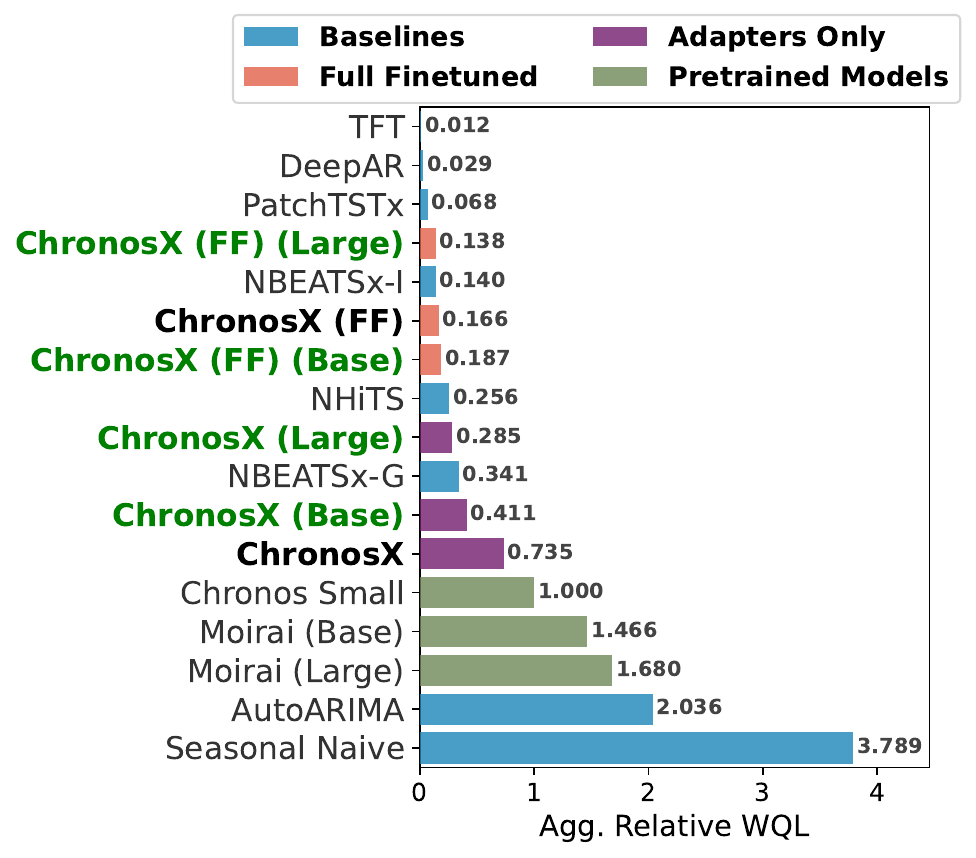}
        \caption{WQL on \textbf{Simple} Synthetic Data.}
        \label{fig:datasets_synthetic_simple_wql_model_size_main}
    \end{subfigure}
    \begin{subfigure}{.325\linewidth}
        \includegraphics[width=1\linewidth]{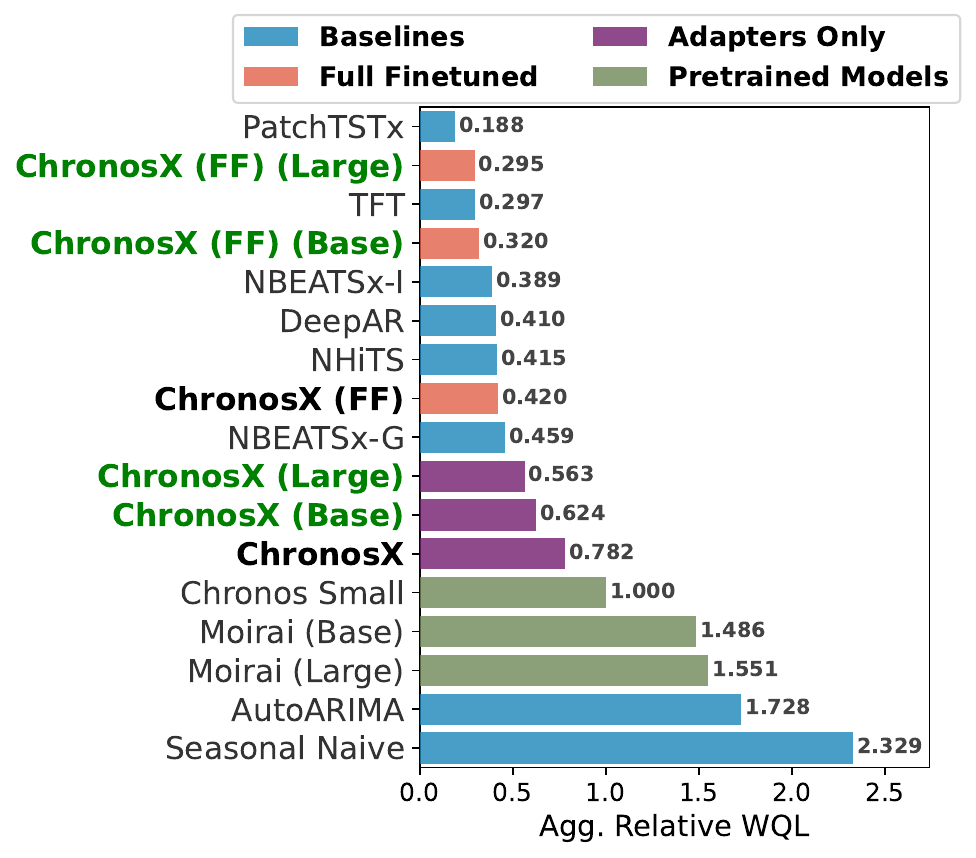}
        \caption{WQL on \textbf{Complex} Synthetic Data.}
        \label{fig:datasets_synthetic_complex_wql_model_size_main}
    \end{subfigure}
\begin{subfigure}{.33\linewidth}
    \includegraphics[width=1\linewidth]{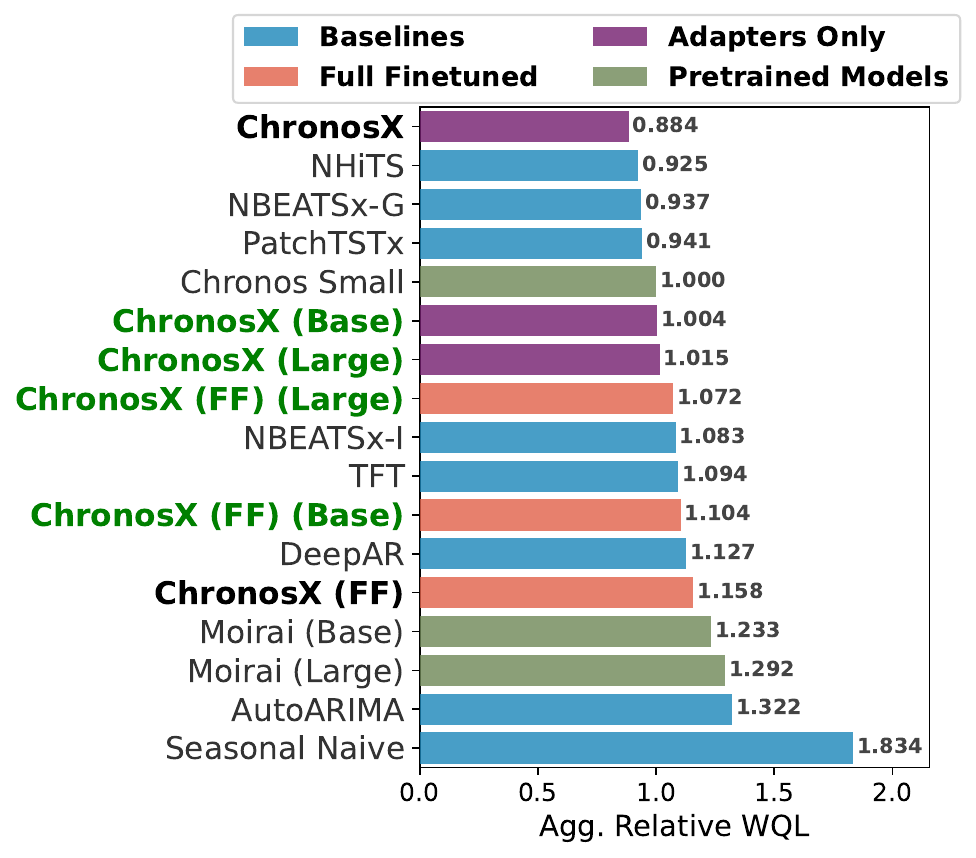}
    \caption{WQL  \textbf{Real}  Datasets}
    \label{fig:datasets_synthetic_complex_wql_model_size_main}
\end{subfigure}
\caption{
    Model size ablation. \thismodel (Base) and \thismodel (Large) are variants of \thismodel with backbone models with 200M and 710M parameters. Overall, larger backbone models improve performance with full fine-tuning.
}
\vspace{-0.5em}
\label{fig:wql_datasets_chronos_sizes}
\end{figure*}

\subsection{Ablations of Covariate Adapters}
In this section we explore the impact of covariates and backbone model sizes in ChronosX.

\textbf{Impact of covariates in ChronosX}.
We analyze if the performance improvement of ChronosX is obtained from the covariate information or the extra parameters included by the covariate modules. We run experiments using the Input and Output blocks, but omitting the covariates by dropping the matrices related to covariates, inducing the following variations:
\begin{equation}
    \begin{aligned}
        f_{\mathrm{IIB-NC}}&\left(
        \mathbf{z}_{t-1},
        \mathbf{x}_{t-1}
        \right) 
        = h_{\mathrm{emb}}(\mathbf{z}_{t-1}) +\\
        &\mathrm{FFN}\left(\mathrm{ReLU}\left(h_{\mathrm{\text{emb}}}(\mathbf{z}_{t-1})W_{{\mathrm{IIB}}}^{({\mathrm{emb}})}  \right)\right)
    \end{aligned}
\end{equation}
\begin{equation}
    \begin{aligned}
        f_{\mathrm{OIB-NC}}&\left(
        \mathbf{z}_{t-1},
        \mathbf{x}_{t}
        \right)
        = h_{\mathrm{out}}(\mathbf{z}_{t-1}) W_{\mathrm{out}} +\\
        &\mathrm{FFN} \left(\mathrm{ReLU} \left(h_{\mathrm{\text{out}}}(\mathbf{z}_{t-1})W_{{\mathrm{OIB}}}^{({\mathrm{out}})} \right)\right)
    \end{aligned}
\end{equation}

Evaluations of these ablations are presented in Fig.~\ref{fig:wql_nocovs_main}, where the ablations without covariates are denoted \thismodel(NC) and  \thismodel(FF) (NC).
We can see across synthetic and real datasets that our model \thismodel and its full-finetuned version \thismodel(FF) perform better than its corresponding ablations, \thismodel(NC) and  \thismodel(FF) (NC), respectively. Thus, we verify the performance improvement observed by our proposed model \thismodel is due to effectively consuming the information encoded through covariates. Finally, we provide an adapter-only variant that consumes covariates of the future through a linear transformation, denoted by \thismodel(RS)
\begin{equation}
f_{\mathrm{RS}}\left(
\mathbf{z}_{1:t-1},
\mathbf{x}_{t}
\right) 
= 
h_{\mathrm{out}}(\mathbf{z}_{1:t-1})W_{\mathrm{out}}
+ 
\mathbf{x}_{t}W_{{\mathrm{OIB}}}^{({\mathrm{cov}})}
  \label{eq:residual_block}
\end{equation}
We observe that this residual-based model \thismodel(RS) systematically outperforms its adapter-only ablation \thismodel(NC), further verifying the relevance of covariate information to improve performance.

\textbf{Impact of Backbone Size in ChronosX}.
In our experiments we have used for our model \thismodel the backbone model \thatmodel in its small version (46M). In this ablation, we study the effect of using the size variants \textit{Base} and \textit{Large} of \thatmodel, which have 200M and 710M parameters, respectively. The results in Fig.~\ref{fig:wql_datasets_chronos_sizes} demonstrate that increasing the model capacity is particularly helpful in complex synthetic datasets.
For real datasets we can see that full fine-tuning approaches improve their performance with larger backbone models, i.e. \thismodel(FF)(Base) and \thismodel(FF)(Large), whereas adapter-only variants decrease in performance.
Hence, we observe that using the backbone size \textit{Small}, as done with our model \thismodel, provides a good balance between model size and forecast quality.

\begin{table*}[t]
    \centering
    \caption{Definition of three alternative versions of \thismodel: with a single linear layer (OL), without linear layer but with activation (NL), and without linear layer and activation (NL-NR).}
    \resizebox{\textwidth}{!}{
        \begin{tabular}{ccc} \bottomrule
            \multirow{2}{*}{\textbf{Ablation}} &  \textbf{Input Injection Block} & \textbf{Output Injection Block}\\
              &  $g_{\mathrm{IIB}}(h_{\mathrm{\text{emb}}}(\mathbf{z}_{t-1}),\mathbf{x}_{t-1})$ & $g_{\mathrm{OIB}}(h_{\mathrm{\mathrm{out}}}(\mathbf{z}_{t-1}),\mathbf{x}_{t})$\\  \midrule
            One Linear (OL) &          $\mathrm{FFN}\left(\mathrm{ReLU}\left(\left(h_{\mathrm{\text{emb}}}(\mathbf{z}_{t-1})   \oplus \mathbf{x}_{t-1} \right)W_{\text{IIB}}^{\text{(OL)}}\right)\right)$ & $ \mathrm{FFN} \left(\mathrm{ReLU} \left(\left(h_{\mathrm{\text{out}}}(\mathbf{z}_{t-1})   \oplus \mathbf{x}_{t} \right)W_{\text{OIB}}^{\text{(OL)}}\right)\right) $ \\ 
            
            No Linear (NL) &  $\mathrm{FFN}\left(\mathrm{ReLU}\left(h_{\mathrm{\text{emb}}}(\mathbf{z}_{t-1})   \oplus \mathbf{x}_{t-1}\right)\right)$ & 
            $\mathrm{FFN} \left(\mathrm{ReLU}\left(h_{\mathrm{\text{out}}}(\mathbf{z}_{t-1})   \oplus \mathbf{x}_{t} \right)\right)$ \\
    
            No Linear - No ReLU & $ \text{FFN}(h_{\text{emb}}(\mathbf{z}_{t-1}) \oplus \mathbf{x}_{t-1})$ &  $\text{FFN}(h_{\text{out}}(\mathbf{z}_{t-1}) \oplus \mathbf{x}_{t})$ \\ \bottomrule
        \end{tabular}
    }
        \label{tab:adapters_impact_arch}
\end{table*}

\begin{figure*}
    \centering
    \begin{subfigure}{.325\linewidth}
        \includegraphics[width=1\linewidth]{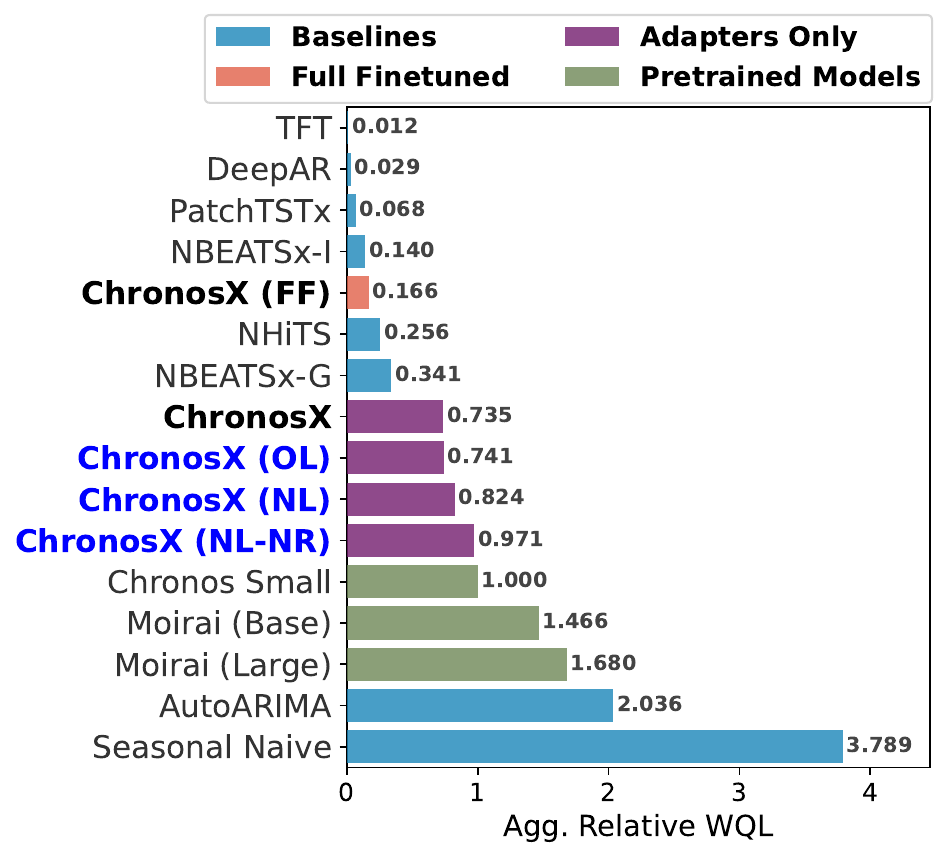}
        \caption{WQL on \textbf{Simple} Synthetic Data.}
        \label{fig:impact_arch1}
    \end{subfigure}
    \begin{subfigure}{.325\linewidth}
        \includegraphics[width=1\linewidth]{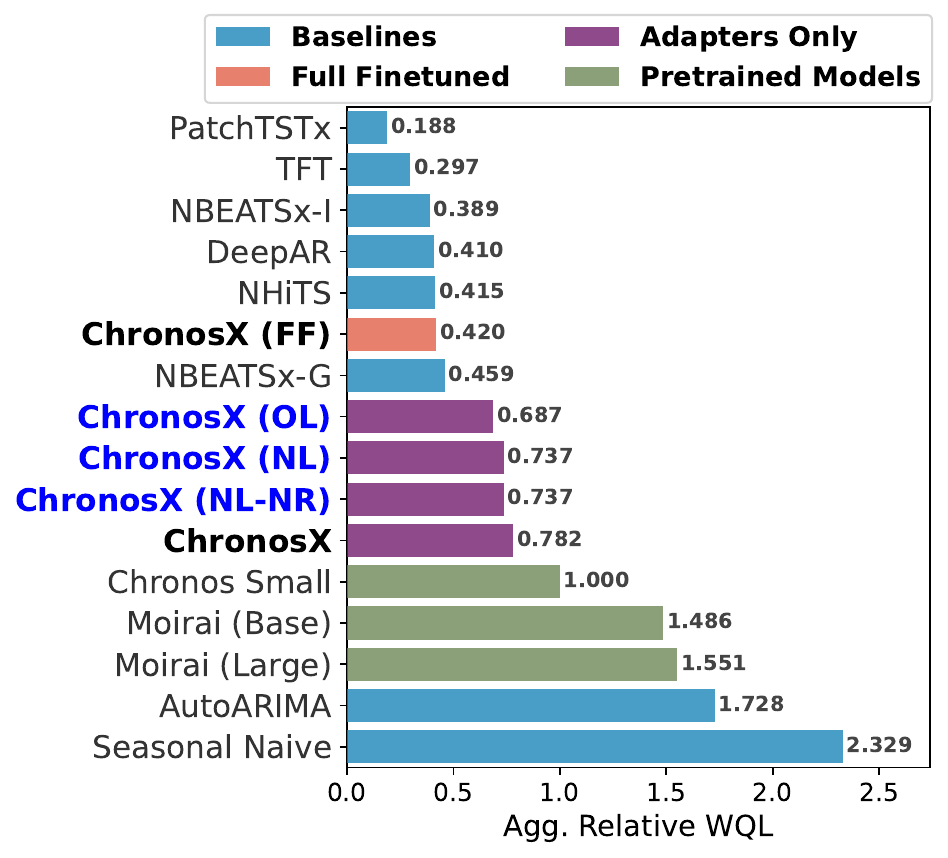}
        \caption{WQL on \textbf{Complex} Synthetic Data.}
        \label{fig:impact_arch2}
    \end{subfigure}
\begin{subfigure}{.325\linewidth}
    \includegraphics[width=1\linewidth]{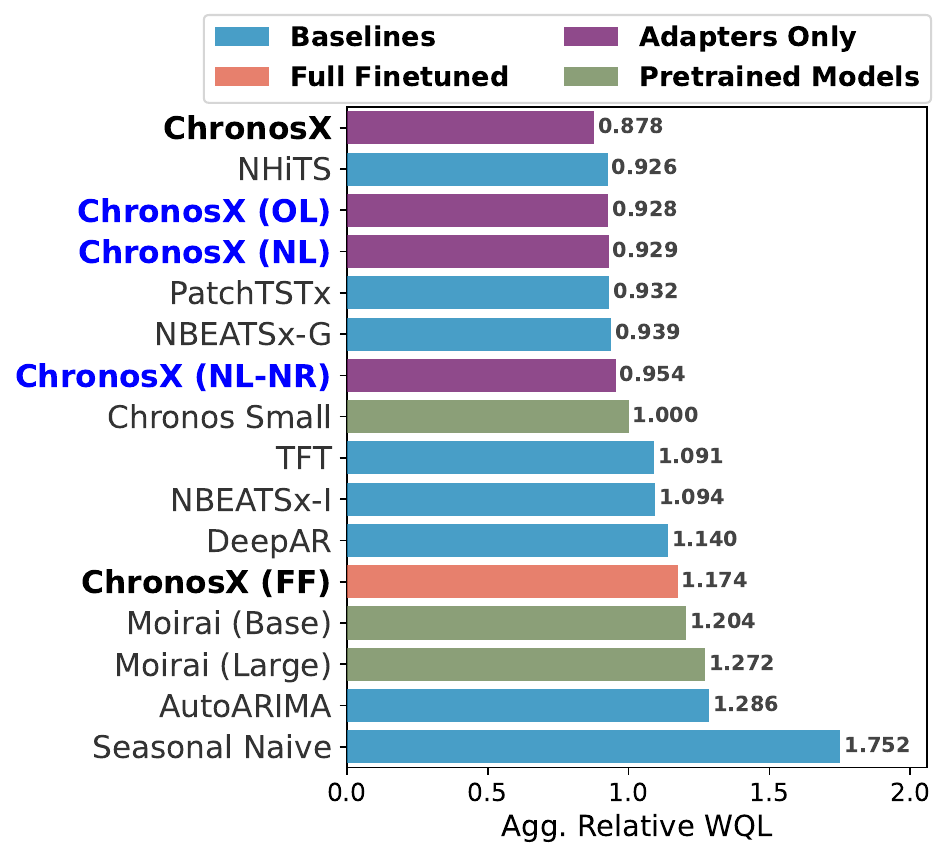}
    \caption{WQL  \textbf{Real}  Datasets.}
    \label{fig:impact_arch3}
\end{subfigure}
\caption{Results of ablations of \thismodel with different covariate adapters' architectures, i.e. \thismodel (OL), \thismodel (NL), and \thismodel (NL-NR). Overall, our proposed architecture outperforms these variants.}
\label{fig:impact_arch}
\end{figure*}

\textbf{Impact of Adapters Architecture}. Our proposed adapters consider two independent linear layers, followed by a Feed-Forward Network, as explained in Section \ref{sec_proposed_method}. One linear layer operates on embeddings related to the target time series (token embedding and hidden state), while operates on the covariates. This design allows the model to learn interactions independently, and has been explored in previous work~\citep{pineda2023_deeppipe}. However, to further justify our design, we compare it with three other alternatives. Firstly, we consider \thismodel with one linear layer \textit{(OL)} that receives the concatenated inputs of the embeddings of the target time series and their respective covariates. We also study \thismodel without the initial linear layers \textit{(NL)}, and just an activation function before FFN. Finally, we ablate \thismodel without initial linear layers and activations \textit{(NL-NR)}, just using a FFN. In Table \ref{tab:adapters_impact_arch}, we formalize these variants mathematically both for the input and output modules, following the same notation as in our original formulation. Figure \ref{fig:impact_arch} reports the results after evaluating the architectures in our datasets. In general, we observe that our original formulation \thismodel outperforms the other alternatives in real datasets and simple synthetic data. In complex data, \thismodel(OL) obtained the best results among the adapter-only variants, but still is outperformed by \thismodel(FF). Interestingly, we observe that \thismodel(OL) is consistently better than \thismodel(NL) and \thismodel(NL-NR), which highlights the importance of the additional linear layer. \thismodel(NL) outperforms \thismodel(NL-NR) in all datasets, further motivating the use of the activation function before the Feed-Forward Network. Please refer to the Appendix for more details. 
\section{Conclusion}\label{sec:conclusions}
In this paper we presented extensions of pretrained time series models to incorporate exogenous variables via modules that inject information from past and future covariates. We showcase our approach with \thismodel and show that the same approach can be applied to other pretrained models with minimal modifications, resulting in extensions like \timesfmx and \momentx. We study the cases where we finetune the entire model or we only fit the parameters of the covariate-related modules, and introduced a family of 32 synthetic datasets with covariates to evaluate the efficacy of forecasting models to integrate covariates. We test the adapters in our models \thismodel, \timesfmx, and \momentx and demonstrate in synthetic and real datasets that our approach enables pretrained models to capture covariate information and perform favorably when compared to other baselines.

\textbf{Limitations - Future Work}. 
A limitation of this work is that the adapter approach for integrating covariate data requires training of the adapter module, therefore losing the benefit of pretrained models to perform zero-shot inference (arguable in exchange for lower forecasting error). An interesting research direction is to integrate covariates into pretrained forecasting models in inference akin to in-context learning, using directly the covariate data to reduce the forecast error.

\typeout{}
\bibliography{main}
\bibliographystyle{abbrvnat}

\section*{Checklist}

 \begin{enumerate}

 \item For all models and algorithms presented, check if you include:
 \begin{enumerate}
   \item A clear description of the mathematical setting, assumptions, algorithm, and/or model. [Yes] $\rightarrow$ Refer to Section \ref{sec_proposed_method} and Section \ref{sec:experiments_and_results}.
   \item An analysis of the properties and complexity (time, space, sample size) of any algorithm. [Yes] $\rightarrow$ Refer to Section \ref{app:time_effiency}.
   \item (Optional) Anonymized source code, with specification of all dependencies, including external libraries. [No] $\rightarrow$ Released upon acceptance.
 \end{enumerate}

 \item For any theoretical claim, check if you include:
 \begin{enumerate}
   \item Statements of the full set of assumptions of all theoretical results. [Not Applicable]
   \item Complete proofs of all theoretical results. [Not Applicable]
   \item Clear explanations of any assumptions. [Not Applicable]     
 \end{enumerate}

 \item For all figures and tables that present empirical results, check if you include:
 \begin{enumerate}
   \item The code, data, and instructions needed to reproduce the main experimental results (either in the supplemental material or as a URL). [Yes]. 
   \item All the training details (e.g., data splits, hyperparameters, how they were chosen). [Yes] $\leftarrow$ Refer to Section \ref{app:hyperparameter_details}.
         \item A clear definition of the specific measure or statistics and error bars (e.g., with respect to the random seed after running experiments multiple times). [Yes] $\leftarrow$ Refer to Section \ref{app:metrics}.
         \item A description of the computing infrastructure used. (e.g., type of GPUs, internal cluster, or cloud provider). [Yes] $\rightarrow$ Refer to Section \ref{app:compute_resources}.
 \end{enumerate}

 \item If you are using existing assets (e.g., code, data, models) or curating/releasing new assets, check if you include:
 \begin{enumerate}
   \item Citations of the creator If your work uses existing assets. [Yes] $\rightarrow$ Refer to Section \ref{app:datasets_with_covariates}.
   \item The license information of the assets, if applicable. [Not Applicable]
   \item New assets either in the supplemental material or as a URL, if applicable. [Not Applicable]
   \item Information about consent from data providers/curators. [Not Applicable]
   \item Discussion of sensible content if applicable, e.g., personally identifiable information or offensive content. [Not Applicable]
 \end{enumerate}

 \item If you used crowdsourcing or conducted research with human subjects, check if you include:
 \begin{enumerate}
   \item The full text of instructions given to participants and screenshots. [Not Applicable]
   \item Descriptions of potential participant risks, with links to Institutional Review Board (IRB) approvals if applicable. [Not Applicable]
   \item The estimated hourly wage paid to participants and the total amount spent on participant compensation. [Not Applicable]
 \end{enumerate}

 \end{enumerate}

\appendix
\onecolumn
\newpage\clearpage
\section{Appendix Structure} \label{app:appendix_structure}

In this appendix, we specify details on different aspects of our work. We address the following items:
\begin{itemize}

    \item In Appendix \ref{app:hyperparameter_details}, we detail the hyperparameters used for the main method.
    \item In Appendix \ref{app:compute_resources}, we mention the computing resource used in the experiments.
    \item In Appendix \ref{app:chronos_variants}, we describe the Chronos variants.
    \item In Appendix \ref{app:synthetic-dataset-creation}, we give more information about the synthetic dataset creation.
    \item In Appendix \ref{app:datasets_with_covariates}, we detail information on the datasets with covariates.
    \item In Appendix \ref{app:baselines}, we specify the baselines configurations.
    \item In Appendix \ref{app:metrics}, we describe the metrics that we used in our experiments.
    \item In Appendix \ref{app:additional_results}, we present additional results:
    \begin{itemize}
        \item In Appendix \ref{app:time_effiency}, we compare the execution time and the number of parameters for different chronos variants.
        \item Section \ref{sec::appendix:full_set_of_evaluations}: full set of evaluations with \thismodel, \timesfmx, and \momentx
        \item Section \ref{sec::appendix:rq2}: evaluations on covariate injection on pretrained models with either past covariates, future covariates, or both.
        \item Section \ref{appendix:ablation_improvement_from_covariates}: ablations showing that  performance improvement is brought by covariates.
        \item Section \ref{appendix:ablation_larger_versions_of_chronos}: ablations showing the performance with larger version of Chronos.
        \item Section \ref{appendix:ablation_adapter_arch}: ablations showing the performance of different adapter architectures for covariates.
        \item Section \ref{sec::appendix:rq4}: Probabilistic forecasts visualizations.
    \end{itemize}
\end{itemize}

\section{Hyperparameter Details} \label{app:hyperparameter_details}
For all the experiments, we use the output dimension of the linear layers and the hidden dimension of the Feed Forward Networks is 256. We train the models for 5000 steps, and checkpointing the best observed model every 100 steps. The rest of the training setup such as the optimizer and learning rate scheduler were set following previous work~\cite{ansari2024Chronos}. For \timesfmx we used a hidden dimension of 256, learning rate 1.e-5, and cosine annealing scheduler as in~\citep{das2023decoder}. For \momentx we used a hidden dimension of 64, maximum learning 1.e-4 with one cycle learning rate schedule as in~\citep{goswami2024moment}. Remaining parameters of \timesfmx a \momentx were set up following previous works~\citep{das2023decoder, goswami2024moment}.

\section{Compute Resource Information} \label{app:compute_resources}
For the experiments, we use GPU Nvidia A10G. More expensive experiments used NVIDIA Tesla V100. For baselines we used cpu instances with 16 virtual cpus and 32 GiB of memory.

\section{Chronos Variants} \label{app:chronos_variants}

\subsection{Additional Covariate Modules: Hidden State Only and Covariates Only (Residual) Injection Block.}
To understand the impact of the proposed output and input injection blocks, we present the following variants.
First, we propose a hidden-state base adapter that does not take covariates into account and works as an alternative to full finetuning without covariates, defined as
\begin{equation}
    f_{\mathrm{HS}}(h_{\mathrm{\text{out}}}(\mathbf{z}_{1:t-1}))=  \mathrm{FFN} \left( \mathrm{ReLU}\left(h_{\mathrm{\text{out}}}(\mathbf{z}_{1:t-1})W^{(\mathrm{out})}_{{\mathrm{OIB}}}  \right) \right)
    \label{eq:hidden_state_block}
\end{equation}
and second, we present an adapter to incorporate covariates of the future by updating logits through a linear transformation. 
\begin{equation}
f_{\mathrm{RS}}\left(
\mathbf{z}_{1:t-1},
\mathbf{x}_{t}
\right) 
= 
h_{\mathrm{out}}(\mathbf{z}_{1:t-1})W_{\mathrm{out}}
+ 
\mathbf{x}_{t}W_{{\mathrm{OIB}}}^{({\mathrm{cov}})}
  \label{eq:residual_block}
\end{equation}

\subsection{Variants List}
Here we explain the baselines, for more information please refer to the section \ref{sec_proposed_method}. In the notation below, we refer to the IIB (Input Injection Block) as the network that receives the past covariates and updates the token embedding, while the OIB (Output Injection Block) indicates the network that receives the hidden state and the future covariates. 

\begin{itemize}
    \item \textsc{Chronos} is the model with the pretrained weights (not finetuned).
    \item \textsc{Chronos(FF)} is the pretrained model full-finetuned on the corresponding dataset.
    \item \textsc{ChronosX}, represented as well as \textsc{ChronosX(IIB+OIB)}, is the frozen pretrained model with the additional Input and Output Injection Blocks.
    \item \textsc{ChronosX(FF)}, represented as well as \textsc{ChronosX(FF+IIB+OIB)}, is the full-finetuned \textsc{Chronos} model with the additional Input and Output Injection Block.
    \item \textsc{ChronosX(FF+IIB)} is the full-finetuned \textsc{Chronos} model with Input Injection Block.
    \item \textsc{ChronosX(FF+OIB)} is the full-finetuned \textsc{Chronos} model with Output Injection Block.
    \item \textsc{ChronosX(OIB)} is the frozen pretrained model only with Output Injection Block.
    \item  \textsc{ChronosX(IIB)} is the frozen pretrained model only with Input Injection Block.
    \item  \textsc{ChronosX(HS)} is the frozen pretrained model only with hidden state, as in Equation \ref{eq:hidden_state_block}.
    \item  \textsc{ChronosX(RS)} is the frozen pretrained model only with modules accepting covariates, as in Equation \ref{eq:residual_block}.
\end{itemize}

\newpage\clearpage
\section{Synthetic Dataset Creation} \label{app:synthetic-dataset-creation}

We created a suite of 32 synthetic datasets comprising a 100 time series with length $L=1827$ and $30$ steps for the prediction length, assuming a "fictional" daily frequency. Each dataset is created by combining a \textbf{main signal} $\hat z_t$ with a \textbf{covariate} $x_t$ using a \textbf{combination operator} according to Eq. \ref{eq:combination_for_synthetics}.

\begin{equation}\label{eq:combination_for_synthetics}
z_t = \hat z_t \odot x_t, \quad  t=1,...,L
\end{equation}

\begin{figure}[H]
    \centering
    \newcolumntype{C}{ >{\centering\arraybackslash} m{0.12\linewidth} }

    \begin{tabular}{CCC | CCC}
      
         \includegraphics[trim={0.2cm 0.2cm 0.2cm 0.2cm}, width=\linewidth]{Figures/cipt_synthetic_generation/cipt_main_signal/single_spikes_add_target_before_covariates.pdf}
         &
         \includegraphics[trim={0.2cm 0.2cm 0.2cm 0.2cm}, width=\linewidth]{Figures/cipt_synthetic_generation/cipt_covariates/single_spikes_add_covariate.pdf}
         &
         \includegraphics[trim={0.2cm 0.2cm 0.2cm 0.2cm}, width=\linewidth]{Figures/cipt_synthetic_generation/cipt_result_signal_add/single_spikes_add_target_after_covariates_with_add.pdf}
        & 
        \includegraphics[trim={0.2cm 0.2cm 0.2cm 0.2cm}, width=\linewidth]{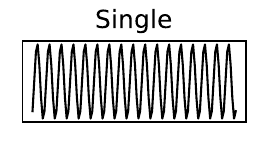}
         &
         \includegraphics[trim={0.2cm 0.2cm 0.2cm 0.2cm}, width=\linewidth]{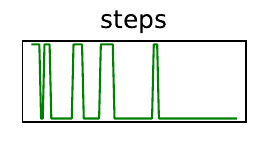}
         &
         \includegraphics[trim={0.2cm 0.2cm 0.2cm 0.2cm}, width=\linewidth]{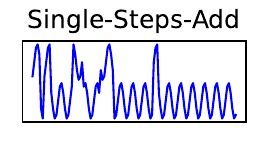} \\

         \includegraphics[trim={0.2cm 0.2cm 0.2cm 0.2cm}, width=\linewidth]{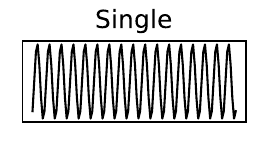}
         &
         \includegraphics[trim={0.2cm 0.2cm 0.2cm 0.2cm}, width=\linewidth]{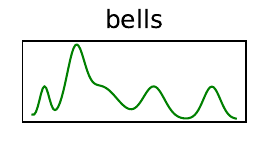}
         &
         \includegraphics[trim={0.2cm 0.2cm 0.2cm 0.2cm}, width=\linewidth]{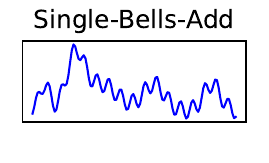}
        & 
        \includegraphics[trim={0.2cm 0.2cm 0.2cm 0.2cm}, width=\linewidth]{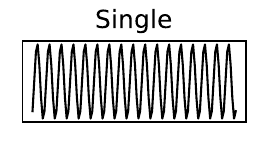}
         &
         \includegraphics[trim={0.2cm 0.2cm 0.2cm 0.2cm}, width=\linewidth]{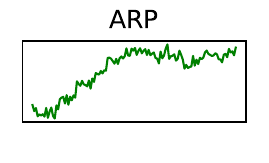}
         &
         \includegraphics[trim={0.2cm 0.2cm 0.2cm 0.2cm}, width=\linewidth]{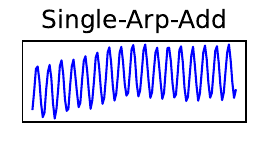} \\

  \includegraphics[trim={0.2cm 0.2cm 0.2cm 0.2cm}, width=\linewidth]{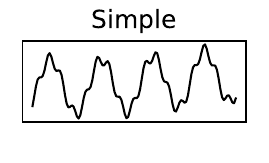}
         &
         \includegraphics[trim={0.2cm 0.2cm 0.2cm 0.2cm}, width=\linewidth]{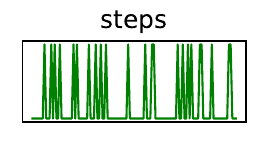}
         &
         \includegraphics[trim={0.2cm 0.2cm 0.2cm 0.2cm}, width=\linewidth]{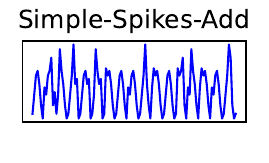}
        & 
        \includegraphics[trim={0.2cm 0.2cm 0.2cm 0.2cm}, width=\linewidth]{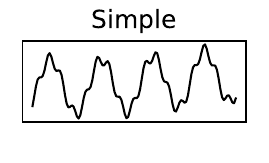}
         &
         \includegraphics[trim={0.2cm 0.2cm 0.2cm 0.2cm}, width=\linewidth]{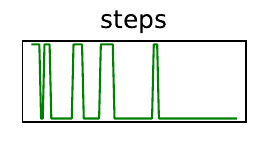}
         &
         \includegraphics[trim={0.2cm 0.2cm 0.2cm 0.2cm}, width=\linewidth]{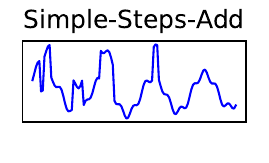} \\

         \includegraphics[trim={0.2cm 0.2cm 0.2cm 0.2cm}, width=\linewidth]{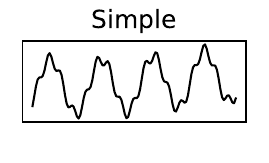}
         &
         \includegraphics[trim={0.2cm 0.2cm 0.2cm 0.2cm}, width=\linewidth]{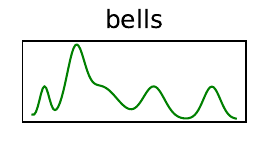}
         &
         \includegraphics[trim={0.2cm 0.2cm 0.2cm 0.2cm}, width=\linewidth]{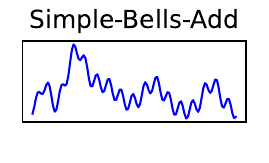}
        & 
        \includegraphics[trim={0.2cm 0.2cm 0.2cm 0.2cm}, width=\linewidth]{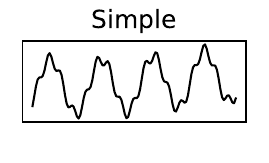}
         &
         \includegraphics[trim={0.2cm 0.2cm 0.2cm 0.2cm}, width=\linewidth]{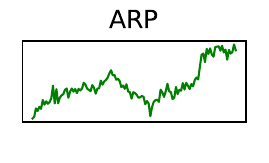}
         &
         \includegraphics[trim={0.2cm 0.2cm 0.2cm 0.2cm}, width=\linewidth]{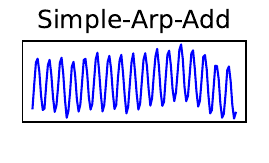} \\

        \includegraphics[trim={0.2cm 0.2cm 0.2cm 0.2cm}, width=\linewidth]{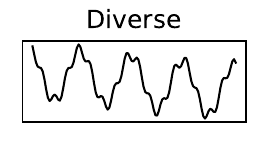}
         &
         \includegraphics[trim={0.2cm 0.2cm 0.2cm 0.2cm}, width=\linewidth]{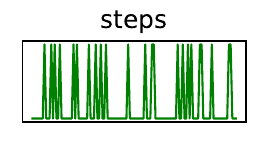}
         &
         \includegraphics[trim={0.2cm 0.2cm 0.2cm 0.2cm}, width=\linewidth]{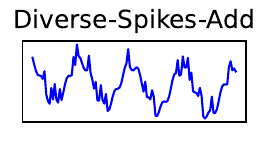}
        & 
        \includegraphics[trim={0.2cm 0.2cm 0.2cm 0.2cm}, width=\linewidth]{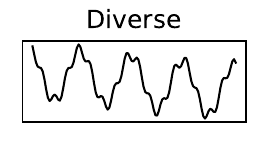}
         &
         \includegraphics[trim={0.2cm 0.2cm 0.2cm 0.2cm}, width=\linewidth]{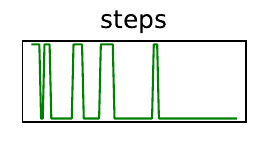}
         &
         \includegraphics[trim={0.2cm 0.2cm 0.2cm 0.2cm}, width=\linewidth]{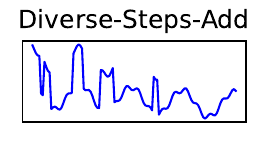} \\

         \includegraphics[trim={0.2cm 0.2cm 0.2cm 0.2cm}, width=\linewidth]{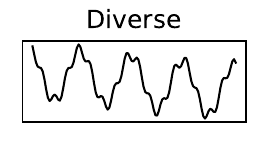}
         &
         \includegraphics[trim={0.2cm 0.2cm 0.2cm 0.2cm}, width=\linewidth]{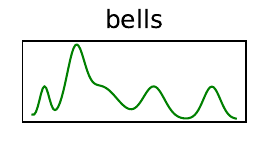}
         &
         \includegraphics[trim={0.2cm 0.2cm 0.2cm 0.2cm}, width=\linewidth]{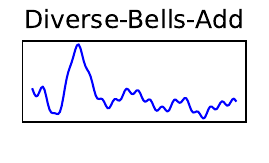}
        & 
        \includegraphics[trim={0.2cm 0.2cm 0.2cm 0.2cm}, width=\linewidth]{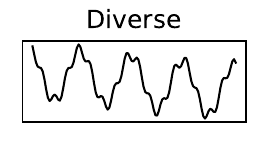}
         &
         \includegraphics[trim={0.2cm 0.2cm 0.2cm 0.2cm}, width=\linewidth]{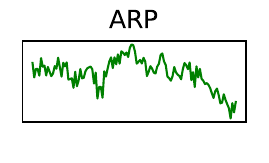}
         &
         \includegraphics[trim={0.2cm 0.2cm 0.2cm 0.2cm}, width=\linewidth]{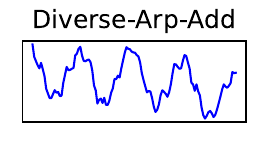} \\

         \includegraphics[trim={0.2cm 0.2cm 0.2cm 0.2cm}, width=\linewidth]{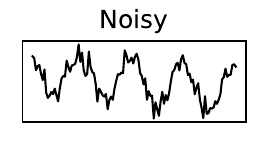}
         &
         \includegraphics[trim={0.2cm 0.2cm 0.2cm 0.2cm}, width=\linewidth]{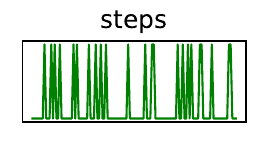}
         &
         \includegraphics[trim={0.2cm 0.2cm 0.2cm 0.2cm}, width=\linewidth]{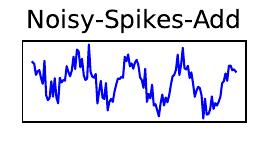}
        & 
        \includegraphics[trim={0.2cm 0.2cm 0.2cm 0.2cm}, width=\linewidth]{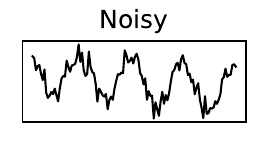}
         &
         \includegraphics[trim={0.2cm 0.2cm 0.2cm 0.2cm}, width=\linewidth]{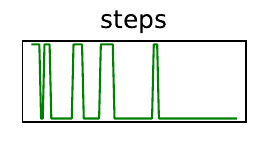}
         &
         \includegraphics[trim={0.2cm 0.2cm 0.2cm 0.2cm}, width=\linewidth]{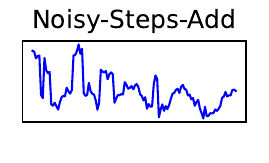} \\

         \includegraphics[trim={0.2cm 0.2cm 0.2cm 0.2cm}, width=\linewidth]{Figures/cipt_synthetic_generation/cipt_main_signal/noisy_bells_add_target_before_covariates.pdf}
         &
         \includegraphics[trim={0.2cm 0.2cm 0.2cm 0.2cm}, width=\linewidth]{Figures/cipt_synthetic_generation/cipt_covariates/noisy_bells_add_covariate.pdf}
         &
         \includegraphics[trim={0.2cm 0.2cm 0.2cm 0.2cm}, width=\linewidth]{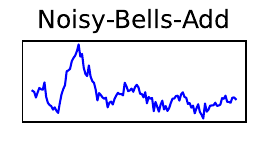}
        & 
        \includegraphics[trim={0.2cm 0.2cm 0.2cm 0.2cm}, width=\linewidth]{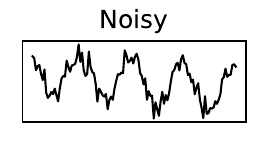}
         &
         \includegraphics[trim={0.2cm 0.2cm 0.2cm 0.2cm}, width=\linewidth]{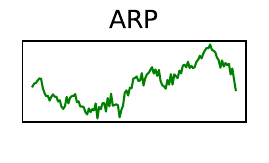}
         &
         \includegraphics[trim={0.2cm 0.2cm 0.2cm 0.2cm}, width=\linewidth]{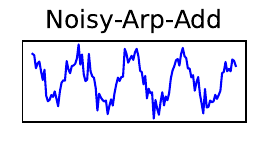}
    \end{tabular}
    \caption{\label{fig:add_synthetic_all_comb}This figure provides a single time series example from all possible combinations of \textbf{covariates} and \textbf{main signal} with the add $+$ \textbf{operator.}}

\end{figure}

\begin{figure}[H]
    \centering
    \newcolumntype{C}{ >{\centering\arraybackslash} m{0.12\linewidth} }

    \begin{tabular}{CCC | CCC}

         \includegraphics[trim={0.2cm 0.2cm 0.2cm 0.2cm}, width=\linewidth]{Figures/cipt_synthetic_generation/cipt_main_signal/single_spikes_add_target_before_covariates.pdf}
         &
         \includegraphics[trim={0.2cm 0.2cm 0.2cm 0.2cm}, width=\linewidth]{Figures/cipt_synthetic_generation/cipt_covariates/single_spikes_add_covariate.pdf}
         &
         \includegraphics[trim={0.2cm 0.2cm 0.2cm 0.2cm}, width=\linewidth]{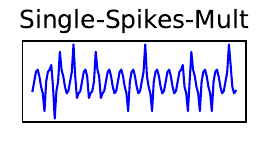}
        & 
        \includegraphics[trim={0.2cm 0.2cm 0.2cm 0.2cm}, width=\linewidth]{Figures/cipt_synthetic_generation/cipt_main_signal/single_steps_add_target_before_covariates.pdf}
         &
         \includegraphics[trim={0.2cm 0.2cm 0.2cm 0.2cm}, width=\linewidth]{Figures/cipt_synthetic_generation/cipt_covariates/single_steps_add_covariate.pdf}
         &
         \includegraphics[trim={0.2cm 0.2cm 0.2cm 0.2cm}, width=\linewidth]{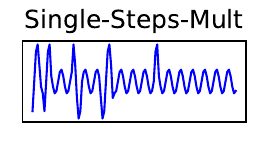} \\

         \includegraphics[trim={0.2cm 0.2cm 0.2cm 0.2cm}, width=\linewidth]{Figures/cipt_synthetic_generation/cipt_main_signal/single_bells_add_target_before_covariates.pdf}
         &
         \includegraphics[trim={0.2cm 0.2cm 0.2cm 0.2cm}, width=\linewidth]{Figures/cipt_synthetic_generation/cipt_covariates/single_bells_add_covariate.pdf}
         &
         \includegraphics[trim={0.2cm 0.2cm 0.2cm 0.2cm}, width=\linewidth]{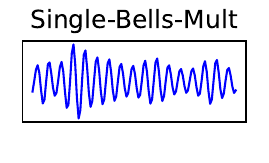}
        & 
        \includegraphics[trim={0.2cm 0.2cm 0.2cm 0.2cm}, width=\linewidth]{Figures/cipt_synthetic_generation/cipt_main_signal/single_arp_add_target_before_covariates.pdf}
         &
         \includegraphics[trim={0.2cm 0.2cm 0.2cm 0.2cm}, width=\linewidth]{Figures/cipt_synthetic_generation/cipt_covariates/single_arp_add_covariate.pdf}
         &
         \includegraphics[trim={0.2cm 0.2cm 0.2cm 0.2cm}, width=\linewidth]{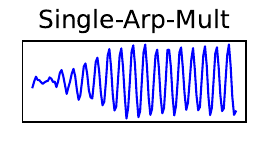} \\

        \includegraphics[trim={0.2cm 0.2cm 0.2cm 0.2cm}, width=\linewidth]{Figures/cipt_synthetic_generation/cipt_main_signal/simple_spikes_add_target_before_covariates.pdf}
         &
         \includegraphics[trim={0.2cm 0.2cm 0.2cm 0.2cm}, width=\linewidth]{Figures/cipt_synthetic_generation/cipt_covariates/simple_spikes_add_covariate.pdf}
         &
         \includegraphics[trim={0.2cm 0.2cm 0.2cm 0.2cm}, width=\linewidth]{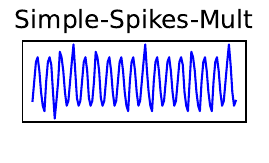}
        & 
        \includegraphics[trim={0.2cm 0.2cm 0.2cm 0.2cm}, width=\linewidth]{Figures/cipt_synthetic_generation/cipt_main_signal/simple_steps_add_target_before_covariates.pdf}
         &
         \includegraphics[trim={0.2cm 0.2cm 0.2cm 0.2cm}, width=\linewidth]{Figures/cipt_synthetic_generation/cipt_covariates/simple_steps_add_covariate.pdf}
         &
         \includegraphics[trim={0.2cm 0.2cm 0.2cm 0.2cm}, width=\linewidth]{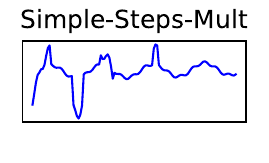} \\

         \includegraphics[trim={0.2cm 0.2cm 0.2cm 0.2cm}, width=\linewidth]{Figures/cipt_synthetic_generation/cipt_main_signal/simple_bells_add_target_before_covariates.pdf}
         &
         \includegraphics[trim={0.2cm 0.2cm 0.2cm 0.2cm}, width=\linewidth]{Figures/cipt_synthetic_generation/cipt_covariates/simple_bells_add_covariate.pdf}
         &
         \includegraphics[trim={0.2cm 0.2cm 0.2cm 0.2cm}, width=\linewidth]{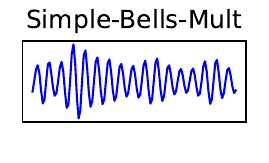}
        & 
        \includegraphics[trim={0.2cm 0.2cm 0.2cm 0.2cm}, width=\linewidth]{Figures/cipt_synthetic_generation/cipt_main_signal/simple_arp_add_target_before_covariates.pdf}
         &
         \includegraphics[trim={0.2cm 0.2cm 0.2cm 0.2cm}, width=\linewidth]{Figures/cipt_synthetic_generation/cipt_covariates/simple_arp_add_covariate.pdf}
         &
         \includegraphics[trim={0.2cm 0.2cm 0.2cm 0.2cm}, width=\linewidth]{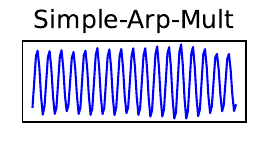} \\

        \includegraphics[trim={0.2cm 0.2cm 0.2cm 0.2cm}, width=\linewidth]{Figures/cipt_synthetic_generation/cipt_main_signal/diverse_spikes_add_target_before_covariates.pdf}
         &
         \includegraphics[trim={0.2cm 0.2cm 0.2cm 0.2cm}, width=\linewidth]{Figures/cipt_synthetic_generation/cipt_covariates/diverse_spikes_add_covariate.pdf}
         &
         \includegraphics[trim={0.2cm 0.2cm 0.2cm 0.2cm}, width=\linewidth]{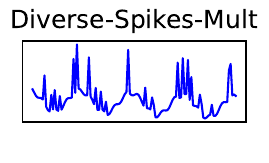}
        & 
        \includegraphics[trim={0.2cm 0.2cm 0.2cm 0.2cm}, width=\linewidth]{Figures/cipt_synthetic_generation/cipt_main_signal/diverse_steps_add_target_before_covariates.pdf}
         &
         \includegraphics[trim={0.2cm 0.2cm 0.2cm 0.2cm}, width=\linewidth]{Figures/cipt_synthetic_generation/cipt_covariates/diverse_steps_add_covariate.pdf}
         &
         \includegraphics[trim={0.2cm 0.2cm 0.2cm 0.2cm}, width=\linewidth]{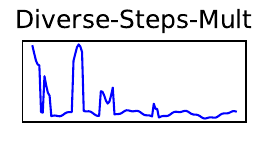} \\

         \includegraphics[trim={0.2cm 0.2cm 0.2cm 0.2cm}, width=\linewidth]{Figures/cipt_synthetic_generation/cipt_main_signal/diverse_bells_add_target_before_covariates.pdf}
         &
         \includegraphics[trim={0.2cm 0.2cm 0.2cm 0.2cm}, width=\linewidth]{Figures/cipt_synthetic_generation/cipt_covariates/diverse_bells_add_covariate.pdf}
         &
         \includegraphics[trim={0.2cm 0.2cm 0.2cm 0.2cm}, width=\linewidth]{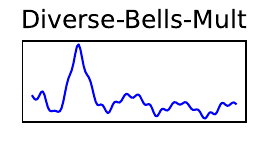}
        & 
        \includegraphics[trim={0.2cm 0.2cm 0.2cm 0.2cm}, width=\linewidth]{Figures/cipt_synthetic_generation/cipt_main_signal/diverse_arp_add_target_before_covariates.pdf}
         &
         \includegraphics[trim={0.2cm 0.2cm 0.2cm 0.2cm}, width=\linewidth]{Figures/cipt_synthetic_generation/cipt_covariates/diverse_arp_add_covariate.pdf}
         &
         \includegraphics[trim={0.2cm 0.2cm 0.2cm 0.2cm}, width=\linewidth]{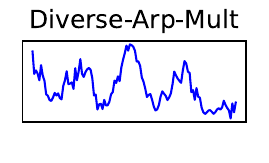} \\

         \includegraphics[trim={0.2cm 0.2cm 0.2cm 0.2cm}, width=\linewidth]{Figures/cipt_synthetic_generation/cipt_main_signal/noisy_spikes_add_target_before_covariates.pdf}
         &
         \includegraphics[trim={0.2cm 0.2cm 0.2cm 0.2cm}, width=\linewidth]{Figures/cipt_synthetic_generation/cipt_covariates/noisy_spikes_add_covariate.pdf}
         &
         \includegraphics[trim={0.2cm 0.2cm 0.2cm 0.2cm}, width=\linewidth]{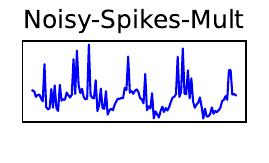}
        & 
        \includegraphics[trim={0.2cm 0.2cm 0.2cm 0.2cm}, width=\linewidth]{Figures/cipt_synthetic_generation/cipt_main_signal/noisy_steps_add_target_before_covariates.pdf}
         &
         \includegraphics[trim={0.2cm 0.2cm 0.2cm 0.2cm}, width=\linewidth]{Figures/cipt_synthetic_generation/cipt_covariates/noisy_steps_add_covariate.pdf}
         &
         \includegraphics[trim={0.2cm 0.2cm 0.2cm 0.2cm}, width=\linewidth]{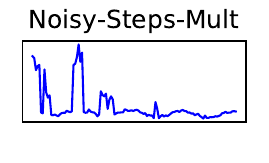} \\

         \includegraphics[trim={0.2cm 0.2cm 0.2cm 0.2cm}, width=\linewidth]{Figures/cipt_synthetic_generation/cipt_main_signal/noisy_bells_add_target_before_covariates.pdf}
         &
         \includegraphics[trim={0.2cm 0.2cm 0.2cm 0.2cm}, width=\linewidth]{Figures/cipt_synthetic_generation/cipt_covariates/noisy_bells_add_covariate.pdf}
         &
         \includegraphics[trim={0.2cm 0.2cm 0.2cm 0.2cm}, width=\linewidth]{Figures/cipt_synthetic_generation/cipt_result_signal_mult/noisy_bells_add_target_after_covariates_with_mult.pdf}
        & 
        \includegraphics[trim={0.2cm 0.2cm 0.2cm 0.2cm}, width=\linewidth]{Figures/cipt_synthetic_generation/cipt_main_signal/noisy_arp_add_target_before_covariates.pdf}
         &
         \includegraphics[trim={0.2cm 0.2cm 0.2cm 0.2cm}, width=\linewidth]{Figures/cipt_synthetic_generation/cipt_covariates/noisy_arp_add_covariate.pdf}
         &
         \includegraphics[trim={0.2cm 0.2cm 0.2cm 0.2cm}, width=\linewidth]{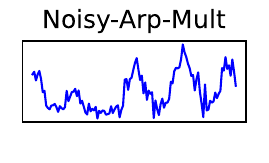}
    \end{tabular}
    \caption{\label{fig:mult_synthetic_all_comb}This figure provides a single time series example from all possible combinations of \textbf{covariates} and \textbf{main signal} with the mult $\times$ \textbf{operator.}}

\end{figure}

Once a combination is chosen, we randomly sample the parameters for the main signal \cref{app:main_signal_generation} and the covariates \cref{app:covariates} to get 100 different time series. All possible combinations are illustrated in \cref{fig:add_synthetic_all_comb} and \cref{fig:mult_synthetic_all_comb}.

\subsection{Main signal generation}\label{app:main_signal_generation}

The main signal to be affected by the covariates can be generated in different ways. We describe four variants in an increasing order of complexity by changing the set of possible parameters in Eq. \ref{equation:main_signal_shape}.
\begin{equation}\label{equation:main_signal_shape}
\hat z_t = \sum_{i=1}^3 a_i \sin \left(f_i t + \phi_i \right)
                                + b_1 t  + b_2
                                + \epsilon
\end{equation}

\begin{itemize}[leftmargin=15pt]
    \item \textbf{Single Sinuoid}: This represents the simplest time series, keeping the same amplitude, phase and a weekly frequency among time series. 
    
    $$\hat z_t^ {\text{(single)}} = \sin \left(2\pi \frac{t}{7} \right)$$
    
    \item \textbf{Simple Sinusoids}: Every time series in the dataset is the superposition of three sinusoids with different frequencies, simulating weekly, monthly and yearly seasonal patterns. For each time series, a different amplitude is sampled from $a_i \sim U(1,5)$.
    $$\hat z_t^ {\text{(sinusoids)}} = a_1 \sin \left(2\pi \frac{t}{7} \right) +
     a_2 \sin \left(2\pi \frac{t}{30} \right) +
     a_3 \sin \left(2\pi \frac{t}{365} \right)$$ 

    \item \textbf{Diverse Sinusoids}: Every time series is similar to \textit{simple sinusoids}, but we sample a different phase for every sinusiodal component with a random amplitude $a_i \sim U(1,5)$ and a random phase $\phi_i \sim U(-\pi,\pi)$. A trend component is also added with random coefficient $b_1, b_2 \sim  U(-1,1)$.
    $$\hat z_t^ {\text{(diverse)}} = a_1 \sin \left(2\pi \frac{t}{7} + \phi_1 \right) +
     a_2 \sin \left(2\pi \frac{t}{30} + \phi_2 \right) +
     a_3 \sin \left(2\pi \frac{t}{365} + \phi_3 \right) + b_1 \frac{t}{365}  + b_2$$

    \item \textbf{Noisy Sinusoids}. The Noisy Sinusoid are \textit{Diverse sinusoids} with noise added at every step. The variance is chosen to be proportional to the scale $s=\frac{1}{L}\sum^L_t |\hat z_t|$ of the main diverse sinusoid $\epsilon \sim \mathcal{N}\left(0,\frac{s}{4} \right)$.

    $$\hat z_t^ {\text{(noisy)}}  = \hat z_t^ {\text{(diverse)}} + \epsilon $$
\end{itemize}

\subsection{Covariate}\label{app:covariates}
Similarly to the main signal, the parameters of the covariates are randomly chosen depending on the types and the scale of the signal.

\begin{itemize}[leftmargin=15pt]
    \item \textbf{Spikes} are used to represent short time events by considering large value for single "active" time steps as in Eq. \ref{eq:spikes}. The $N=500$ times steps $\mathbb{T} = \{t_1, \ldots, t_N \}$ are uniformly sampled without replacement.
    The strength of all the spikes is randomly sampled in an interval limited to $5$ times the scale $\gamma \sim U(1,5s)$.
    
    \begin{equation} \label{eq:spikes}
         x_t = \begin{cases}
        \gamma & \text{if } t \in \mathbb{T}_N \\
        1 &  \text{otherwise}
    \end{cases} 
     \end{equation}

    \item \textbf{Steps} are used to study the influence of sudden but sustained events. Steps covariates share the same formula as the spikes in Eq. \ref{eq:steps} but the generation of the "active" index is different.  We consider $N=125$ non-overlapping intervals $T_i = \{t_i, t_{i+1}, \ldots, t_{i}+\delta_i\}$ by uniformly sampling the starting steps $t_i \sim U(0,L)$ and the associated duration $\delta_i \sim U(1,30)$ with $T_i \cap T_j = \emptyset$. The resulting signal is then composed of a union of the different intervals : $\mathbb{T} = \cup_i^N T_i$. All steps have the save amplitude $\gamma$, randomly sampled in the same manner as the spikes.
    
    \begin{equation} \label{eq:steps}
         x_t = \begin{cases}
        \gamma & \text{if } t \in \mathbb{T} \\
        1 & \text{if } \text{otherwise}
    \end{cases} 
     \end{equation}

    \item \textbf{Bells} are considered to represent trends with smoother changes. In this case, the signal is a mixture of gaussian bells as in Eq. \ref{eq:bells}. We consider mixtures of $N = 125$ bells with randomly sampled mean $\hat{\mu}_i \sim U(0,L)$ and randomly sampled standard deviation $\sigma_i \sim U(1, 15)$. The scale of the bells is randomly sampled in an interval limited to $5$ times the scale $\gamma \sim U(1,5s)$.

    \begin{equation}\label{eq:bells}
    x_{\mathrm{t}}=\gamma \sum_i^N \exp \left(\frac{-\left\|t-\hat{\mu}_i\right\|^2}{\sigma_i^2}\right)
    \end{equation}
    
    \item \textbf{ARP} is created following Eq. \ref{eq:arp}. Only second order auto-regressive processes are created with coefficients $a_1\sim U(0,1)$ and $a_2=1-a_1$,
    and noise amplitude equal to one. The scale of the final covariates is randomly sampled in an interval limited to $5$ times the scale $\gamma \sim U(1,5s)$.

    \begin{equation}\label{eq:arp}
    x_t=a_1 \cdot x_{t-1} +a_2 \cdot x_{t-2}+\epsilon    
    \end{equation}
    
\end{itemize}

\section{Datasets With Covariates}\label{app:datasets_with_covariates}
In this section we present a summary of the datasets with covariates used for evaluations.
\begin{table}[!ht]
\caption{Real Datasets Descriptors.}
    \centering
\label{tab:real_datasets_with_covariates}
\rowcolors{2}{gray!25}{white}
\renewcommand\theadfont{}
\resizebox{\textwidth}{!}{
    \begin{tabular}{ccccccc}
    \hline
     \toprule
        \textbf{Dataset Name} & \textbf{Num. Series} & \textbf{Num. Covariates} & \textbf{Freq.}  & \textbf{Prediction Length} & \textbf{Source}\\ \midrule
        ETT (15 Min.)    & 2 & 5 & 15min & 24 & \citep{haoyietal-informer-2021}\\
        ETT (Hourly)     & 2 & 5 & 1H & 24 & \citep{haoyietal-informer-2021} \\ 
        M5               & 30490 & 12 & 1D & 28 & \citep{makridakis2022m5}\\ 
        Electricity-BE   & 1 & 2 & 1H & 24 & \citep{LAGO2021116983, olivares2023neural}\\ 
        Electricity-DE   & 1 & 2 & 1H & 24 & \citep{LAGO2021116983, olivares2023neural}\\ 
        Electricity-FR   & 1 & 2 & 1H & 24 & \citep{LAGO2021116983, olivares2023neural}\\ 
        Electricity-NP   & 1 & 2 & 1H & 24 & \citep{LAGO2021116983, olivares2023neural}\\ 
        Electricity-PJM  & 1 & 2 & 1H & 24 & \citep{LAGO2021116983, olivares2023neural}\\ 
        BDG-2 Hog        & 24 & 5 & 1H & 24 & \citep{woo2024unified, wang2023benchmarks}\\ 
        BDG-2 Bull       & 41 & 3 & 1H & 24 & \citep{woo2024unified, wang2023benchmarks}\\ 
        BDG-2 Cockatoo   & 1 & 5 & 1H & 24 & \citep{woo2024unified, wang2023benchmarks}\\ 
        Covid19Energy    & 1 & 6 & 1H & 24 & \citep{woo2024unified, wang2023benchmarks}\\ 
        GEF12            & 20 & 1 & 1H & 24 & \citep{woo2024unified, wang2023benchmarks}\\ 
        GEF14            & 1 & 1 & 1H & 24 & \citep{woo2024unified, wang2023benchmarks}\\ 
        GEF17            & 8 & 1 & 1H & 24 & \citep{woo2024unified, wang2023benchmarks}\\ 
        PDB              & 1 & 1 & 1H & 24 & \citep{woo2024unified, wang2023benchmarks}\\ 
        Spanish          & 1 & 20 & 1H & 24 & \citep{woo2024unified, wang2023benchmarks}\\ 
        Rideshare        & 156 & 8 & 1H & 24 & \citep{godahewa2021monash}\\ 
        \bottomrule
    \end{tabular}
    }
\end{table}
\newpage
\section{Baselines}
\label{app:baselines}

For task-specific deep learning architectures DeepAR~\citep{salinas2020deepar}, PatchTST \citep{Nie2023PatchTST}, TFT~\citep{lim2021temporal} we based evaluations on  implementations in GluonTS~\citep{alexandrov2020gluonts}.
However, NBEATSx-I, NBEATSx-G~\citep{olivares2023neural} and NHiTS~\citep{challu2023nhits} experiments were based on implementations in the NeuralForecast \citep{olivares2022library_neuralforecast} library.
The original PatchTST model does not naturally accept additional time features as input, so we extend it in the following way. Similar to target time series, covariates are converted to patches and appended to the target patches. Since PatchTST is encoder only model, we combine past and future covariates (if available) and (left) shift them by prediction length so that the covariates are aligned with the target input. We call this version PatchTSTx.

The baselines DeepAR, PatchTST, TFT, NBEATSx-I, NBEATSx-G, , and N-HiTS were trained and evaluated three times and their performance averaged in order to account for high variance inherent in their optimization. 

Statistical baselines SeasonalNaive was used with their default hyperparameters in StatsForecast~\citep{garza2022statsforecast}, but with season length implied by their frequencies. 
For example, daily frequency data had season length set to 7, hourly data 24, etc. 
For this heuristic, we used the helper function {\tt get\_seasonality} from GluonTS.

Default hyperparameter configurations provided in baseline implementations were kept as is, and no dataset specific or global hyperparameter tuning was performed.
GluonTS-based implementations were optimized with a batch size of 128.

\begin{table}[t]
\centering
\caption{Baseline models and hyperparameter choices. Hyperparameters not specified are set to defaults in their respective implementations. $C$ stands for context length, $d_h$ for hidden layer dimension, $n_L$ for number of layers, $n_H$ for number of heads, and $\eta$ for learning rate.}
\label{tab:all-baselines}
\rowcolors{2}{gray!25}{white}
\renewcommand\theadfont{}
\resizebox{\textwidth}{!}{%

\begin{tabular}{l l l l p{10cm}}
\toprule
\textbf{Model} & \textbf{Model Type} & \textbf{Implementation} & \textbf{Probabilistic} & \textbf{Hyperparameters} \\
\midrule
SeasonalNaive  & Local               & StatsForecast           & Yes                    & N/A                            \\
AutoARIMA      & Local               & StatsForecast           & Yes                    & $C = 1000$               \\
DeepAR         & Task-specific       & GluonTS                 & Yes                    & $d_h = 40, n_{L} = 2$               \\
TFT            & Task-specific       & GluonTS                 & Yes                    & $d_h = 32, n_{H} = 4$               \\
PatchTST       & Task-specific       & GluonTS                 & Yes                    & Patch length: 16, Stride: 8, $d_h = 32, n_L = 2, n_{H} = 4$   \\
DLinear        & Task-specific       & GluonTS                 & Yes                    & Kernel size: 25, $d_h = 20$               \\
TiDE        & Task-specific       & GluonTS                 & Yes                    & encoder ($d_h=512$, $n_L=2$), decoder ($d_h=512$, $n_L=2$, output dimension: 8), temporal hidden dimension: 128, distribution hidden dimension: 512.               \\
N-BEATSx-I        & Task-specific       & NeuralForecast          & No                     & Input size multiplier: 5, scaler: robust\_scaler, stack\_types: Interpretable           \\
N-BEATSx-G        & Task-specific       & NeuralForecast          & No                     & Input size multiplier: 5, scaler: robust\_scaler, stack\_types: Generalized           \\
N-HiTS         & Task-specific       & NeuralForecast          & No                     & Input size multiplier: 5, scaler: robust\_scaler       \\
Moirai-1.0-R        & Pretrained           & Reference               & Yes                    & $C = 1024$, Patch length: auto, batch size:4\\
TimesFM-1.0-200m-pytorch        & Pretrained           & Reference               & No                    & $C = 512$\\
MOMENT        & Pretrained/Task-specific           & Reference               & No                    &Forecasting head training epochs: 100, $C = 512, \eta = 10^{-4}$\\
TTM-R2        & Pretrained           & Reference               & No                    &$C = 512$\\
TTM-CM        & Pretrained/Task-specific           & Reference               & No                    &$C = 512$\\
\bottomrule
\end{tabular}
}%

\end{table}

\section{Evaluation Metrics}\label{app:metrics}

For the evaluation metrics we follow the approach exposed in~\citep{ansari2021deep}.
Let $\{\rvx_i = [x_{i,1},\ldots,x_{i,C+H}]\}_{i=1}^N$ be a dataset with $N$ time series where $C$ and $H$ are the context length and prediction length, respectively. 

The mean absolute scaled error (MASE, \cite{hyndman2006another}) is defined as:
\[
    \mathrm{MASE}(\hat \rvx_i, \rvx_i) = \frac{C-S}{H}\frac{
        \sum_{t=C+1}^{C+H}|\hat x_{i,t} - x_{i,t}|
    }{
        \sum_{t=1}^{C-S}|x_{i,t} - x_{i,t+S}|
    },
\]
where $S$ is a seasonality parameter.

The Weighted Quantile Loss is defined as
\[
   \mathrm{WQL} = \frac{1}{K} \sum_{j=1}^K \mathrm{WQL}_{\alpha_j}.
\]
where $K$ is the number of quantiles, $\alpha$ is the set of quantile levels to evaluate, and 
\[
    \textrm{WQL}_\alpha = \frac{2\sum_{i,t} \mathrm{QL}_\alpha(q^{(\alpha)}_{i, t}, x_{i, t})}{\sum_{i,t} |x_{i, t}|}.
\]
where $\rvq^{(\alpha)}_i = [q^{(\alpha)}_{i,C+1},\ldots,q^{(\alpha)}_{i,C+H}]$ are the predicted quantiles at levels $\alpha\in(0, 1)$, and
\begin{equation}
    \textrm{QL}_{\alpha}(q, x) = \begin{cases}
        \alpha (x - q), & \mbox{if } x > q, \\
        (1-\alpha)(q - x), & \mbox{otherwise.}
    \end{cases}
    \label{eqn:quantile-loss}
\end{equation}

\section{Additional Results}
\label{app:additional_results}

\subsection{Time and Memory Efficiency of Adapters and Finetuning.}
\label{app:time_effiency}
\begin{figure}[h]
    \centering
    \begin{subfigure}{.25\linewidth}
    \includegraphics[width=\linewidth]{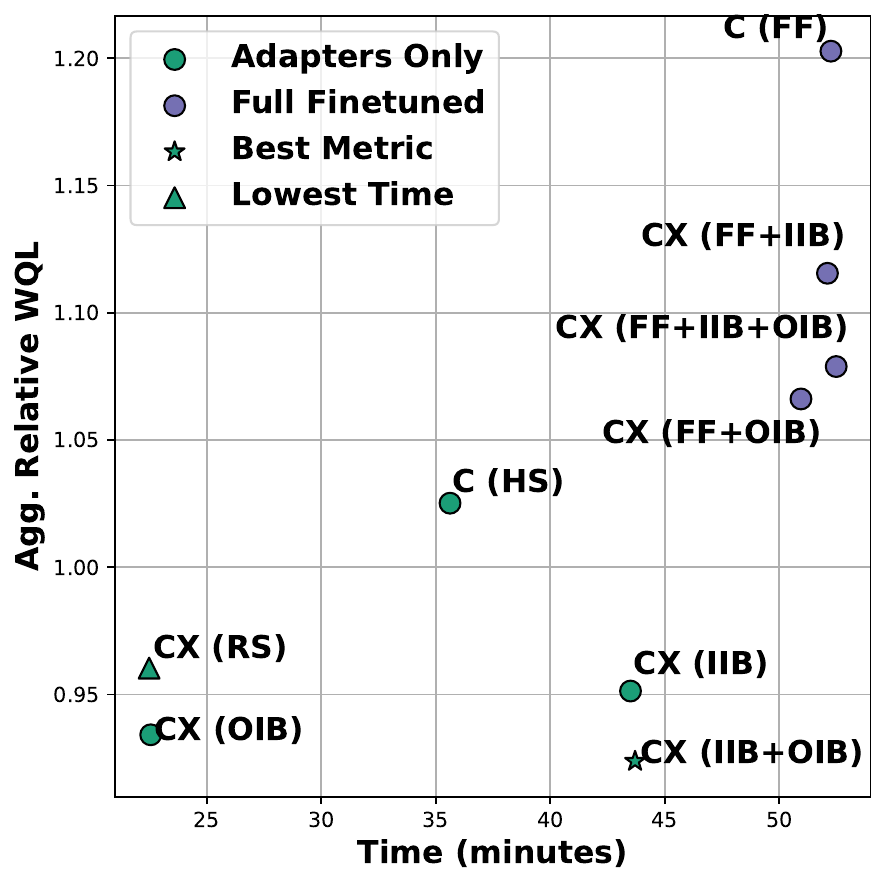}
    \caption{WQL vs Exec. Time}\label{fig:metrics_vs_time_WQL_cd_efficiency}
    \end{subfigure}
    \begin{subfigure}{.25\linewidth}
    \includegraphics[width=\linewidth]{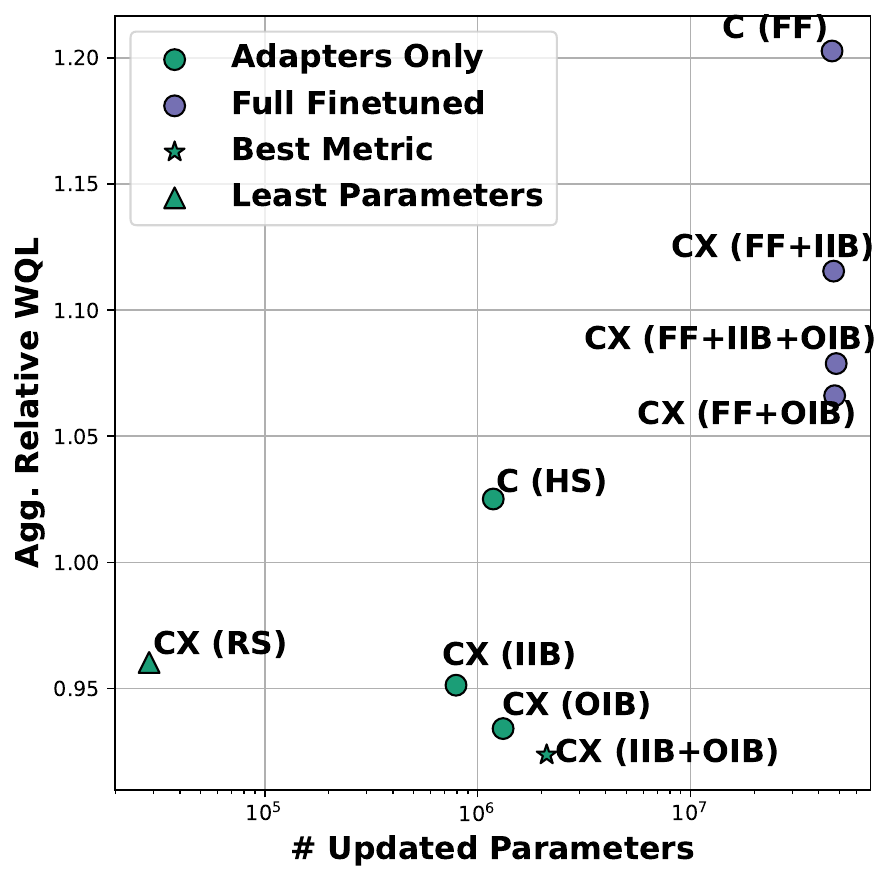}
    \caption{WQL vs $\#$ Parameters}\label{fig:metrics_vs_parameters_WQL_cd_efficiency}
    \end{subfigure}
    \label{fig:metrics_vs_efficiency}
  \caption{Average Time Execution and number of updated parameters of our proposed models against Aggregated Relative Weighted Quantile Loss on real datasets with covariates. 
  }
\end{figure}

We present an empirical analysis of the trade-offs between our models using only adapters and full finetuning against time execution and number of updated parameters. For a better visualization we represent \thismodel and \thatmodel with \textsc{CX} and \textsc{C}, respectively.
For this, we measure the average computational time consumed per run for training and testing across datasets and compared it to the aggregated WQL metric. As can be seen in Figure \ref{fig:metrics_vs_time_WQL_cd_efficiency}, the adapters are among the best performing and fastest variants. This is due to the decreased number of updated parameters as we report in Figure \ref{fig:metrics_vs_parameters_WQL_cd_efficiency}. Overall the most efficient variants in terms of time and number of updated hyperparameters is \textsc{ChronosX(RS)}, while the best performing and still efficient is \textsc{ChronosX}. We can see that full finetuning models have a larger amount of updated parameters and larger execution time than our approaches using only adapters.

\newpage\clearpage
\subsection{Full set of evaluations with \thismodel, \timesfmx, and \momentx}
\label{sec::appendix:full_set_of_evaluations}

\begin{figure}[t]
    \centering
    \begin{subfigure}{.48\linewidth}
    \includegraphics[width=1\linewidth]{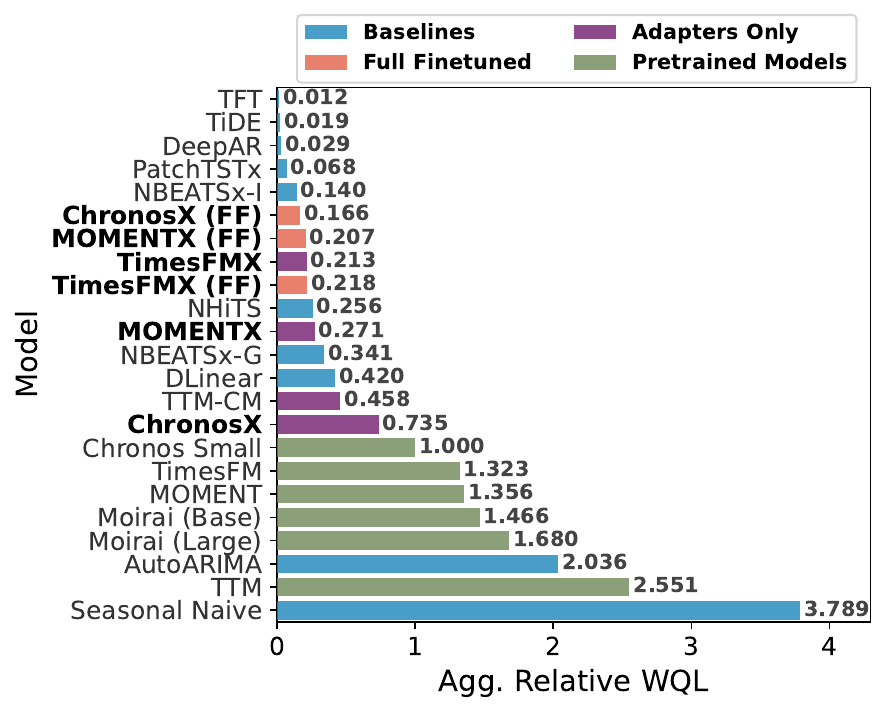}
    \caption{Agg. Rel. WQL on \textbf{Simple} synthetic datasets.}
    \label{fig:appendix:datasets_synthetic_simple_main}
    \end{subfigure}
    \quad
    \begin{subfigure}{.48\linewidth}
    \includegraphics[width=1\linewidth]{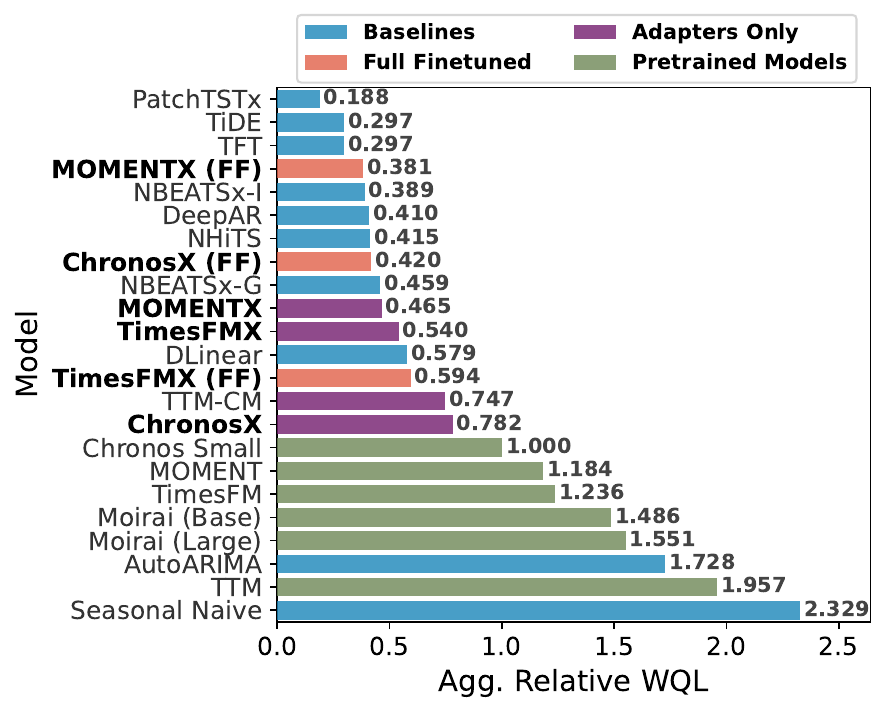}
    \caption{Agg. Rel. WQL on \textbf{Complex} synthetic datasets.}
    \label{fig:appendix:datasets_synthetic_complex_main}
    \end{subfigure}
    \\
    \begin{subfigure}{.48\linewidth}
    \includegraphics[width=1\linewidth]{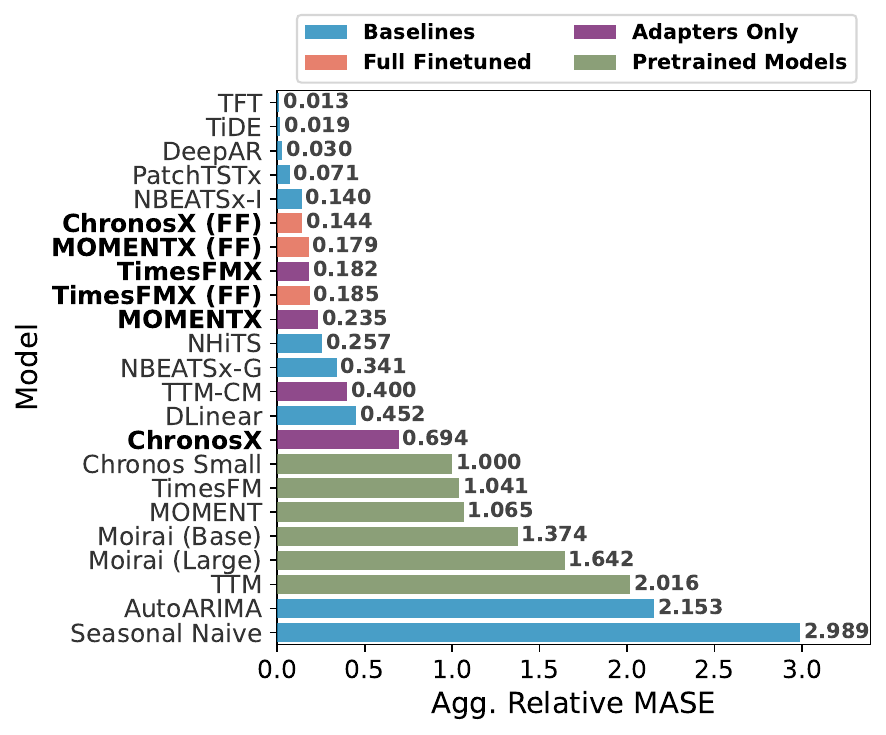}
    \caption{Agg. Rel. MASE on \textbf{Simple} synthetic datasets}
    \label{fig:appendix:datasets_synthetic_simple_main}
    \end{subfigure}
    \quad
    \begin{subfigure}{.48\linewidth}
    \includegraphics[width=1\linewidth]{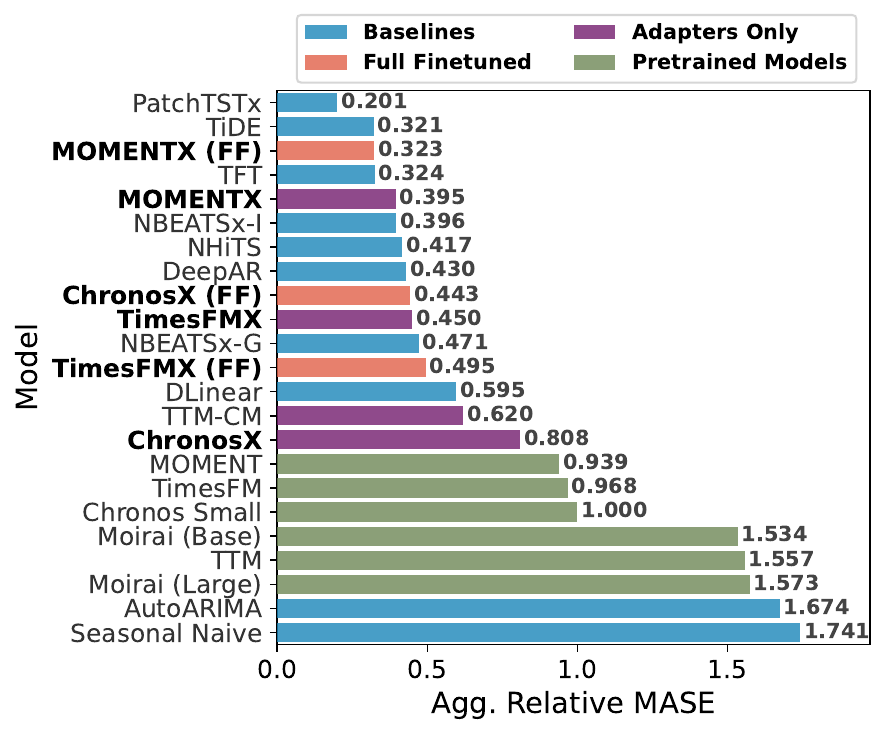}
    \caption{Agg. Rel. MASE on \textbf{Complex} synthetic datasets}
    \label{fig:appendix:datasets_synthetic_complex_main}
    \end{subfigure}
    \\
    \begin{subfigure}{.48\linewidth}
    \includegraphics[width=1\linewidth]{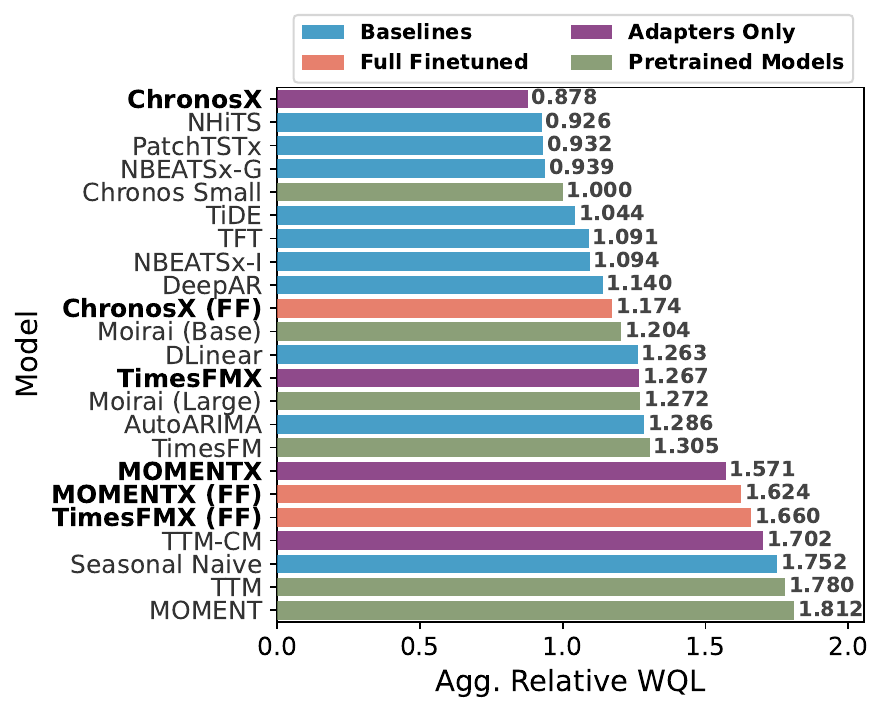}
    \caption{Agg. Rel. WQL on \textbf{real} datasets}
    \label{fig:appendix:datasets_synthetic_simple_main}
    \end{subfigure}
    \quad
    \begin{subfigure}{.48\linewidth}
    \includegraphics[width=1\linewidth]{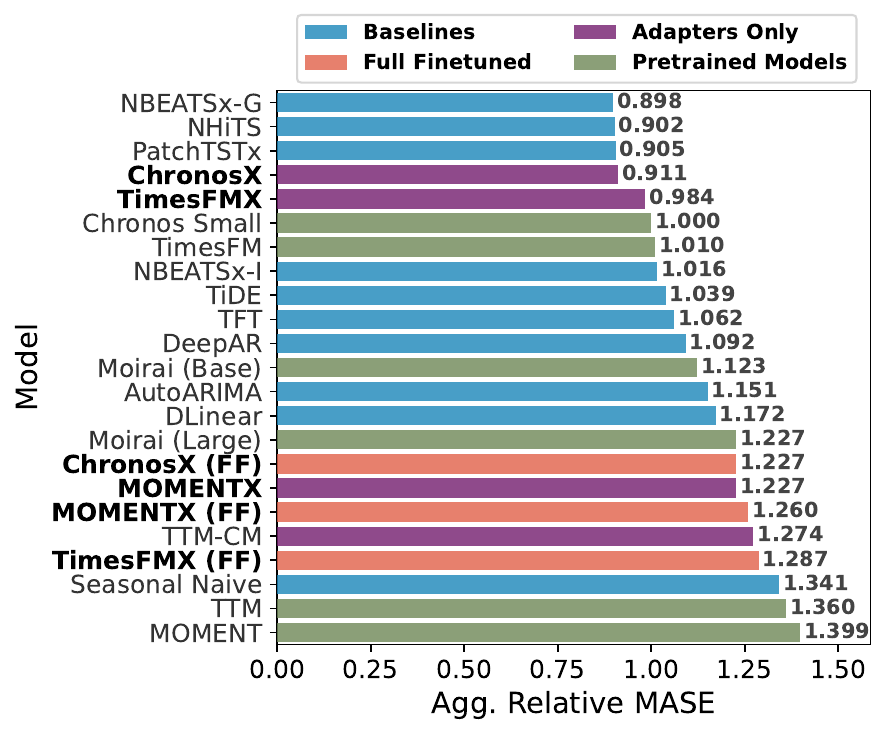}
    \caption{Agg. Rel. MASE on \textbf{real} datasets}
    \label{fig:appendix:datasets_synthetic_complex_main}
    \end{subfigure}
    \caption{Aggregated Relative WQL and MASE on simple, complex, and real datasets. Emphasis on comparisons of pretrained models with extensions to incorporate covariates.
    }
    \label{fig:appendix:ciptfinal_simple_synthetic_grouped_AggRel_ALL_v1}
\end{figure}

We present the following results:
\begin{itemize}[leftmargin=15pt]
    \item In Fig.~\ref{fig:appendix:ciptfinal_simple_synthetic_grouped_AggRel_ALL_v1} we present the Aggregated Relative WQL and MASE on simple, complex, and  real datasets,
    \item In~\Cref{tab:appendix:ciptfinal_simple_synthetic_grouped_wql_v1} and~\Cref{tab:appendix:ciptfinal_complex_synthetic_grouped_wql_v1} we present the  WQL on simple and complex datasets, respectively,
    \item In~\Cref{tab:appendix:ciptfinal_simple_synthetic_grouped_mase_v1} and~\Cref{tab:appendix:ciptfinal_complex_synthetic_grouped_mase_v1} we present the MASE on simple datasets, respectively.
    \item In~\Cref{tab:appendix:ciptfinal_real_grouped_mase_v1} and~\Cref{tab:appendix:ciptfinal_real_grouped_wql_v1} we present the MASE and WQL on real datasets, respectively.
\end{itemize}

\clearpage

\begin{table}[H]
    \centering
    \caption{WQL scores per model in 16 \textbf{simple} synthetic datasets. Models achieving the \underline{\textbf{first}}, \textbf{second}, and \underline{third} best scores have been highlighted. Scores reported are averaged over three random seeds.}
    \label{tab:appendix:ciptfinal_simple_synthetic_grouped_wql_v1}
    \rowcolors{2}{gray!25}{white}
    \renewcommand\theadfont{}
    \resizebox{\textwidth}{!}{%
\begin{tabular}{lccccccccccccccccccccccc}
\toprule
 & \multicolumn{3}{c}{Full Finetuned} & \multicolumn{4}{c}{Adapters Only} & \multicolumn{6}{c}{Pretrained Models} & \multicolumn{10}{c}{Baselines} \\
 \cmidrule(lr){2-4} \cmidrule(lr){5-8} \cmidrule(lr){9-14} \cmidrule(lr){15-24}
\rowcolor{white} & \rotatebox{45}{\thead{ChronosX (FF)}} & \rotatebox{45}{\thead{MOMENTX (FF)}} & \rotatebox{45}{\thead{TimesFMX (FF)}} & \rotatebox{45}{\thead{ChronosX}} & \rotatebox{45}{\thead{MOMENTX}} & \rotatebox{45}{\thead{TimesFMX}} & \rotatebox{45}{\thead{TTM-CM}} & \rotatebox{45}{\thead{Chronos Small}} & \rotatebox{45}{\thead{Moirai (Base)}} & \rotatebox{45}{\thead{Moirai (Large)}} & \rotatebox{45}{\thead{MOMENT}} & \rotatebox{45}{\thead{TimesFM}} & \rotatebox{45}{\thead{TTM}} & \rotatebox{45}{\thead{AutoARIMA}} & \rotatebox{45}{\thead{TFT}} & \rotatebox{45}{\thead{DeepAR}} & \rotatebox{45}{\thead{PatchTSTx}} & \rotatebox{45}{\thead{NBEATSx-I}} & \rotatebox{45}{\thead{NBEATSx-G}} & \rotatebox{45}{\thead{NHiTS}} & \rotatebox{45}{\thead{Seasonal Naive}} & \rotatebox{45}{\thead{TiDE}} & \rotatebox{45}{\thead{DLinear}} \\
\midrule
Simple ARP (Add) & 0.164 & 0.070 & 0.024 & 0.168 & 0.074 & 0.036 & 0.063 & 0.056 & 0.099 & 0.148 & 0.075 & 0.059 & 0.276 & 0.198 & 0.007 & 0.007 & \bftable \underline{0.002} & \underline{0.005} & 0.015 & 0.012 & 0.402 & \bftable 0.003 & 0.022 \\
Simple ARP (Mult.) & 0.042 & 0.039 & 0.038 & 0.136 & 0.047 & 0.026 & 0.044 & 0.037 & 0.114 & 0.153 & 0.058 & 0.048 & 0.374 & 0.313 & \bftable \underline{0.001} & 0.004 & \underline{0.003} & 0.012 & 0.018 & 0.014 & 0.572 & \bftable 0.002 & 0.030 \\
Simple Bells (Add) & 0.287 & 0.035 & 0.079 & 0.225 & 0.051 & 0.060 & 0.280 & 0.319 & 0.346 & 0.383 & 0.402 & 0.453 & 0.415 & 0.176 & \underline{0.005} & 0.056 & \bftable \underline{0.003} & 0.048 & 0.061 & 0.037 & 0.582 & \bftable 0.005 & 0.060 \\
\bottomrule
\end{tabular}

    }
\end{table}

\begin{table}[H]
    \centering
    \caption{WQL scores per model in 16 \textbf{complex} synthetic datasets. Models achieving the \underline{\textbf{first}}, \textbf{second}, and \underline{third} best scores have been highlighted. Scores reported are averaged over three random seeds.}
    \label{tab:appendix:ciptfinal_complex_synthetic_grouped_wql_v1}
    \rowcolors{2}{gray!25}{white}
    \renewcommand\theadfont{}
    \resizebox{\textwidth}{!}{%
    \begin{tabular}{lccccccccccccccccccccccc}
\toprule
 & \multicolumn{3}{c}{Full Finetuned} & \multicolumn{4}{c}{Adapters Only} & \multicolumn{6}{c}{Pretrained Models} & \multicolumn{10}{c}{Baselines} \\
 \cmidrule(lr){2-4} \cmidrule(lr){5-8} \cmidrule(lr){9-14} \cmidrule(lr){15-24}
\rowcolor{white} & \rotatebox{45}{\thead{ChronosX (FF)}} & \rotatebox{45}{\thead{MOMENTX (FF)}} & \rotatebox{45}{\thead{TimesFMX (FF)}} & \rotatebox{45}{\thead{ChronosX}} & \rotatebox{45}{\thead{MOMENTX}} & \rotatebox{45}{\thead{TimesFMX}} & \rotatebox{45}{\thead{TTM-CM}} & \rotatebox{45}{\thead{Chronos Small}} & \rotatebox{45}{\thead{Moirai (Base)}} & \rotatebox{45}{\thead{Moirai (Large)}} & \rotatebox{45}{\thead{MOMENT}} & \rotatebox{45}{\thead{TimesFM}} & \rotatebox{45}{\thead{TTM}} & \rotatebox{45}{\thead{AutoARIMA}} & \rotatebox{45}{\thead{TFT}} & \rotatebox{45}{\thead{DeepAR}} & \rotatebox{45}{\thead{PatchTSTx}} & \rotatebox{45}{\thead{NBEATSx-I}} & \rotatebox{45}{\thead{NBEATSx-G}} & \rotatebox{45}{\thead{NHiTS}} & \rotatebox{45}{\thead{Seasonal Naive}} & \rotatebox{45}{\thead{TiDE}} & \rotatebox{45}{\thead{DLinear}} \\
\midrule
Diverse ARP (Add) & 0.092 & 0.052 & 0.050 & 0.117 & 0.052 & 0.052 & 0.052 & 0.063 & 0.138 & 0.167 & 0.061 & 0.063 & 0.227 & 0.265 & 0.023 & 0.032 & \bftable \underline{0.007} & \bftable 0.009 & 0.019 & \underline{0.012} & 0.250 & 0.019 & 0.028 \\
Diverse ARP (Mult.) & 0.036 & 0.047 & 0.062 & 0.077 & 0.054 & 0.048 & 0.057 & 0.059 & 0.137 & 0.153 & 0.059 & 0.057 & 0.202 & 0.269 & 0.030 & 0.029 & \bftable \underline{0.016} & \bftable 0.024 & 0.028 & \underline{0.024} & 0.221 & 0.027 & 0.041 \\
Diverse Bells (Add) & 0.208 & 0.066 & 0.134 & 0.271 & 0.088 & 0.117 & 0.255 & 0.257 & 0.313 & 0.317 & 0.322 & 0.355 & 0.369 & 0.168 & \bftable 0.027 & 0.161 & \bftable \underline{0.011} & 1.162 & 0.105 & 0.124 & 0.464 & \underline{0.035} & 0.081 \\
\bottomrule
\end{tabular}

    }
\end{table}

\begin{table}[H]
    \centering
    \caption{MASE scores per model in 16 \textbf{simple} synthetic datasets. Models achieving the \underline{\textbf{first}}, \textbf{second}, and \underline{third} best scores have been highlighted. Scores reported are averaged over three random seeds.}
    \label{tab:appendix:ciptfinal_simple_synthetic_grouped_mase_v1}
    \rowcolors{2}{gray!25}{white}
    \renewcommand\theadfont{}
    \resizebox{\textwidth}{!}{%
    \begin{tabular}{lccccccccccccccccccccccc}
\toprule
 & \multicolumn{3}{c}{Full Finetuned} & \multicolumn{4}{c}{Adapters Only} & \multicolumn{6}{c}{Pretrained Models} & \multicolumn{10}{c}{Baselines} \\
 \cmidrule(lr){2-4} \cmidrule(lr){5-8} \cmidrule(lr){9-14} \cmidrule(lr){15-24}
\rowcolor{white} & \rotatebox{45}{\thead{ChronosX (FF)}} & \rotatebox{45}{\thead{MOMENTX (FF)}} & \rotatebox{45}{\thead{TimesFMX (FF)}} & \rotatebox{45}{\thead{ChronosX}} & \rotatebox{45}{\thead{MOMENTX}} & \rotatebox{45}{\thead{TimesFMX}} & \rotatebox{45}{\thead{TTM-CM}} & \rotatebox{45}{\thead{Chronos Small}} & \rotatebox{45}{\thead{Moirai (Base)}} & \rotatebox{45}{\thead{Moirai (Large)}} & \rotatebox{45}{\thead{MOMENT}} & \rotatebox{45}{\thead{TimesFM}} & \rotatebox{45}{\thead{TTM}} & \rotatebox{45}{\thead{AutoARIMA}} & \rotatebox{45}{\thead{TFT}} & \rotatebox{45}{\thead{DeepAR}} & \rotatebox{45}{\thead{PatchTSTx}} & \rotatebox{45}{\thead{NBEATSx-I}} & \rotatebox{45}{\thead{NBEATSx-G}} & \rotatebox{45}{\thead{NHiTS}} & \rotatebox{45}{\thead{Seasonal Naive}} & \rotatebox{45}{\thead{TiDE}} & \rotatebox{45}{\thead{DLinear}} \\
\midrule
Simple ARP (Add) & 0.947 & 0.305 & 0.108 & 0.951 & 0.328 & 0.163 & 0.283 & 0.322 & 0.571 & 0.849 & 0.328 & 0.264 & 1.228 & 1.103 & 0.039 & 0.034 & \bftable \underline{0.013} & \underline{0.029} & 0.086 & 0.067 & 1.788 & \bftable 0.019 & 0.122 \\
Simple ARP (Mult.) & 0.161 & 0.130 & 0.134 & 0.500 & 0.160 & 0.093 & 0.148 & 0.157 & 0.454 & 0.642 & 0.204 & 0.175 & 1.260 & 1.616 & \bftable \underline{0.006} & 0.020 & \underline{0.014} & 0.051 & 0.070 & 0.057 & 1.930 & \bftable 0.009 & 0.128 \\
Simple Bells (Add) & 3.203 & 0.310 & 0.656 & 2.459 & 0.439 & 0.499 & 2.976 & 3.483 & 3.689 & 4.069 & 3.408 & 3.858 & 3.520 & 1.993 & \bftable 0.055 & 0.613 & \bftable \underline{0.035} & 0.491 & 0.641 & 0.392 & 4.951 & \underline{0.059} & 0.682 \\
\bottomrule
\end{tabular}

    }
\end{table}

\begin{table}[H]
    \centering
    \caption{MASE scores per model in 16 \textbf{complex} synthetic datasets. Models achieving the \underline{\textbf{first}}, \textbf{second}, and \underline{third} best scores have been highlighted. Scores reported are averaged over three random seeds.}
    \label{tab:appendix:ciptfinal_complex_synthetic_grouped_mase_v1}
    \rowcolors{2}{gray!25}{white}
    \renewcommand\theadfont{}
    \resizebox{\textwidth}{!}{%
    \begin{tabular}{lccccccccccccccccccccccc}
\toprule
 & \multicolumn{3}{c}{Full Finetuned} & \multicolumn{4}{c}{Adapters Only} & \multicolumn{6}{c}{Pretrained Models} & \multicolumn{10}{c}{Baselines} \\
 \cmidrule(lr){2-4} \cmidrule(lr){5-8} \cmidrule(lr){9-14} \cmidrule(lr){15-24}
\rowcolor{white} & \rotatebox{45}{\thead{ChronosX (FF)}} & \rotatebox{45}{\thead{MOMENTX (FF)}} & \rotatebox{45}{\thead{TimesFMX (FF)}} & \rotatebox{45}{\thead{ChronosX}} & \rotatebox{45}{\thead{MOMENTX}} & \rotatebox{45}{\thead{TimesFMX}} & \rotatebox{45}{\thead{TTM-CM}} & \rotatebox{45}{\thead{Chronos Small}} & \rotatebox{45}{\thead{Moirai (Base)}} & \rotatebox{45}{\thead{Moirai (Large)}} & \rotatebox{45}{\thead{MOMENT}} & \rotatebox{45}{\thead{TimesFM}} & \rotatebox{45}{\thead{TTM}} & \rotatebox{45}{\thead{AutoARIMA}} & \rotatebox{45}{\thead{TFT}} & \rotatebox{45}{\thead{DeepAR}} & \rotatebox{45}{\thead{PatchTSTx}} & \rotatebox{45}{\thead{NBEATSx-I}} & \rotatebox{45}{\thead{NBEATSx-G}} & \rotatebox{45}{\thead{NHiTS}} & \rotatebox{45}{\thead{Seasonal Naive}} & \rotatebox{45}{\thead{TiDE}} & \rotatebox{45}{\thead{DLinear}} \\
\midrule
Diverse ARP (Add) & 1.004 & 0.547 & 0.509 & 1.282 & 0.551 & 0.511 & 0.520 & 0.768 & 1.848 & 2.056 & 0.608 & 0.613 & 2.130 & 2.801 & 0.317 & 0.419 & \bftable \underline{0.099} & \bftable 0.110 & 0.242 & \underline{0.148} & 1.947 & 0.270 & 0.318 \\
Diverse ARP (Mult.) & 0.500 & 0.515 & 0.659 & 0.987 & 0.579 & 0.501 & 0.598 & 0.772 & 1.791 & 1.928 & 0.648 & 0.571 & 1.973 & 3.091 & 0.404 & 0.370 & \bftable \underline{0.221} & \bftable 0.314 & 0.374 & \underline{0.323} & 1.921 & 0.360 & 0.533 \\
Diverse Bells (Add) & 3.392 & 0.834 & 1.660 & 4.066 & 1.105 & 1.510 & 3.557 & 3.975 & 4.713 & 4.634 & 3.871 & 4.417 & 4.529 & 2.234 & \bftable 0.446 & 2.513 & \bftable \underline{0.169} & 12.561 & 1.584 & 1.512 & 5.563 & \underline{0.586} & 1.249 \\
\bottomrule
\end{tabular}

    }
\end{table}

\begin{table}[H]
    \centering
    \caption{WQL scores per model in real datasets. Models achieving the \underline{\textbf{first}}, \textbf{second}, and \underline{third} best scores have been highlighted. Scores reported are averaged over three random seeds.}
    \label{tab:appendix:ciptfinal_real_grouped_mase_v1}
    \rowcolors{2}{gray!25}{white}
    \renewcommand\theadfont{}
    \resizebox{\textwidth}{!}{%
    \begin{tabular}{lccccccccccccccccccccccc}
\toprule
 & \multicolumn{3}{c}{Full Finetuned} & \multicolumn{4}{c}{Adapters Only} & \multicolumn{6}{c}{Pretrained Models} & \multicolumn{10}{c}{Baselines} \\
 \cmidrule(lr){2-4} \cmidrule(lr){5-8} \cmidrule(lr){9-14} \cmidrule(lr){15-24}
\rowcolor{white} & \rotatebox{45}{\thead{ChronosX (FF)}} & \rotatebox{45}{\thead{MOMENTX (FF)}} & \rotatebox{45}{\thead{TimesFMX (FF)}} & \rotatebox{45}{\thead{ChronosX}} & \rotatebox{45}{\thead{MOMENTX}} & \rotatebox{45}{\thead{TimesFMX}} & \rotatebox{45}{\thead{TTM-CM}} & \rotatebox{45}{\thead{Chronos Small}} & \rotatebox{45}{\thead{Moirai (Base)}} & \rotatebox{45}{\thead{Moirai (Large)}} & \rotatebox{45}{\thead{MOMENT}} & \rotatebox{45}{\thead{TimesFM}} & \rotatebox{45}{\thead{TTM}} & \rotatebox{45}{\thead{AutoARIMA}} & \rotatebox{45}{\thead{TFT}} & \rotatebox{45}{\thead{DeepAR}} & \rotatebox{45}{\thead{PatchTSTx}} & \rotatebox{45}{\thead{NBEATSx-G}} & \rotatebox{45}{\thead{NBEATSx-I}} & \rotatebox{45}{\thead{NHiTS}} & \rotatebox{45}{\thead{Seasonal Naive}} & \rotatebox{45}{\thead{TiDE}} & \rotatebox{45}{\thead{DLinear}} \\
\midrule
ETT (15 Min.) & 0.022 & 0.036 & 0.036 & \bftable 0.016 & 0.026 & 0.038 & 0.024 & 0.016 & 0.019 & 0.040 & 0.043 & 0.023 & 0.024 & 0.020 & 0.020 & 0.072 & 0.035 & \underline{0.016} & 0.027 & \bftable \underline{0.015} & 0.065 & 0.034 & 0.028 \\
ETT (Hourly) & \bftable 0.025 & 0.050 & 0.045 & \underline{0.025} & 0.050 & 0.049 & 0.041 & \bftable \underline{0.022} & 0.037 & 0.076 & 0.068 & 0.048 & 0.093 & 0.045 & 0.045 & 0.026 & 0.036 & 0.065 & 0.063 & 0.035 & 0.101 & 0.048 & 0.030 \\
M5 & \bftable \underline{0.538} & 0.750 & 0.763 & \underline{0.557} & 0.759 & 0.778 & 0.824 & \bftable 0.549 & 0.586 & 0.576 & 0.759 & 0.756 & 0.778 & - & 0.560 & 0.617 & 0.812 & 0.607 & 0.604 & 0.748 & 0.934 & 0.595 & 0.728 \\
\bottomrule
\end{tabular}

    }
\end{table}

\begin{table}[H]
    \centering
    \caption{MASE scores per model in real datasets. Models achieving the \underline{\textbf{first}}, \textbf{second}, and \underline{third} best scores have been highlighted. Scores reported are averaged over three random seeds.}
    \label{tab:appendix:ciptfinal_real_grouped_wql_v1}
    \rowcolors{2}{gray!25}{white}
    \renewcommand\theadfont{}
    \resizebox{\textwidth}{!}{%
    \begin{tabular}{lccccccccccccccccccccccc}
\toprule
 & \multicolumn{3}{c}{Full Finetuned} & \multicolumn{4}{c}{Adapters Only} & \multicolumn{6}{c}{Pretrained Models} & \multicolumn{10}{c}{Baselines} \\
 \cmidrule(lr){2-4} \cmidrule(lr){5-8} \cmidrule(lr){9-14} \cmidrule(lr){15-24}
\rowcolor{white} & \rotatebox{45}{\thead{ChronosX (FF)}} & \rotatebox{45}{\thead{MOMENTX (FF)}} & \rotatebox{45}{\thead{TimesFMX (FF)}} & \rotatebox{45}{\thead{ChronosX}} & \rotatebox{45}{\thead{MOMENTX}} & \rotatebox{45}{\thead{TimesFMX}} & \rotatebox{45}{\thead{TTM-CM}} & \rotatebox{45}{\thead{Chronos Small}} & \rotatebox{45}{\thead{Moirai (Base)}} & \rotatebox{45}{\thead{Moirai (Large)}} & \rotatebox{45}{\thead{MOMENT}} & \rotatebox{45}{\thead{TimesFM}} & \rotatebox{45}{\thead{TTM}} & \rotatebox{45}{\thead{AutoARIMA}} & \rotatebox{45}{\thead{TFT}} & \rotatebox{45}{\thead{DeepAR}} & \rotatebox{45}{\thead{PatchTSTx}} & \rotatebox{45}{\thead{NBEATSx-G}} & \rotatebox{45}{\thead{NBEATSx-I}} & \rotatebox{45}{\thead{NHiTS}} & \rotatebox{45}{\thead{Seasonal Naive}} & \rotatebox{45}{\thead{TiDE}} & \rotatebox{45}{\thead{DLinear}} \\
\midrule
ETT (15 Min.) & 0.338 & 0.361 & 0.368 & \bftable 0.213 & 0.263 & 0.394 & 0.258 & \underline{0.224} & 0.259 & 0.513 & 0.440 & 0.261 & 0.254 & 0.225 & 0.269 & 0.790 & 0.451 & \bftable \underline{0.208} & 0.404 & 0.239 & 0.677 & 0.472 & 0.377 \\
ETT (Hourly) & 0.354 & 0.475 & 0.433 & 0.352 & 0.490 & 0.456 & 0.401 & \bftable \underline{0.297} & 0.489 & 0.900 & 0.635 & 0.437 & 0.834 & 0.469 & 0.599 & \underline{0.351} & 0.463 & 0.752 & 0.677 & 0.443 & 0.950 & 0.618 & \bftable 0.311 \\
M5 & \bftable \underline{0.907} & 1.071 & 1.104 & 0.925 & 1.091 & 1.151 & 1.182 & \underline{0.919} & 0.959 & 0.925 & 1.088 & 1.064 & 1.094 & - & \bftable 0.911 & 0.938 & 1.038 & 0.946 & 0.953 & 1.326 & 1.268 & 0.919 & 1.056 \\
\bottomrule
\end{tabular}

    }
\end{table}
\clearpage
\newpage\clearpage
\subsection{Evaluations on covariate injection on pretrained models with either IIB, OIB or both.}
\label{sec::appendix:rq2}

In this section we present results where where \thismodel has either covariates of the past (IIB), covariates of the future (OIB), or both (IIB and OIB).

\begin{figure}[t]
    \centering
    \begin{subfigure}{.48\linewidth}
    \includegraphics[width=1\linewidth]{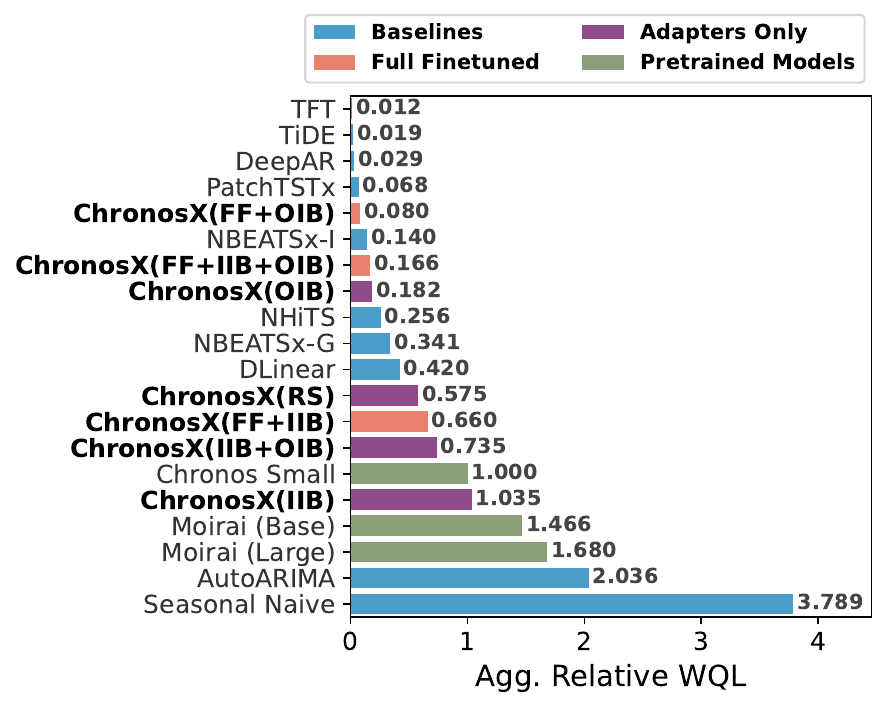}
    \caption{Agg. Rel. WQL on \textbf{Simple} synthetic datasets.}
    \label{fig:appendix:datasets_synthetic_simple_main_wql_v2}
    \end{subfigure}
    \quad
    \begin{subfigure}{.48\linewidth}
    \includegraphics[width=1\linewidth]{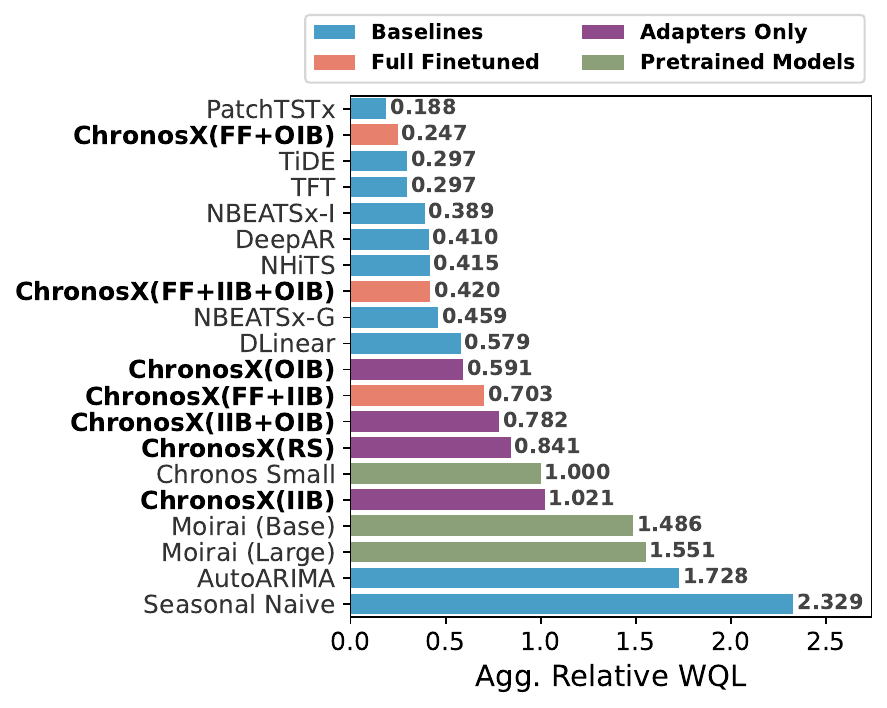}
    \caption{Agg. Rel. WQL on \textbf{Complex} synthetic datasets.}
    \label{fig:appendix:datasets_synthetic_complex_main_mase_v2}
    \end{subfigure}
    \\
    \begin{subfigure}{.48\linewidth}
    \includegraphics[width=1\linewidth]{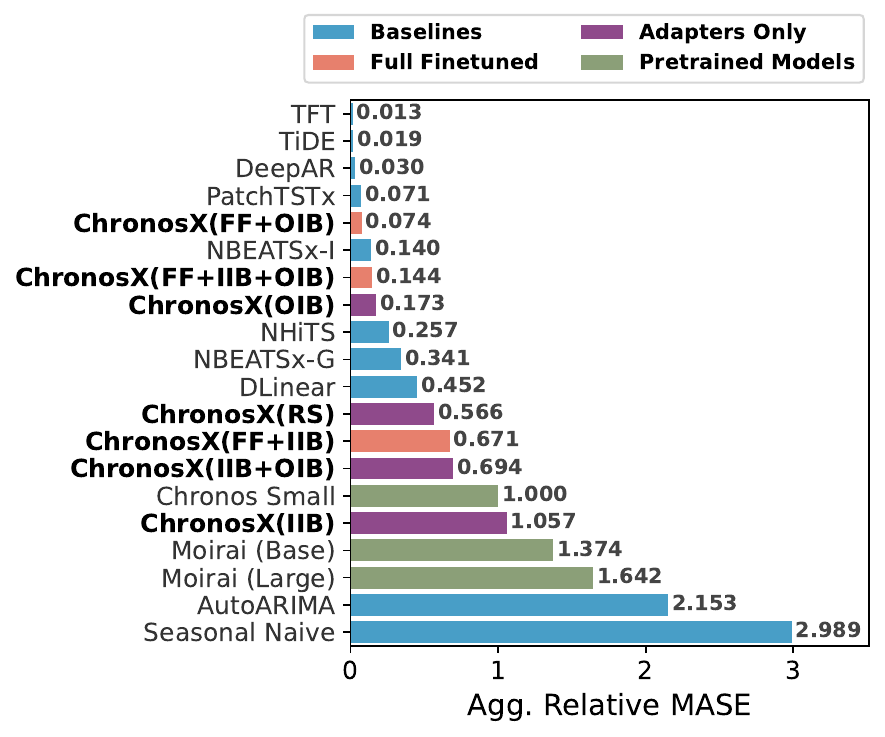}
    \caption{Agg. Rel. MASE on \textbf{Simple} synthetic datasets}
    \label{fig:appendix:datasets_synthetic_simple_main_wql_v2}
    \end{subfigure}
    \quad
    \begin{subfigure}{.48\linewidth}
    \includegraphics[width=1\linewidth]{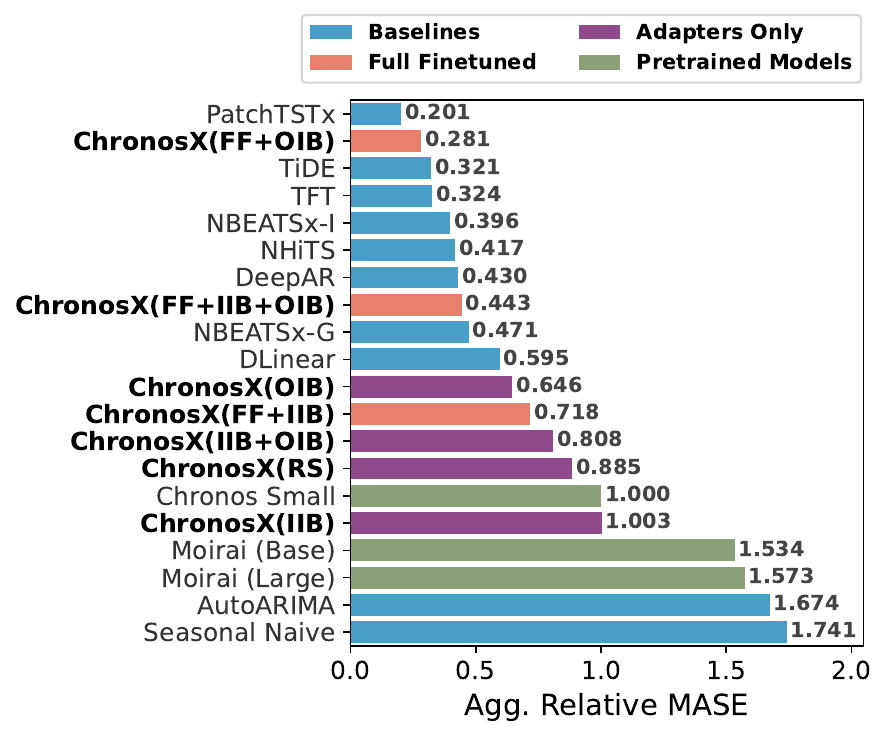}
    \caption{Agg. Rel. MASE on \textbf{Complex} synthetic datasets}
    \label{fig:appendix:datasets_synthetic_complex_main_mase_v2}
    \end{subfigure}
    \\
    \begin{subfigure}{.48\linewidth}
    \includegraphics[width=1\linewidth]{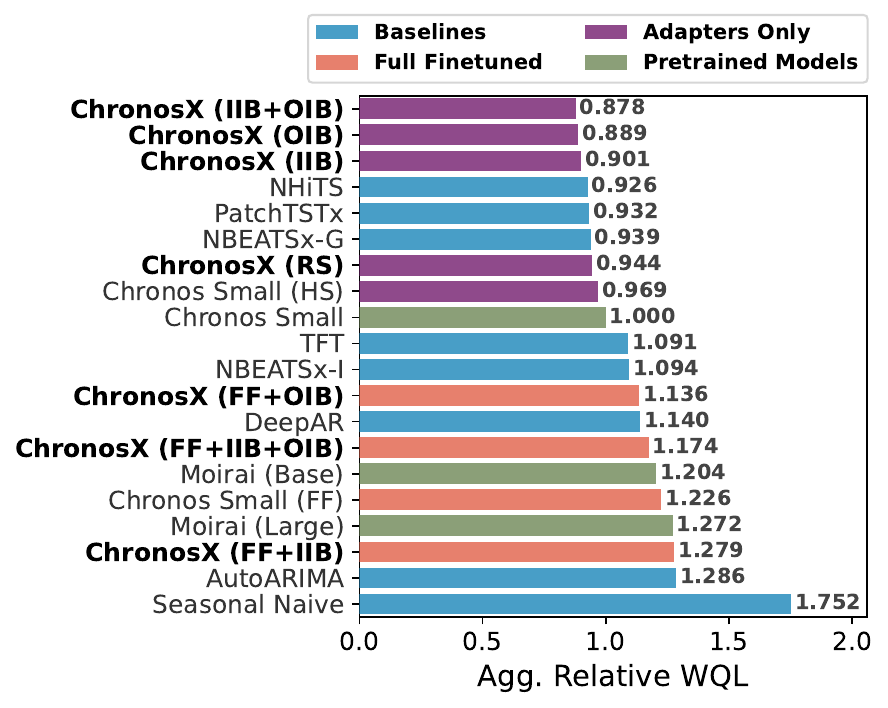}
    \caption{Agg. Rel. WQL on real datasets with covariates.}
    \label{fig:appendix:ciptfinal_real_grouped_aggrel_wql_v2}
    \end{subfigure}
    \quad
    \begin{subfigure}{.475\linewidth}
    \includegraphics[width=1\linewidth]{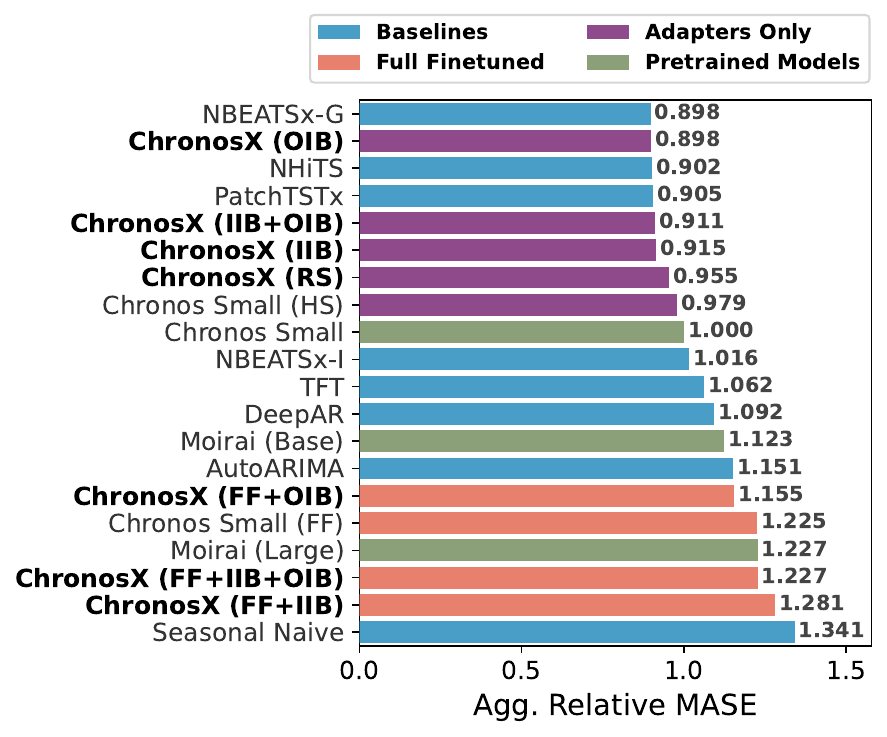}
    \caption{Agg. Rel. MASE on real datasets with covariates.}
    \label{fig:appendix:ciptfinal_real_grouped_aggrel_mase_v2}
    \end{subfigure}
    \caption{Aggregated Relative WQL and MASE on simple, complex, and real datasets. Emphasis on ablations of \thismodel with either past covariates, future covariates, or both.
    }
    \label{fig:appendix:ciptfinal_simple_synthetic_grouped_AggRel_ALL_v2}
\end{figure}

We present the following results:
\begin{itemize}[leftmargin=15pt]
    \item In Fig.~\ref{fig:appendix:ciptfinal_simple_synthetic_grouped_AggRel_ALL_v2} we present the Aggregated Relative WQL and MASE on simple, complex, and real datasets,
    \item In~\Cref{tab:appendix:datasets_synthetic_simple_main_wql_v2} and~\Cref{tab:appendix:datasets_synthetic_complex_main_wql_v2} we present the  WQL on simple and complex datasets, respectively,
    \item In~\Cref{tab:appendix:datasets_synthetic_simple_main_mase_v2} and~\Cref{tab:appendix:datasets_synthetic_complex_main_mase_v2} we present the MASE on simple datasets, respectively.
    \item In~\Cref{tab:appendix:ciptfinal_real_grouped_wql_v2} and~\Cref{tab:appendix:ciptfinal_real_grouped_mase_v2} we present WQL and MASE on real datasets, respectively.
\end{itemize}

\clearpage

\begin{table}[H]
    \centering
    \caption{WQL scores per model in 16 \textbf{simple} synthetic datasets. Models achieving the \underline{\textbf{first}}, \textbf{second}, and \underline{third} best scores have been highlighted. Scores reported are averaged over three random seeds.}
    \label{tab:appendix:datasets_synthetic_simple_main_wql_v2}
    \rowcolors{2}{gray!25}{white}
    \renewcommand\theadfont{}
    \resizebox{\textwidth}{!}{%
\begin{tabular}{lcccccccccccccccccccc}
\toprule
 & \multicolumn{3}{c}{Full Finetuned} & \multicolumn{4}{c}{Adapters Only} & \multicolumn{3}{c}{Pretrained Models} & \multicolumn{10}{c}{Baselines} \\
 \cmidrule(lr){2-4} \cmidrule(lr){5-8} \cmidrule(lr){9-11} \cmidrule(lr){12-21}
\rowcolor{white} & \rotatebox{45}{\thead{ChronosX(FF+IIB+OIB)}} & \rotatebox{45}{\thead{ChronosX(FF+OIB)}} & \rotatebox{45}{\thead{ChronosX(FF+IIB)}} & \rotatebox{45}{\thead{ChronosX(IIB+OIB)}} & \rotatebox{45}{\thead{ChronosX(OIB)}} & \rotatebox{45}{\thead{ChronosX(IIB)}} & \rotatebox{45}{\thead{ChronosX(RS)}} & \rotatebox{45}{\thead{Chronos Small}} & \rotatebox{45}{\thead{Moirai (Large)}} & \rotatebox{45}{\thead{Moirai (Base)}} & \rotatebox{45}{\thead{AutoARIMA}} & \rotatebox{45}{\thead{TFT}} & \rotatebox{45}{\thead{DeepAR}} & \rotatebox{45}{\thead{PatchTSTx}} & \rotatebox{45}{\thead{NBEATSx-I}} & \rotatebox{45}{\thead{NBEATSx-G}} & \rotatebox{45}{\thead{NHiTS}} & \rotatebox{45}{\thead{Seasonal Naive}} & \rotatebox{45}{\thead{TiDE}} & \rotatebox{45}{\thead{DLinear}} \\
\midrule
Simple ARP (Add) & 0.164 & 0.018 & 0.045 & 0.168 & 0.043 & 0.060 & 0.055 & 0.056 & 0.148 & 0.099 & 0.198 & 0.007 & 0.007 & \bftable \underline{0.002} & \underline{0.005} & 0.015 & 0.012 & 0.402 & \bftable 0.003 & 0.022 \\
Simple ARP (Mult.) & 0.042 & 0.007 & 0.028 & 0.136 & 0.024 & 0.041 & 0.037 & 0.037 & 0.153 & 0.114 & 0.313 & \bftable \underline{0.001} & 0.004 & \underline{0.003} & 0.012 & 0.018 & 0.014 & 0.572 & \bftable 0.002 & 0.030 \\
Simple Bells (Add) & 0.287 & 0.271 & 0.288 & 0.225 & 0.185 & 0.252 & 0.318 & 0.319 & 0.383 & 0.346 & 0.176 & \underline{0.005} & 0.056 & \bftable \underline{0.003} & 0.048 & 0.061 & 0.037 & 0.582 & \bftable 0.005 & 0.060 \\
Simple Bells (Mult.) & 0.041 & 0.008 & 0.040 & 0.176 & 0.022 & 0.083 & 0.089 & 0.089 & 0.156 & 0.136 & 0.273 & \bftable \underline{0.001} & 0.005 & \underline{0.004} & 0.040 & 0.037 & 0.033 & 0.417 & \bftable 0.002 & 0.093 \\
\bottomrule
\end{tabular}

    }
\end{table}

\begin{table}[H]
    \centering
    \caption{WQL scores per model in 16 \textbf{complex} synthetic datasets. Models achieving the \underline{\textbf{first}}, \textbf{second}, and \underline{third} best scores have been highlighted. Scores reported are averaged over three random seeds.}
    \label{tab:appendix:datasets_synthetic_complex_main_wql_v2}
    \rowcolors{2}{gray!25}{white}
    \renewcommand\theadfont{}
    \resizebox{\textwidth}{!}{%
    \begin{tabular}{lcccccccccccccccccccc}
\toprule
 & \multicolumn{3}{c}{Full Finetuned} & \multicolumn{4}{c}{Adapters Only} & \multicolumn{3}{c}{Pretrained Models} & \multicolumn{10}{c}{Baselines} \\
 \cmidrule(lr){2-4} \cmidrule(lr){5-8} \cmidrule(lr){9-11} \cmidrule(lr){12-21}
\rowcolor{white} & \rotatebox{45}{\thead{ChronosX(FF+IIB+OIB)}} & \rotatebox{45}{\thead{ChronosX(FF+OIB)}} & \rotatebox{45}{\thead{ChronosX(FF+IIB)}} & \rotatebox{45}{\thead{ChronosX(IIB+OIB)}} & \rotatebox{45}{\thead{ChronosX(OIB)}} & \rotatebox{45}{\thead{ChronosX(IIB)}} & \rotatebox{45}{\thead{ChronosX(RS)}} & \rotatebox{45}{\thead{Chronos Small}} & \rotatebox{45}{\thead{Moirai (Large)}} & \rotatebox{45}{\thead{Moirai (Base)}} & \rotatebox{45}{\thead{AutoARIMA}} & \rotatebox{45}{\thead{TFT}} & \rotatebox{45}{\thead{DeepAR}} & \rotatebox{45}{\thead{PatchTSTx}} & \rotatebox{45}{\thead{NBEATSx-I}} & \rotatebox{45}{\thead{NBEATSx-G}} & \rotatebox{45}{\thead{NHiTS}} & \rotatebox{45}{\thead{Seasonal Naive}} & \rotatebox{45}{\thead{TiDE}} & \rotatebox{45}{\thead{DLinear}} \\
\midrule
Diverse ARP (Add) & 0.092 & 0.034 & 0.037 & 0.117 & 0.054 & 0.062 & 0.063 & 0.063 & 0.167 & 0.138 & 0.265 & 0.023 & 0.032 & \bftable \underline{0.007} & \bftable 0.009 & 0.019 & \underline{0.012} & 0.250 & 0.019 & 0.028 \\
Diverse ARP (Mult.) & 0.036 & 0.031 & 0.036 & 0.077 & 0.048 & 0.060 & 0.059 & 0.059 & 0.153 & 0.137 & 0.269 & 0.030 & 0.029 & \bftable \underline{0.016} & \bftable 0.024 & 0.028 & \underline{0.024} & 0.221 & 0.027 & 0.041 \\
Diverse Bells (Add) & 0.208 & 0.192 & 0.211 & 0.271 & 0.210 & 0.247 & 0.259 & 0.257 & 0.317 & 0.313 & 0.168 & \bftable 0.027 & 0.161 & \bftable \underline{0.011} & 1.162 & 0.105 & 0.124 & 0.464 & \underline{0.035} & 0.081 \\
Diverse Bells (Mult.) & 0.051 & \bftable 0.020 & 0.052 & 0.116 & 0.074 & 0.097 & 0.088 & 0.089 & 0.162 & 0.155 & 0.233 & \underline{0.036} & 0.081 & \bftable \underline{0.012} & 0.050 & 0.067 & 0.055 & 0.213 & 0.040 & 0.084 \\
\bottomrule
\end{tabular}

    }
\end{table}

\begin{table}[H]
    \centering
    \caption{MASE scores per model in 16 \textbf{simple} synthetic datasets. Models achieving the \underline{\textbf{first}}, \textbf{second}, and \underline{third} best scores have been highlighted. Scores reported are averaged over three random seeds.}
    \label{tab:appendix:datasets_synthetic_simple_main_mase_v2}
    \rowcolors{2}{gray!25}{white}
    \renewcommand\theadfont{}
    \resizebox{\textwidth}{!}{%
    \begin{tabular}{lcccccccccccccccccccc}
\toprule
 & \multicolumn{3}{c}{Full Finetuned} & \multicolumn{4}{c}{Adapters Only} & \multicolumn{3}{c}{Pretrained Models} & \multicolumn{10}{c}{Baselines} \\
 \cmidrule(lr){2-4} \cmidrule(lr){5-8} \cmidrule(lr){9-11} \cmidrule(lr){12-21}
\rowcolor{white} & \rotatebox{45}{\thead{ChronosX(FF+IIB+OIB)}} & \rotatebox{45}{\thead{ChronosX(FF+OIB)}} & \rotatebox{45}{\thead{ChronosX(FF+IIB)}} & \rotatebox{45}{\thead{ChronosX(IIB+OIB)}} & \rotatebox{45}{\thead{ChronosX(OIB)}} & \rotatebox{45}{\thead{ChronosX(IIB)}} & \rotatebox{45}{\thead{ChronosX(RS)}} & \rotatebox{45}{\thead{Chronos Small}} & \rotatebox{45}{\thead{Moirai (Large)}} & \rotatebox{45}{\thead{Moirai (Base)}} & \rotatebox{45}{\thead{AutoARIMA}} & \rotatebox{45}{\thead{TFT}} & \rotatebox{45}{\thead{DeepAR}} & \rotatebox{45}{\thead{PatchTSTx}} & \rotatebox{45}{\thead{NBEATSx-I}} & \rotatebox{45}{\thead{NBEATSx-G}} & \rotatebox{45}{\thead{NHiTS}} & \rotatebox{45}{\thead{Seasonal Naive}} & \rotatebox{45}{\thead{TiDE}} & \rotatebox{45}{\thead{DLinear}} \\
\midrule
Simple ARP (Add) & 0.947 & 0.139 & 0.262 & 0.951 & 0.261 & 0.344 & 0.317 & 0.322 & 0.849 & 0.571 & 1.103 & 0.039 & 0.034 & \bftable \underline{0.013} & \underline{0.029} & 0.086 & 0.067 & 1.788 & \bftable 0.019 & 0.122 \\
Simple ARP (Mult.) & 0.161 & 0.033 & 0.127 & 0.500 & 0.110 & 0.176 & 0.156 & 0.157 & 0.642 & 0.454 & 1.616 & \bftable \underline{0.006} & 0.020 & \underline{0.014} & 0.051 & 0.070 & 0.057 & 1.930 & \bftable 0.009 & 0.128 \\
Simple Bells (Add) & 3.203 & 3.044 & 3.227 & 2.459 & 2.119 & 2.805 & 3.474 & 3.483 & 4.069 & 3.689 & 1.993 & \bftable 0.055 & 0.613 & \bftable \underline{0.035} & 0.491 & 0.641 & 0.392 & 4.951 & \underline{0.059} & 0.682 \\
Simple Bells (Mult.) & 0.220 & 0.046 & 0.214 & 0.858 & 0.130 & 0.462 & 0.472 & 0.473 & 0.847 & 0.691 & 1.698 & \bftable \underline{0.004} & 0.025 & \underline{0.019} & 0.212 & 0.206 & 0.177 & 1.882 & \bftable 0.012 & 0.583 \\
\bottomrule
\end{tabular}

    }
\end{table}

\begin{table}[H]
    \centering
    \caption{MASE scores per model in 16 \textbf{complex} synthetic datasets. Models achieving the \underline{\textbf{first}}, \textbf{second}, and \underline{third} best scores have been highlighted. Scores reported are averaged over three random seeds.}
    \label{tab:appendix:datasets_synthetic_complex_main_mase_v2}
    \rowcolors{2}{gray!25}{white}
    \renewcommand\theadfont{}
    \resizebox{\textwidth}{!}{%
    \begin{tabular}{lcccccccccccccccccccc}
\toprule
 & \multicolumn{3}{c}{Full Finetuned} & \multicolumn{4}{c}{Adapters Only} & \multicolumn{3}{c}{Pretrained Models} & \multicolumn{10}{c}{Baselines} \\
 \cmidrule(lr){2-4} \cmidrule(lr){5-8} \cmidrule(lr){9-11} \cmidrule(lr){12-21}
\rowcolor{white} & \rotatebox{45}{\thead{ChronosX(FF+IIB+OIB)}} & \rotatebox{45}{\thead{ChronosX(FF+OIB)}} & \rotatebox{45}{\thead{ChronosX(FF+IIB)}} & \rotatebox{45}{\thead{ChronosX(IIB+OIB)}} & \rotatebox{45}{\thead{ChronosX(OIB)}} & \rotatebox{45}{\thead{ChronosX(IIB)}} & \rotatebox{45}{\thead{ChronosX(RS)}} & \rotatebox{45}{\thead{Chronos Small}} & \rotatebox{45}{\thead{Moirai (Large)}} & \rotatebox{45}{\thead{Moirai (Base)}} & \rotatebox{45}{\thead{AutoARIMA}} & \rotatebox{45}{\thead{TFT}} & \rotatebox{45}{\thead{DeepAR}} & \rotatebox{45}{\thead{PatchTSTx}} & \rotatebox{45}{\thead{NBEATSx-I}} & \rotatebox{45}{\thead{NBEATSx-G}} & \rotatebox{45}{\thead{NHiTS}} & \rotatebox{45}{\thead{Seasonal Naive}} & \rotatebox{45}{\thead{TiDE}} & \rotatebox{45}{\thead{DLinear}} \\
\midrule
Diverse ARP (Add) & 1.004 & 0.461 & 0.488 & 1.282 & 0.687 & 0.758 & 0.769 & 0.768 & 2.056 & 1.848 & 2.801 & 0.317 & 0.419 & \bftable \underline{0.099} & \bftable 0.110 & 0.242 & \underline{0.148} & 1.947 & 0.270 & 0.318 \\
Diverse ARP (Mult.) & 0.500 & 0.446 & 0.484 & 0.987 & 0.655 & 0.754 & 0.767 & 0.772 & 1.928 & 1.791 & 3.091 & 0.404 & 0.370 & \bftable \underline{0.221} & \bftable 0.314 & 0.374 & \underline{0.323} & 1.921 & 0.360 & 0.533 \\
Diverse Bells (Add) & 3.392 & 3.178 & 3.508 & 4.066 & 3.378 & 3.838 & 4.000 & 3.975 & 4.634 & 4.713 & 2.234 & \bftable 0.446 & 2.513 & \bftable \underline{0.169} & 12.561 & 1.584 & 1.512 & 5.563 & \underline{0.586} & 1.249 \\
Diverse Bells (Mult.) & 0.925 & \bftable 0.456 & 0.929 & 1.959 & 1.414 & 1.731 & 1.591 & 1.620 & 2.834 & 2.755 & 3.453 & \underline{0.651} & 1.462 & \bftable \underline{0.249} & 0.920 & 1.171 & 1.000 & 2.698 & 0.721 & 1.619 \\
\bottomrule
\end{tabular}

    }
\end{table}

\begin{table}[H]
    \centering
    \caption{WQL scores per model in real datasets with covariates. Models achieving the \underline{\textbf{first}}, \textbf{second}, and \underline{third} best scores have been highlighted. Scores reported are averaged over three random seeds.}
    \label{tab:appendix:ciptfinal_real_grouped_wql_v2}
    \rowcolors{2}{gray!25}{white}
    \renewcommand\theadfont{}
    \resizebox{\textwidth}{!}{    \begin{tabular}{lcccccccccccccccccccc}
\toprule
 & \multicolumn{4}{c}{Full Finetuned} & \multicolumn{5}{c}{Adapters Only} & \multicolumn{3}{c}{Pretrained Models} & \multicolumn{8}{c}{Baselines} \\
 \cmidrule(lr){2-5} \cmidrule(lr){6-10} \cmidrule(lr){11-13} \cmidrule(lr){14-21}
\rowcolor{white} & \rotatebox{45}{\thead{ChronosX (FF+IIB+OIB)}} & \rotatebox{45}{\thead{ChronosX (FF+OIB)}} & \rotatebox{45}{\thead{ChronosX (FF+IIB)}} & \rotatebox{45}{\thead{Chronos Small (FF)}} & \rotatebox{45}{\thead{ChronosX (IIB+OIB)}} & \rotatebox{45}{\thead{ChronosX (OIB)}} & \rotatebox{45}{\thead{ChronosX (IIB)}} & \rotatebox{45}{\thead{ChronosX (RS)}} & \rotatebox{45}{\thead{Chronos Small (HS)}} & \rotatebox{45}{\thead{Chronos Small}} & \rotatebox{45}{\thead{Moirai (Base)}} & \rotatebox{45}{\thead{Moirai (Large)}} & \rotatebox{45}{\thead{AutoARIMA}} & \rotatebox{45}{\thead{TFT}} & \rotatebox{45}{\thead{DeepAR}} & \rotatebox{45}{\thead{PatchTSTx}} & \rotatebox{45}{\thead{NBEATSx-G}} & \rotatebox{45}{\thead{NBEATSx-I}} & \rotatebox{45}{\thead{NHiTS}} & \rotatebox{45}{\thead{Seasonal Naive}} \\
\midrule
ETT (15 Min.) & 0.022 & 0.020 & 0.026 & 0.023 & \underline{0.016} & 0.017 & 0.016 & \bftable 0.015 & 0.017 & 0.016 & 0.019 & 0.040 & 0.020 & 0.020 & 0.072 & 0.035 & 0.016 & 0.027 & \bftable \underline{0.015} & 0.065 \\
ETT (Hourly) & 0.025 & 0.024 & 0.025 & 0.024 & 0.025 & \bftable 0.023 & 0.024 & 0.024 & \underline{0.023} & \bftable \underline{0.022} & 0.037 & 0.076 & 0.045 & 0.045 & 0.026 & 0.036 & 0.065 & 0.063 & 0.035 & 0.101 \\
M5 & \underline{0.538} & 0.538 & \bftable \underline{0.537} & \bftable 0.537 & 0.557 & 0.551 & 0.552 & 0.551 & 0.549 & 0.549 & 0.586 & 0.576 & - & 0.560 & 0.617 & 0.812 & 0.607 & 0.604 & 0.748 & 0.934 \\
Electricity-BE & 0.058 & 0.056 & 0.058 & 0.054 & 0.049 & 0.047 & 0.051 & 0.047 & 0.047 & 0.049 & 0.069 & 0.075 & 0.080 & 0.087 & 0.078 & \underline{0.040} & \bftable \underline{0.036} & 0.054 & \bftable 0.038 & 0.071 \\
\bottomrule
\end{tabular}

    }
\end{table}

\begin{table}[H]
    \centering
    \caption{MASE scores per model in real datasets with covariates. Models achieving the \underline{\textbf{first}}, \textbf{second}, and \underline{third} best scores have been highlighted. Scores reported are averaged over three random seeds.}
    \label{tab:appendix:ciptfinal_real_grouped_mase_v2}
    \rowcolors{2}{gray!25}{white}
    \renewcommand\theadfont{}
    \resizebox{\textwidth}{!}{%
    \begin{tabular}{lcccccccccccccccccccc}
\toprule
 & \multicolumn{4}{c}{Full Finetuned} & \multicolumn{5}{c}{Adapters Only} & \multicolumn{3}{c}{Pretrained Models} & \multicolumn{8}{c}{Baselines} \\
 \cmidrule(lr){2-5} \cmidrule(lr){6-10} \cmidrule(lr){11-13} \cmidrule(lr){14-21}
\rowcolor{white} & \rotatebox{45}{\thead{ChronosX (FF+IIB+OIB)}} & \rotatebox{45}{\thead{ChronosX (FF+OIB)}} & \rotatebox{45}{\thead{ChronosX (FF+IIB)}} & \rotatebox{45}{\thead{Chronos Small (FF)}} & \rotatebox{45}{\thead{ChronosX (IIB+OIB)}} & \rotatebox{45}{\thead{ChronosX (OIB)}} & \rotatebox{45}{\thead{ChronosX (IIB)}} & \rotatebox{45}{\thead{ChronosX (RS)}} & \rotatebox{45}{\thead{Chronos Small (HS)}} & \rotatebox{45}{\thead{Chronos Small}} & \rotatebox{45}{\thead{Moirai (Base)}} & \rotatebox{45}{\thead{Moirai (Large)}} & \rotatebox{45}{\thead{AutoARIMA}} & \rotatebox{45}{\thead{TFT}} & \rotatebox{45}{\thead{DeepAR}} & \rotatebox{45}{\thead{PatchTSTx}} & \rotatebox{45}{\thead{NBEATSx-G}} & \rotatebox{45}{\thead{NBEATSx-I}} & \rotatebox{45}{\thead{NHiTS}} & \rotatebox{45}{\thead{Seasonal Naive}} \\
\midrule
ETT (15 Min.) & 0.338 & 0.297 & 0.352 & 0.337 & 0.213 & \bftable 0.207 & 0.212 & \bftable \underline{0.206} & 0.211 & 0.224 & 0.259 & 0.513 & 0.225 & 0.269 & 0.790 & 0.451 & \underline{0.208} & 0.404 & 0.239 & 0.677 \\
ETT (Hourly) & 0.354 & 0.341 & 0.346 & 0.333 & 0.352 & \underline{0.331} & 0.331 & 0.338 & \bftable 0.313 & \bftable \underline{0.297} & 0.489 & 0.900 & 0.469 & 0.599 & 0.351 & 0.463 & 0.752 & 0.677 & 0.443 & 0.950 \\
M5 & \underline{0.907} & 0.907 & \bftable 0.904 & \bftable \underline{0.903} & 0.925 & 0.918 & 0.925 & 0.918 & 0.915 & 0.919 & 0.959 & 0.925 & - & 0.911 & 0.938 & 1.038 & 0.946 & 0.953 & 1.326 & 1.268 \\
Electricity-BE & 0.432 & 0.413 & 0.435 & 0.400 & 0.374 & 0.373 & 0.378 & 0.372 & 0.374 & 0.380 & 0.458 & 0.510 & 0.306 & 0.621 & 0.568 & \underline{0.292} & \bftable \underline{0.276} & 0.406 & \bftable 0.276 & 0.405 \\
\bottomrule
\end{tabular}

    }
\end{table}
\newpage
\clearpage

\newpage
\clearpage

\newpage\clearpage
\subsection{Ablations showing that  performance improvement is brought by covariates}
\label{appendix:ablation_improvement_from_covariates}

In this section we present results where \thismodel (NC) and \thismodel (FF) (NC) are variants with no-covariates of our models \thismodel and \thismodel (FF), respectively.

\begin{figure}[H]
    \centering
    \begin{subfigure}{.325\linewidth}
        \includegraphics[width=1\linewidth]{Figures/ciptfinal_simple_synthetic_without_covariates_aistats/aggregated_plot_chronos_style_WQL_sd.pdf}
        \caption{WQL on \textbf{Simple} Synthetic Data.}
        \label{fig:ciptfinal_simple_synthetic_without_covariates_aistats_wql}
    \end{subfigure}
    \begin{subfigure}{.325\linewidth}
        \includegraphics[width=1\linewidth]{Figures/ciptfinal_complex_synthetic_without_covariates_aistats/aggregated_plot_chronos_style_WQL_sd.pdf}
        \caption{WQL on \textbf{Complex} Synthetic Data.}
        \label{fig:ciptfinal_complex_synthetic_without_covariates_aistats_wql}
    \end{subfigure}
\begin{subfigure}{.325\linewidth}
    \includegraphics[width=1\linewidth]{Figures/ciptfinal_real_without_covariates_aistats/aggregated_plot_chronos_style_WQL_cd.pdf}
    \caption{WQL  \textbf{Real}  Datasets}
    \label{fig:ciptfinal_real_without_covariates_aistats_wql}
\end{subfigure}
    \begin{subfigure}{.325\linewidth}
        \includegraphics[width=1\linewidth]{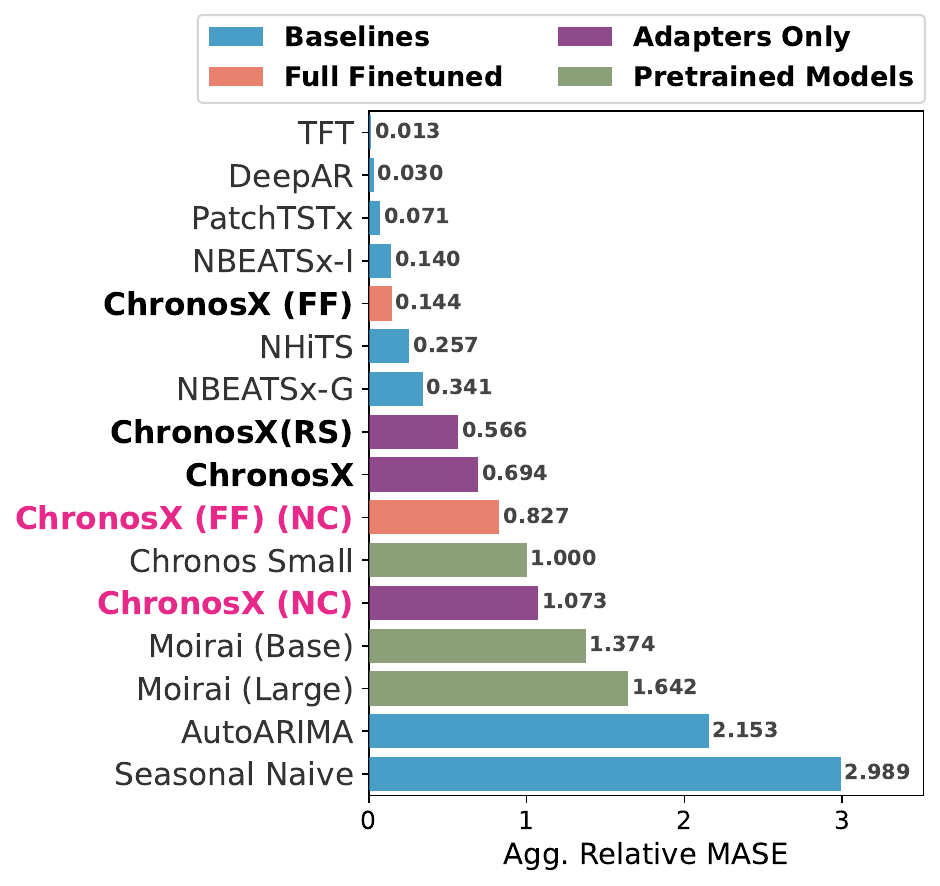}
        \caption{MASE on \textbf{Simple} Synthetic Data.}
        \label{fig:ciptfinal_simple_synthetic_without_covariates_aistats_mase}
    \end{subfigure}
    \begin{subfigure}{.325\linewidth}
        \includegraphics[width=1\linewidth]{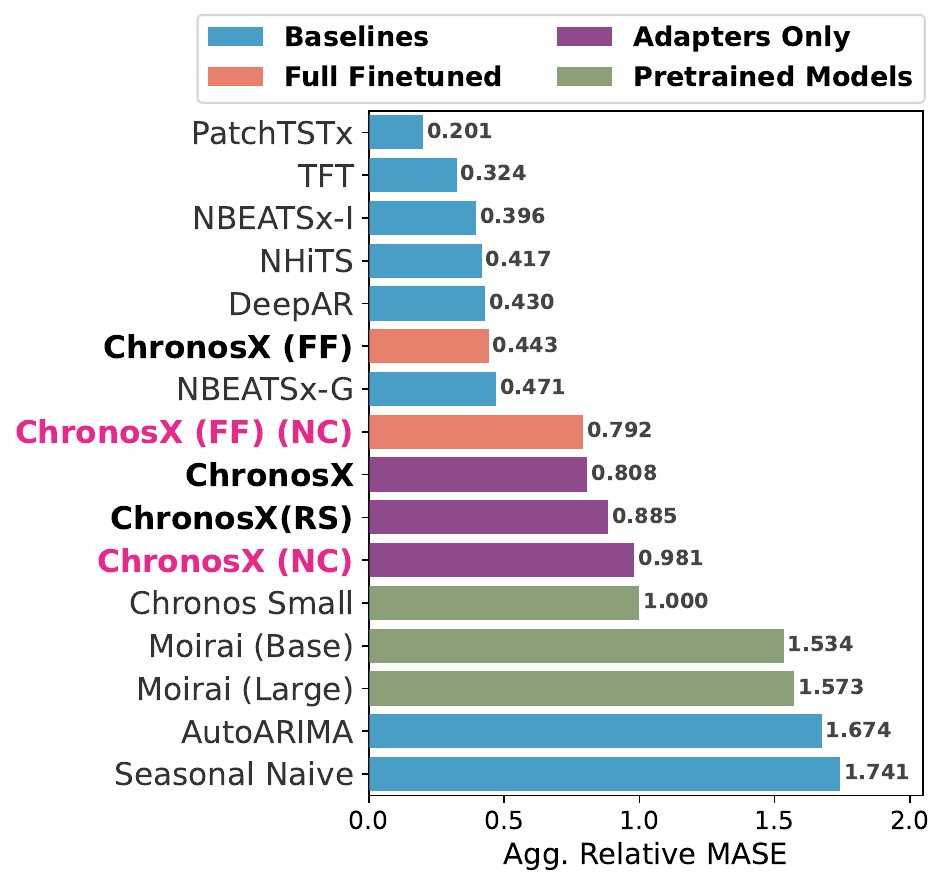}
        \caption{MASE on \textbf{Complex} Synthetic Data.}
        \label{fig:ciptfinal_complex_synthetic_without_covariates_aistats_mase}
    \end{subfigure}
\begin{subfigure}{.325\linewidth}
    \includegraphics[width=1\linewidth]{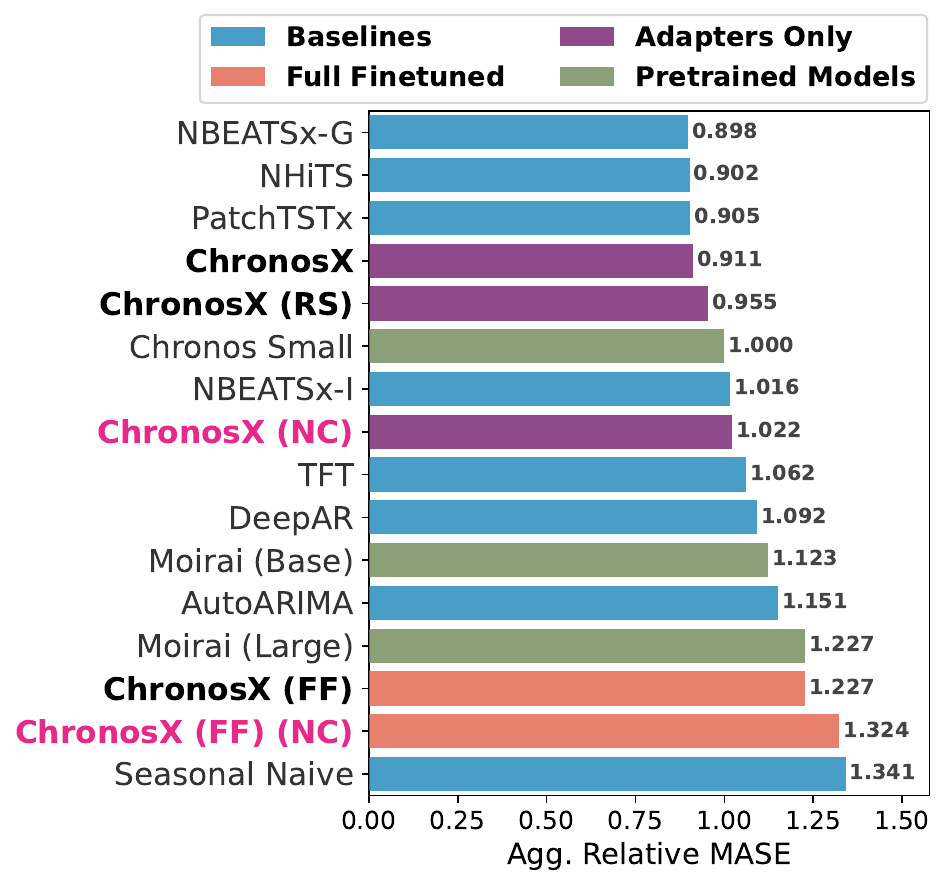}
    \caption{MASE on \textbf{Real}  Datasets}
    \label{fig:ciptfinal_real_without_covariates_aistats_mase}
\end{subfigure}
    \caption{
        Covariate ablation study. \thismodel (NC) and \thismodel (FF) (NC) are variants with no-covariates of our models \thismodel and \thismodel (FF), respectively. Overall model ablations with no-covariates perform worse.
    }
    \label{fig:nocovs_appendix}
\end{figure}

We present the following results:
\begin{itemize}[leftmargin=15pt]
    \item In Fig.~\ref{fig:nocovs_appendix} we present the Aggregated Relative WQL and MASE on simple, complex, and real datasets.
    \item In~\Cref{tab:appendix:ciptfinal_simple_synthetic_without_covariates_aistats_wql} and~\Cref{tab:appendix:ciptfinal_complex_synthetic_without_covariates_aistats_wql} we present the  WQL on simple and complex datasets, respectively,
    \item In~\Cref{tab:appendix:ciptfinal_simple_synthetic_without_covariates_aistats_mase} and~\Cref{tab:appendix:ciptfinal_complex_synthetic_without_covariates_aistats_mase} we present the MASE on simple datasets, respectively.
    \item In~\Cref{tab:appendix:ciptfinal_real_without_covariates_aistats_wql} and~\Cref{tab:appendix:ciptfinal_real_without_covariates_aistats_mase} we present WQL and MASE on real datasets, respectively.
\end{itemize}

\newpage\clearpage
\begin{table}[H]
    \centering
    \caption{WQL scores per model in 16 \textbf{simple} synthetic datasets. Models achieving the \underline{\textbf{first}}, \textbf{second}, and \underline{third} best scores have been highlighted. Scores reported are averaged over three random seeds.}
    \label{tab:appendix:ciptfinal_simple_synthetic_without_covariates_aistats_wql}
    \rowcolors{2}{gray!25}{white}
    \renewcommand\theadfont{}
    \resizebox{\textwidth}{!}{%
\begin{tabular}{lcccccccccccccccc}
\toprule
 & \multicolumn{2}{c}{Full Finetuned} & \multicolumn{3}{c}{Adapters Only} & \multicolumn{3}{c}{Pretrained Models} & \multicolumn{8}{c}{Baselines} \\
 \cmidrule(lr){2-3} \cmidrule(lr){4-6} \cmidrule(lr){7-9} \cmidrule(lr){10-17}
\rowcolor{white} & \rotatebox{45}{\thead{ChronosX (FF)}} & \rotatebox{45}{\thead{ChronosX (FF) (NC)}} & \rotatebox{45}{\thead{ChronosX}} & \rotatebox{45}{\thead{ChronosX (NC)}} & \rotatebox{45}{\thead{ChronosX(RS)}} & \rotatebox{45}{\thead{Chronos Small}} & \rotatebox{45}{\thead{Moirai (Large)}} & \rotatebox{45}{\thead{Moirai (Base)}} & \rotatebox{45}{\thead{NHiTS}} & \rotatebox{45}{\thead{AutoARIMA}} & \rotatebox{45}{\thead{TFT}} & \rotatebox{45}{\thead{NBEATSx-I}} & \rotatebox{45}{\thead{NBEATSx-G}} & \rotatebox{45}{\thead{DeepAR}} & \rotatebox{45}{\thead{PatchTSTx}} & \rotatebox{45}{\thead{Seasonal Naive}} \\
\midrule
Simple ARP (Add) & 0.164 & 0.045 & 0.168 & 0.071 & 0.055 & 0.056 & 0.148 & 0.099 & 0.012 & 0.198 & 0.007 & \bftable 0.005 & 0.015 & \underline{0.007} & \bftable \underline{0.002} & 0.402 \\
Simple ARP (Mult.) & 0.042 & 0.027 & 0.136 & 0.051 & 0.037 & 0.037 & 0.153 & 0.114 & 0.014 & 0.313 & \bftable \underline{0.001} & 0.012 & 0.018 & \underline{0.004} & \bftable 0.003 & 0.572 \\
Simple Bells (Add) & 0.287 & 0.292 & 0.225 & 0.347 & 0.318 & 0.319 & 0.383 & 0.346 & \underline{0.037} & 0.176 & \bftable 0.005 & 0.048 & 0.061 & 0.056 & \bftable \underline{0.003} & 0.582 \\
Simple Bells (Mult.) & 0.041 & 0.062 & 0.176 & 0.085 & 0.089 & 0.089 & 0.156 & 0.136 & 0.033 & 0.273 & \bftable \underline{0.001} & 0.040 & 0.037 & \underline{0.005} & \bftable 0.004 & 0.417 \\
Simple Spikes (Add) & 0.007 & 0.338 & 0.250 & 0.449 & 0.157 & 0.431 & 0.515 & 0.498 & 0.083 & 0.381 & \bftable \underline{0.003} & 0.036 & 0.117 & \bftable 0.003 & \underline{0.004} & 0.978 \\
\bottomrule
\end{tabular}

    }
\end{table}
\begin{table}[H]
    \centering
    \caption{WQL scores per model in 16 \textbf{complex} synthetic datasets. Models achieving the \underline{\textbf{first}}, \textbf{second}, and \underline{third} best scores have been highlighted. Scores reported are averaged over three random seeds.asdf}
    \label{tab:appendix:ciptfinal_complex_synthetic_without_covariates_aistats_wql}
    \rowcolors{2}{gray!25}{white}
    \renewcommand\theadfont{}
    \resizebox{\textwidth}{!}{%
    \begin{tabular}{lcccccccccccccccc}
\toprule
 & \multicolumn{2}{c}{Full Finetuned} & \multicolumn{3}{c}{Adapters Only} & \multicolumn{3}{c}{Pretrained Models} & \multicolumn{8}{c}{Baselines} \\
 \cmidrule(lr){2-3} \cmidrule(lr){4-6} \cmidrule(lr){7-9} \cmidrule(lr){10-17}
\rowcolor{white} & \rotatebox{45}{\thead{ChronosX (FF)}} & \rotatebox{45}{\thead{ChronosX (FF) (NC)}} & \rotatebox{45}{\thead{ChronosX}} & \rotatebox{45}{\thead{ChronosX (NC)}} & \rotatebox{45}{\thead{ChronosX(RS)}} & \rotatebox{45}{\thead{Chronos Small}} & \rotatebox{45}{\thead{Moirai (Large)}} & \rotatebox{45}{\thead{Moirai (Base)}} & \rotatebox{45}{\thead{NHiTS}} & \rotatebox{45}{\thead{AutoARIMA}} & \rotatebox{45}{\thead{TFT}} & \rotatebox{45}{\thead{NBEATSx-I}} & \rotatebox{45}{\thead{NBEATSx-G}} & \rotatebox{45}{\thead{DeepAR}} & \rotatebox{45}{\thead{PatchTSTx}} & \rotatebox{45}{\thead{Seasonal Naive}} \\
\midrule
Diverse ARP (Add) & 0.092 & 0.039 & 0.117 & 0.062 & 0.063 & 0.063 & 0.167 & 0.138 & \underline{0.012} & 0.265 & 0.023 & \bftable 0.009 & 0.019 & 0.032 & \bftable \underline{0.007} & 0.250 \\
Diverse ARP (Mult.) & 0.036 & 0.036 & 0.077 & 0.058 & 0.059 & 0.059 & 0.153 & 0.137 & \underline{0.024} & 0.269 & 0.030 & \bftable 0.024 & 0.028 & 0.029 & \bftable \underline{0.016} & 0.221 \\
Diverse Bells (Add) & 0.208 & 0.220 & 0.271 & 0.271 & 0.259 & 0.257 & 0.317 & 0.313 & 0.124 & 0.168 & \bftable 0.027 & 1.162 & \underline{0.105} & 0.161 & \bftable \underline{0.011} & 0.464 \\
Diverse Bells (Mult.) & 0.051 & 0.060 & 0.116 & 0.093 & 0.088 & 0.089 & 0.162 & 0.155 & 0.055 & 0.233 & \bftable 0.036 & \underline{0.050} & 0.067 & 0.081 & \bftable \underline{0.012} & 0.213 \\
Diverse Spikes (Add) & \bftable 0.016 & 0.180 & 0.075 & 0.219 & 0.108 & 0.230 & 0.288 & 0.279 & 0.049 & 0.352 & \underline{0.019} & 0.029 & 0.046 & 0.036 & \bftable \underline{0.006} & 0.460 \\
\bottomrule
\end{tabular}

    }
\end{table}

\begin{table}[H]
    \centering
    \caption{MASE scores per model in 16 \textbf{simple} synthetic datasets. Models achieving the \underline{\textbf{first}}, \textbf{second}, and \underline{third} best scores have been highlighted. Scores reported are averaged over three random seeds.}
    \label{tab:appendix:ciptfinal_simple_synthetic_without_covariates_aistats_mase}
    \rowcolors{2}{gray!25}{white}
    \renewcommand\theadfont{}
    \resizebox{\textwidth}{!}{%
    \begin{tabular}{lcccccccccccccccc}
\toprule
 & \multicolumn{2}{c}{Full Finetuned} & \multicolumn{3}{c}{Adapters Only} & \multicolumn{3}{c}{Pretrained Models} & \multicolumn{8}{c}{Baselines} \\
 \cmidrule(lr){2-3} \cmidrule(lr){4-6} \cmidrule(lr){7-9} \cmidrule(lr){10-17}
\rowcolor{white} & \rotatebox{45}{\thead{ChronosX (FF)}} & \rotatebox{45}{\thead{ChronosX (FF) (NC)}} & \rotatebox{45}{\thead{ChronosX}} & \rotatebox{45}{\thead{ChronosX (NC)}} & \rotatebox{45}{\thead{ChronosX(RS)}} & \rotatebox{45}{\thead{Chronos Small}} & \rotatebox{45}{\thead{Moirai (Large)}} & \rotatebox{45}{\thead{Moirai (Base)}} & \rotatebox{45}{\thead{NHiTS}} & \rotatebox{45}{\thead{AutoARIMA}} & \rotatebox{45}{\thead{TFT}} & \rotatebox{45}{\thead{NBEATSx-I}} & \rotatebox{45}{\thead{NBEATSx-G}} & \rotatebox{45}{\thead{DeepAR}} & \rotatebox{45}{\thead{PatchTSTx}} & \rotatebox{45}{\thead{Seasonal Naive}} \\
\midrule
Simple ARP (Add) & 0.947 & 0.268 & 0.951 & 0.403 & 0.317 & 0.322 & 0.849 & 0.571 & 0.067 & 1.103 & 0.039 & \bftable 0.029 & 0.086 & \underline{0.034} & \bftable \underline{0.013} & 1.788 \\
Simple ARP (Mult.) & 0.161 & 0.126 & 0.500 & 0.200 & 0.156 & 0.157 & 0.642 & 0.454 & 0.057 & 1.616 & \bftable \underline{0.006} & 0.051 & 0.070 & \underline{0.020} & \bftable 0.014 & 1.930 \\
Simple Bells (Add) & 3.203 & 3.255 & 2.459 & 3.717 & 3.474 & 3.483 & 4.069 & 3.689 & \underline{0.392} & 1.993 & \bftable 0.055 & 0.491 & 0.641 & 0.613 & \bftable \underline{0.035} & 4.951 \\
Simple Bells (Mult.) & 0.220 & 0.377 & 0.858 & 0.485 & 0.472 & 0.473 & 0.847 & 0.691 & 0.177 & 1.698 & \bftable \underline{0.004} & 0.212 & 0.206 & \underline{0.025} & \bftable 0.019 & 1.882 \\
Simple Spikes (Add) & \underline{0.010} & 0.515 & 0.372 & 0.676 & 0.292 & 0.653 & 0.823 & 0.756 & 0.146 & 0.644 & 0.013 & 0.071 & 0.231 & \bftable 0.009 & \bftable \underline{0.007} & 1.258 \\
\bottomrule
\end{tabular}

    }
\end{table}
\begin{table}[H]
    \centering
    \caption{MASE scores per model in 16 \textbf{complex} synthetic datasets. Models achieving the \underline{\textbf{first}}, \textbf{second}, and \underline{third} best scores have been highlighted. Scores reported are averaged over three random seeds.}
    \label{tab:appendix:ciptfinal_complex_synthetic_without_covariates_aistats_mase}
    \rowcolors{2}{gray!25}{white}
    \renewcommand\theadfont{}
    \resizebox{\textwidth}{!}{%
    \begin{tabular}{lcccccccccccccccc}
\toprule
 & \multicolumn{2}{c}{Full Finetuned} & \multicolumn{3}{c}{Adapters Only} & \multicolumn{3}{c}{Pretrained Models} & \multicolumn{8}{c}{Baselines} \\
 \cmidrule(lr){2-3} \cmidrule(lr){4-6} \cmidrule(lr){7-9} \cmidrule(lr){10-17}
\rowcolor{white} & \rotatebox{45}{\thead{ChronosX (FF)}} & \rotatebox{45}{\thead{ChronosX (FF) (NC)}} & \rotatebox{45}{\thead{ChronosX}} & \rotatebox{45}{\thead{ChronosX (NC)}} & \rotatebox{45}{\thead{ChronosX(RS)}} & \rotatebox{45}{\thead{Chronos Small}} & \rotatebox{45}{\thead{Moirai (Large)}} & \rotatebox{45}{\thead{Moirai (Base)}} & \rotatebox{45}{\thead{NHiTS}} & \rotatebox{45}{\thead{AutoARIMA}} & \rotatebox{45}{\thead{TFT}} & \rotatebox{45}{\thead{NBEATSx-I}} & \rotatebox{45}{\thead{NBEATSx-G}} & \rotatebox{45}{\thead{DeepAR}} & \rotatebox{45}{\thead{PatchTSTx}} & \rotatebox{45}{\thead{Seasonal Naive}} \\
\midrule
Diverse ARP (Add) & 1.004 & 0.513 & 1.282 & 0.769 & 0.769 & 0.768 & 2.056 & 1.848 & \underline{0.148} & 2.801 & 0.317 & \bftable 0.110 & 0.242 & 0.419 & \bftable \underline{0.099} & 1.947 \\
Diverse ARP (Mult.) & 0.500 & 0.491 & 0.987 & 0.740 & 0.767 & 0.772 & 1.928 & 1.791 & \underline{0.323} & 3.091 & 0.404 & \bftable 0.314 & 0.374 & 0.370 & \bftable \underline{0.221} & 1.921 \\
Diverse Bells (Add) & 3.392 & 3.608 & 4.066 & 4.079 & 4.000 & 3.975 & 4.634 & 4.713 & \underline{1.512} & 2.234 & \bftable 0.446 & 12.561 & 1.584 & 2.513 & \bftable \underline{0.169} & 5.563 \\
Diverse Bells (Mult.) & 0.925 & 1.191 & 1.959 & 1.661 & 1.591 & 1.620 & 2.834 & 2.755 & 1.000 & 3.453 & \bftable 0.651 & \underline{0.920} & 1.171 & 1.462 & \bftable \underline{0.249} & 2.698 \\
Diverse Spikes (Add) & \bftable 0.062 & 0.599 & 0.294 & 0.733 & 0.439 & 0.767 & 1.048 & 0.999 & 0.163 & 1.203 & \underline{0.077} & 0.107 & 0.176 & 0.132 & \bftable \underline{0.022} & 1.286 \\
\bottomrule
\end{tabular}

    }
\end{table}

\begin{table}[H]
    \centering
    \caption{WQL scores per model in real datasets with covariates. Models achieving the \underline{\textbf{first}}, \textbf{second}, and \underline{third} best scores have been highlighted. Scores reported are averaged over three random seeds.}
    \label{tab:appendix:ciptfinal_real_without_covariates_aistats_wql}
    \rowcolors{2}{gray!25}{white}
    \renewcommand\theadfont{}
    \resizebox{\textwidth}{!}{    \begin{tabular}{lcccccccccccccccc}
\toprule
 & \multicolumn{2}{c}{Full Finetuned} & \multicolumn{3}{c}{Adapters Only} & \multicolumn{3}{c}{Pretrained Models} & \multicolumn{8}{c}{Baselines} \\
 \cmidrule(lr){2-3} \cmidrule(lr){4-6} \cmidrule(lr){7-9} \cmidrule(lr){10-17}
\rowcolor{white} & \rotatebox{45}{\thead{ChronosX (FF)}} & \rotatebox{45}{\thead{ChronosX (FF) (NC)}} & \rotatebox{45}{\thead{ChronosX}} & \rotatebox{45}{\thead{ChronosX (NC)}} & \rotatebox{45}{\thead{ChronosX (RS)}} & \rotatebox{45}{\thead{Chronos Small}} & \rotatebox{45}{\thead{Moirai (Base)}} & \rotatebox{45}{\thead{Moirai (Large)}} & \rotatebox{45}{\thead{AutoARIMA}} & \rotatebox{45}{\thead{TFT}} & \rotatebox{45}{\thead{DeepAR}} & \rotatebox{45}{\thead{PatchTSTx}} & \rotatebox{45}{\thead{NBEATSx-G}} & \rotatebox{45}{\thead{NBEATSx-I}} & \rotatebox{45}{\thead{NHiTS}} & \rotatebox{45}{\thead{Seasonal Naive}} \\
\midrule
ETT (15 Min.) & 0.022 & 0.022 & \underline{0.016} & 0.018 & \bftable 0.015 & 0.016 & 0.019 & 0.040 & 0.020 & 0.020 & 0.072 & 0.035 & 0.016 & 0.027 & \bftable \underline{0.015} & 0.065 \\
ETT (Hourly) & 0.025 & 0.024 & 0.025 & \bftable \underline{0.022} & \underline{0.024} & \bftable 0.022 & 0.037 & 0.076 & 0.045 & 0.045 & 0.026 & 0.036 & 0.065 & 0.063 & 0.035 & 0.101 \\
M5 & \bftable 0.538 & \bftable \underline{0.537} & 0.557 & 0.552 & 0.551 & \underline{0.549} & 0.586 & 0.576 & - & 0.560 & 0.617 & 0.812 & 0.607 & 0.604 & 0.748 & 0.934 \\
Electricity-BE & 0.058 & 0.059 & 0.049 & 0.051 & 0.047 & 0.049 & 0.069 & 0.075 & 0.080 & 0.087 & 0.078 & \underline{0.040} & \bftable \underline{0.036} & 0.054 & \bftable 0.038 & 0.071 \\
Electricity-DE & 0.455 & 0.553 & \bftable 0.219 & 0.243 & 0.245 & 0.315 & 0.457 & 0.414 & 0.372 & 0.374 & \underline{0.235} & 0.304 & \bftable \underline{0.195} & 0.378 & 0.295 & 1.048 \\
Electricity-FR & 0.074 & 0.071 & 0.062 & 0.062 & 0.059 & 0.060 & 0.083 & \underline{0.058} & 0.061 & 0.076 & 0.100 & \bftable 0.048 & 0.061 & 0.076 & \bftable \underline{0.041} & 0.135 \\
\bottomrule
\end{tabular}

    }
\end{table}
\begin{table}[H]
    \centering
    \caption{MASE scores per model in real datasets with covariates. Models achieving the \underline{\textbf{first}}, \textbf{second}, and \underline{third} best scores have been highlighted. Scores reported are averaged over three random seeds.}
    \label{tab:appendix:ciptfinal_real_without_covariates_aistats_mase}
    \rowcolors{2}{gray!25}{white}
    \renewcommand\theadfont{}
    \resizebox{\textwidth}{!}{    \begin{tabular}{lcccccccccccccccc}
\toprule
 & \multicolumn{2}{c}{Full Finetuned} & \multicolumn{3}{c}{Adapters Only} & \multicolumn{3}{c}{Pretrained Models} & \multicolumn{8}{c}{Baselines} \\
 \cmidrule(lr){2-3} \cmidrule(lr){4-6} \cmidrule(lr){7-9} \cmidrule(lr){10-17}
\rowcolor{white} & \rotatebox{45}{\thead{ChronosX (FF)}} & \rotatebox{45}{\thead{ChronosX (FF) (NC)}} & \rotatebox{45}{\thead{ChronosX}} & \rotatebox{45}{\thead{ChronosX (NC)}} & \rotatebox{45}{\thead{ChronosX (RS)}} & \rotatebox{45}{\thead{Chronos Small}} & \rotatebox{45}{\thead{Moirai (Base)}} & \rotatebox{45}{\thead{Moirai (Large)}} & \rotatebox{45}{\thead{AutoARIMA}} & \rotatebox{45}{\thead{TFT}} & \rotatebox{45}{\thead{DeepAR}} & \rotatebox{45}{\thead{PatchTSTx}} & \rotatebox{45}{\thead{NBEATSx-G}} & \rotatebox{45}{\thead{NBEATSx-I}} & \rotatebox{45}{\thead{NHiTS}} & \rotatebox{45}{\thead{Seasonal Naive}} \\
\midrule
ETT (15 Min.) & 0.338 & 0.327 & 0.213 & \underline{0.210} & \bftable \underline{0.206} & 0.224 & 0.259 & 0.513 & 0.225 & 0.269 & 0.790 & 0.451 & \bftable 0.208 & 0.404 & 0.239 & 0.677 \\
ETT (Hourly) & 0.354 & 0.343 & 0.352 & \bftable \underline{0.294} & \underline{0.338} & \bftable 0.297 & 0.489 & 0.900 & 0.469 & 0.599 & 0.351 & 0.463 & 0.752 & 0.677 & 0.443 & 0.950 \\
M5 & \bftable 0.907 & \bftable \underline{0.905} & 0.925 & 0.930 & 0.918 & 0.919 & 0.959 & 0.925 & - & \underline{0.911} & 0.938 & 1.038 & 0.946 & 0.953 & 1.326 & 1.268 \\
Electricity-BE & 0.432 & 0.454 & 0.374 & 0.382 & 0.372 & 0.380 & 0.458 & 0.510 & 0.306 & 0.621 & 0.568 & \underline{0.292} & \bftable \underline{0.276} & 0.406 & \bftable 0.276 & 0.405 \\
Electricity-DE & 1.001 & 1.208 & \bftable \underline{0.378} & 0.473 & 0.466 & 0.636 & 0.970 & 0.820 & 0.699 & 0.671 & \underline{0.458} & 0.625 & \bftable 0.381 & 0.646 & 0.526 & 1.582 \\
Electricity-FR & 0.676 & 0.640 & 0.570 & 0.558 & 0.537 & 0.548 & 0.801 & \underline{0.520} & 0.532 & 0.691 & 0.881 & \bftable 0.451 & 0.562 & 0.628 & \bftable \underline{0.382} & 0.966 \\
\bottomrule
\end{tabular}

    }
\end{table}

\newpage\clearpage
\subsection{Ablation: performance with larger version of Chronos}
\label{appendix:ablation_larger_versions_of_chronos}

In this section we present results where \thismodel is based in Chronos-Small, Chronos-Base and Chronos-Large pretrained models.

\begin{figure}[H]
    \centering
    \begin{subfigure}{.325\linewidth}
        \includegraphics[width=1\linewidth]{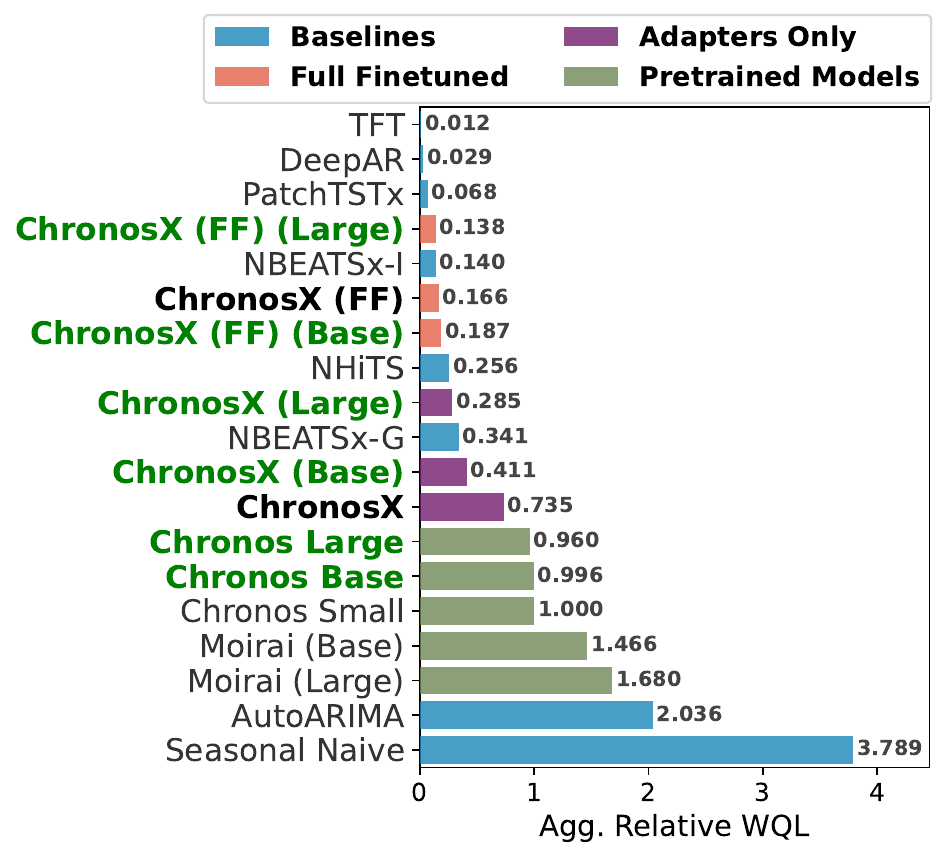}
        \caption{WQL on \textbf{Simple} Synthetic Data.}
        \label{fig:datasets_synthetic_simple_wql_chronosxbaselarge_appendix}
    \end{subfigure}
    \begin{subfigure}{.325\linewidth}
        \includegraphics[width=1\linewidth]{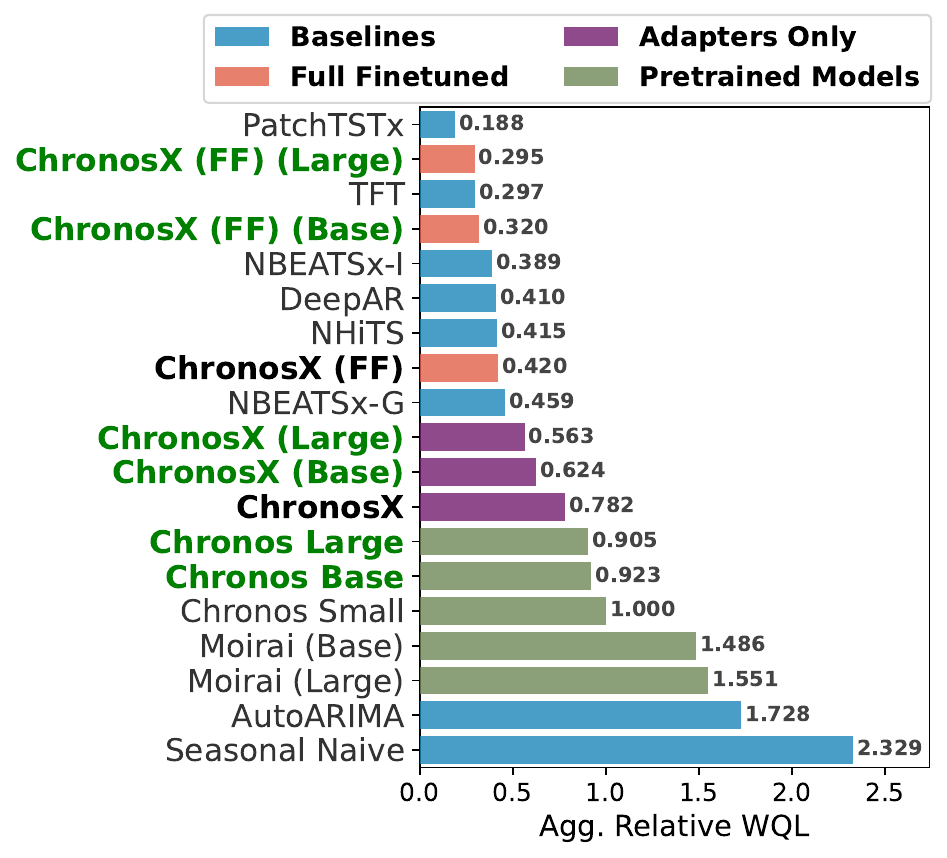}
        \caption{WQL on \textbf{Complex} Synthetic Data.}
        \label{fig:datasets_synthetic_complex_wql_chronosxbaselarge_appendix}
    \end{subfigure}
\begin{subfigure}{.325\linewidth}
    \includegraphics[width=1\linewidth]{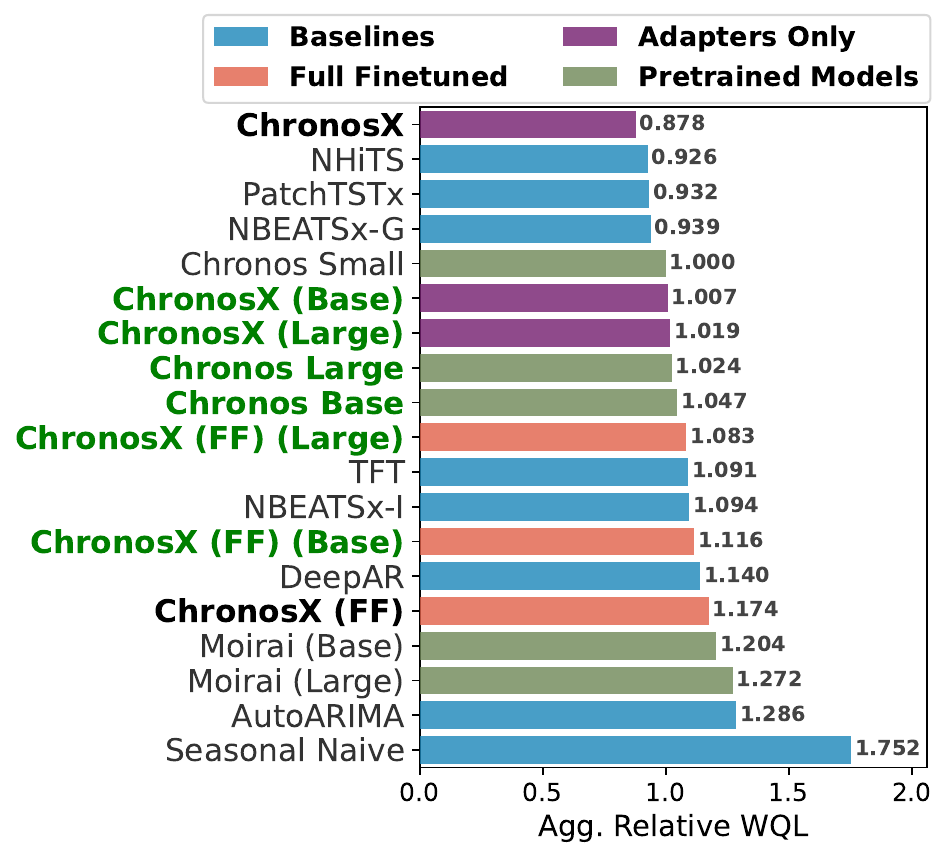}
    \caption{WQL  \textbf{Real}  Datasets}
    \label{fig:datasets_synthetic_real_wql_chronosxbaselarge_appendix}
\end{subfigure}
    \begin{subfigure}{.325\linewidth}
        \includegraphics[width=1\linewidth]{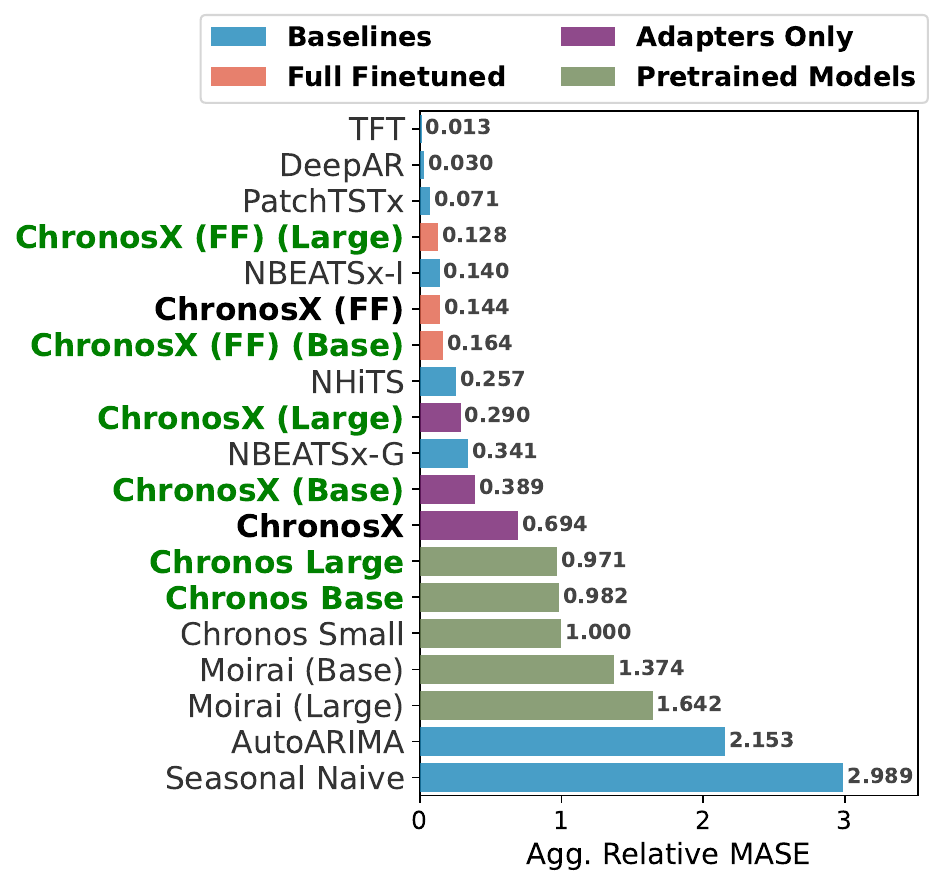}
        \caption{MASE on \textbf{Simple} Synthetic Data.}
        \label{fig:datasets_synthetic_simple_mase_chronosxbaselarge_appendix}
    \end{subfigure}
    \begin{subfigure}{.325\linewidth}
        \includegraphics[width=1\linewidth]{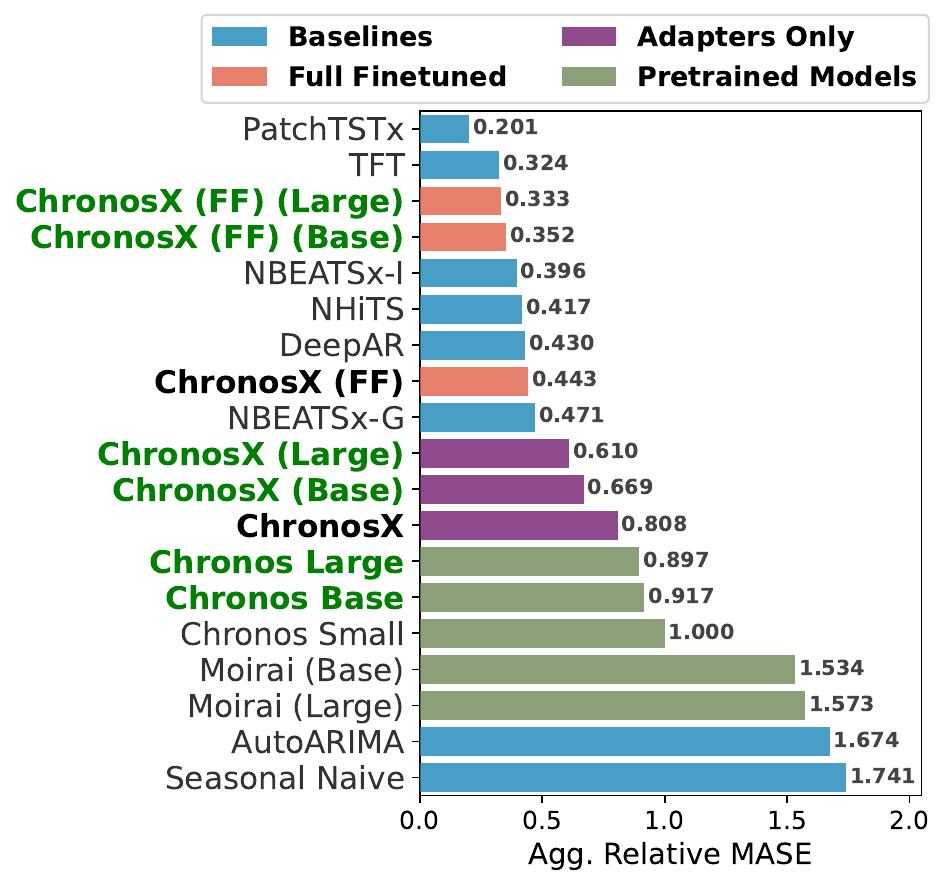}
        \caption{MASE on \textbf{Complex} Synthetic Data.}
        \label{fig:datasets_synthetic_complex_mase_chronosxbaselarge_appendix}
    \end{subfigure}
\begin{subfigure}{.325\linewidth}
    \includegraphics[width=1\linewidth]{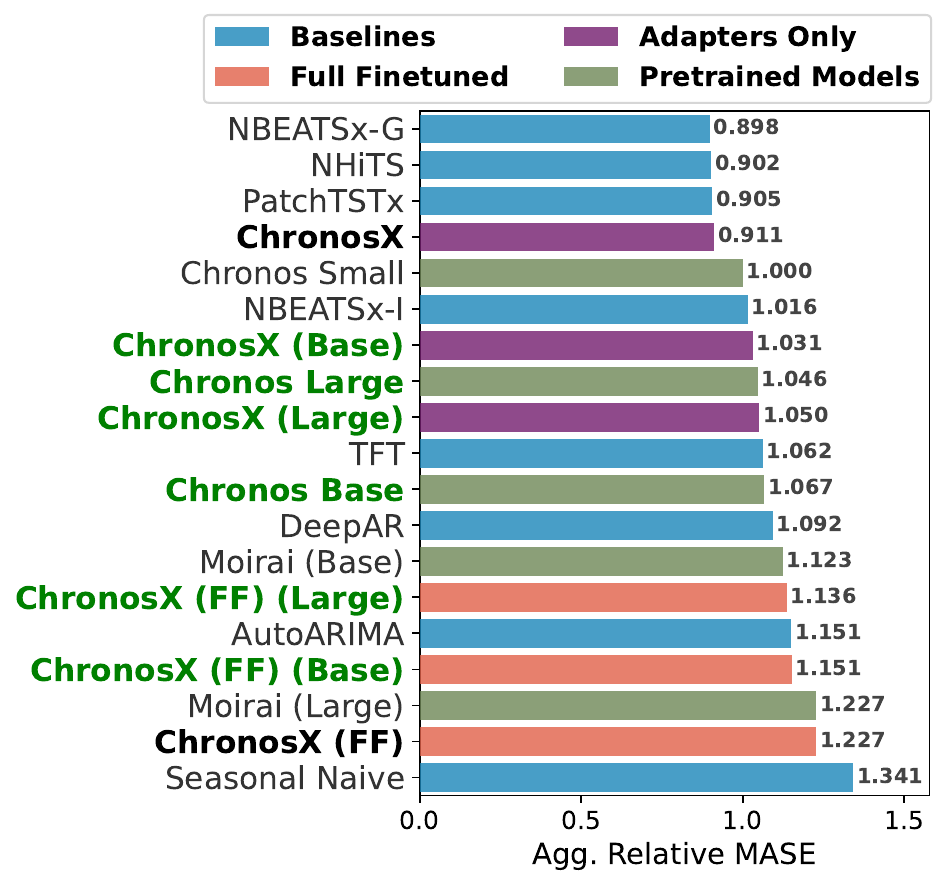}
    \caption{MASE on \textbf{Real}  Datasets}
    \label{fig:datasets_synthetic_real_mase_chronosxbaselarge_appendix}
\end{subfigure}
    \caption{
        Ablations of \textbf{ChronosX} and \textbf{ChronosX (FF)} with larger versions of \thatmodel.
    }
    \label{fig:chronosx_model_sizes}
\end{figure}

We present the following results:
\begin{itemize}[leftmargin=15pt]
    \item In Fig.~\ref{fig:chronosx_model_sizes} we present the Aggregated Relative WQL and MASE on simple, complex, and real datasets, with \thismodel on Chronos-Small, Chronos-Base and Chronos-Large model versions.
    \item In~\Cref{tab:appendix:ciptfinal_simple_synthetic_ChronosXBaseLarge_aistats_supp_wql} and~\Cref{tab:appendix:ciptfinal_complex_synthetic_ChronosXBaseLarge_aistats_supp_wql} we present the  WQL on simple and complex datasets, respectively,
    \item In~\Cref{tab:appendix:ciptfinal_simple_synthetic_ChronosXBaseLarge_aistats_supp_mase} and~\Cref{tab:appendix:ciptfinal_complex_synthetic_ChronosXBaseLarge_aistats_supp_mase} we present the MASE on simple datasets, respectively.
    \item In~\Cref{tab:appendix:ciptfinal_real_ChronosXBaseLarge_aistats_supp_wql} and~\Cref{tab:appendix:ciptfinal_real_ChronosXBaseLarge_aistats_supp_mase} we present WQL and MASE on real datasets, respectively.
\end{itemize}

\newpage\clearpage
\begin{table}[H]
    \centering
    \caption{WQL scores per model in 16 \textbf{simple} synthetic datasets. Models achieving the \underline{\textbf{first}}, \textbf{second}, and \underline{third} best scores have been highlighted. Scores reported are averaged over three random seeds.}
    \label{tab:appendix:ciptfinal_simple_synthetic_ChronosXBaseLarge_aistats_supp_wql}
    \rowcolors{2}{gray!25}{white}
    \renewcommand\theadfont{}
    \resizebox{\textwidth}{!}{%
\begin{tabular}{lccccccccccccccccccc}
\toprule
 & \multicolumn{3}{c}{Full Finetuned} & \multicolumn{3}{c}{Adapters Only} & \multicolumn{5}{c}{Pretrained Models} & \multicolumn{8}{c}{Baselines} \\
 \cmidrule(lr){2-4} \cmidrule(lr){5-7} \cmidrule(lr){8-12} \cmidrule(lr){13-20}
\rowcolor{white} & \rotatebox{45}{\thead{ChronosX (FF)}} & \rotatebox{45}{\thead{ChronosX (FF) (Base)}} & \rotatebox{45}{\thead{ChronosX (FF) (Large)}} & \rotatebox{45}{\thead{ChronosX}} & \rotatebox{45}{\thead{ChronosX (Base)}} & \rotatebox{45}{\thead{ChronosX (Large)}} & \rotatebox{45}{\thead{Chronos Small}} & \rotatebox{45}{\thead{Chronos Base}} & \rotatebox{45}{\thead{Chronos Large}} & \rotatebox{45}{\thead{Moirai (Large)}} & \rotatebox{45}{\thead{Moirai (Base)}} & \rotatebox{45}{\thead{NHiTS}} & \rotatebox{45}{\thead{AutoARIMA}} & \rotatebox{45}{\thead{TFT}} & \rotatebox{45}{\thead{NBEATSx-I}} & \rotatebox{45}{\thead{NBEATSx-G}} & \rotatebox{45}{\thead{DeepAR}} & \rotatebox{45}{\thead{PatchTSTx}} & \rotatebox{45}{\thead{Seasonal Naive}} \\
\midrule
Simple ARP (Add) & 0.164 & 0.545 & 0.043 & 0.168 & 0.079 & 0.052 & 0.056 & 0.057 & 0.054 & 0.148 & 0.099 & 0.012 & 0.198 & 0.007 & \bftable 0.005 & 0.015 & \underline{0.007} & \bftable \underline{0.002} & 0.402 \\
Simple ARP (Mult.) & 0.042 & 0.020 & 0.021 & 0.136 & 0.097 & 0.042 & 0.037 & 0.039 & 0.038 & 0.153 & 0.114 & 0.014 & 0.313 & \bftable \underline{0.001} & 0.012 & 0.018 & \underline{0.004} & \bftable 0.003 & 0.572 \\
Simple Bells (Add) & 0.287 & 0.281 & 0.283 & 0.225 & 0.297 & 0.274 & 0.319 & 0.307 & 0.296 & 0.383 & 0.346 & \underline{0.037} & 0.176 & \bftable 0.005 & 0.048 & 0.061 & 0.056 & \bftable \underline{0.003} & 0.582 \\
Simple Bells (Mult.) & 0.041 & 0.081 & 0.052 & 0.176 & 0.183 & 0.086 & 0.089 & 0.089 & 0.090 & 0.156 & 0.136 & 0.033 & 0.273 & \bftable \underline{0.001} & 0.040 & 0.037 & \underline{0.005} & \bftable 0.004 & 0.417 \\
\bottomrule
\end{tabular}

    }
\end{table}
\begin{table}[H]
    \centering
    \caption{WQL scores per model in 16 \textbf{complex} synthetic datasets. Models achieving the \underline{\textbf{first}}, \textbf{second}, and \underline{third} best scores have been highlighted. Scores reported are averaged over three random seeds.asdf}
    \label{tab:appendix:ciptfinal_complex_synthetic_ChronosXBaseLarge_aistats_supp_wql}
    \rowcolors{2}{gray!25}{white}
    \renewcommand\theadfont{}
    \resizebox{\textwidth}{!}{%
    \begin{tabular}{lccccccccccccccccccc}
\toprule
 & \multicolumn{3}{c}{Full Finetuned} & \multicolumn{3}{c}{Adapters Only} & \multicolumn{5}{c}{Pretrained Models} & \multicolumn{8}{c}{Baselines} \\
 \cmidrule(lr){2-4} \cmidrule(lr){5-7} \cmidrule(lr){8-12} \cmidrule(lr){13-20}
\rowcolor{white} & \rotatebox{45}{\thead{ChronosX (FF)}} & \rotatebox{45}{\thead{ChronosX (FF) (Base)}} & \rotatebox{45}{\thead{ChronosX (FF) (Large)}} & \rotatebox{45}{\thead{ChronosX}} & \rotatebox{45}{\thead{ChronosX (Base)}} & \rotatebox{45}{\thead{ChronosX (Large)}} & \rotatebox{45}{\thead{Chronos Small}} & \rotatebox{45}{\thead{Chronos Base}} & \rotatebox{45}{\thead{Chronos Large}} & \rotatebox{45}{\thead{Moirai (Large)}} & \rotatebox{45}{\thead{Moirai (Base)}} & \rotatebox{45}{\thead{NHiTS}} & \rotatebox{45}{\thead{AutoARIMA}} & \rotatebox{45}{\thead{TFT}} & \rotatebox{45}{\thead{NBEATSx-I}} & \rotatebox{45}{\thead{NBEATSx-G}} & \rotatebox{45}{\thead{DeepAR}} & \rotatebox{45}{\thead{PatchTSTx}} & \rotatebox{45}{\thead{Seasonal Naive}} \\
\midrule
Diverse ARP (Add) & 0.092 & 0.037 & 0.036 & 0.117 & 0.052 & 0.047 & 0.063 & 0.052 & 0.047 & 0.167 & 0.138 & \underline{0.012} & 0.265 & 0.023 & \bftable 0.009 & 0.019 & 0.032 & \bftable \underline{0.007} & 0.250 \\
Diverse ARP (Mult.) & 0.036 & 0.030 & 0.033 & 0.077 & 0.062 & 0.041 & 0.059 & 0.044 & 0.042 & 0.153 & 0.137 & \underline{0.024} & 0.269 & 0.030 & \bftable 0.024 & 0.028 & 0.029 & \bftable \underline{0.016} & 0.221 \\
Diverse Bells (Add) & 0.208 & 0.215 & 0.221 & 0.271 & 0.243 & 0.235 & 0.257 & 0.259 & 0.250 & 0.317 & 0.313 & 0.124 & 0.168 & \bftable 0.027 & 1.162 & \underline{0.105} & 0.161 & \bftable \underline{0.011} & 0.464 \\
Diverse Bells (Mult.) & 0.051 & \underline{0.046} & 0.056 & 0.116 & 0.108 & 0.081 & 0.089 & 0.081 & 0.084 & 0.162 & 0.155 & 0.055 & 0.233 & \bftable 0.036 & 0.050 & 0.067 & 0.081 & \bftable \underline{0.012} & 0.213 \\
\bottomrule
\end{tabular}

    }
\end{table}

\begin{table}[H]
    \centering
    \caption{MASE scores per model in 16 \textbf{simple} synthetic datasets. Models achieving the \underline{\textbf{first}}, \textbf{second}, and \underline{third} best scores have been highlighted. Scores reported are averaged over three random seeds.}
    \label{tab:appendix:ciptfinal_simple_synthetic_ChronosXBaseLarge_aistats_supp_mase}
    \rowcolors{2}{gray!25}{white}
    \renewcommand\theadfont{}
    \resizebox{\textwidth}{!}{%
    \begin{tabular}{lccccccccccccccccccc}
\toprule
 & \multicolumn{3}{c}{Full Finetuned} & \multicolumn{3}{c}{Adapters Only} & \multicolumn{5}{c}{Pretrained Models} & \multicolumn{8}{c}{Baselines} \\
 \cmidrule(lr){2-4} \cmidrule(lr){5-7} \cmidrule(lr){8-12} \cmidrule(lr){13-20}
\rowcolor{white} & \rotatebox{45}{\thead{ChronosX (FF)}} & \rotatebox{45}{\thead{ChronosX (FF) (Base)}} & \rotatebox{45}{\thead{ChronosX (FF) (Large)}} & \rotatebox{45}{\thead{ChronosX}} & \rotatebox{45}{\thead{ChronosX (Base)}} & \rotatebox{45}{\thead{ChronosX (Large)}} & \rotatebox{45}{\thead{Chronos Small}} & \rotatebox{45}{\thead{Chronos Base}} & \rotatebox{45}{\thead{Chronos Large}} & \rotatebox{45}{\thead{Moirai (Large)}} & \rotatebox{45}{\thead{Moirai (Base)}} & \rotatebox{45}{\thead{NHiTS}} & \rotatebox{45}{\thead{AutoARIMA}} & \rotatebox{45}{\thead{TFT}} & \rotatebox{45}{\thead{NBEATSx-I}} & \rotatebox{45}{\thead{NBEATSx-G}} & \rotatebox{45}{\thead{DeepAR}} & \rotatebox{45}{\thead{PatchTSTx}} & \rotatebox{45}{\thead{Seasonal Naive}} \\
\midrule
Simple ARP (Add) & 0.947 & 2.741 & 0.254 & 0.951 & 0.532 & 0.308 & 0.322 & 0.326 & 0.319 & 0.849 & 0.571 & 0.067 & 1.103 & 0.039 & \bftable 0.029 & 0.086 & \underline{0.034} & \bftable \underline{0.013} & 1.788 \\
Simple ARP (Mult.) & 0.161 & 0.091 & 0.095 & 0.500 & 0.362 & 0.175 & 0.157 & 0.172 & 0.172 & 0.642 & 0.454 & 0.057 & 1.616 & \bftable \underline{0.006} & 0.051 & 0.070 & \underline{0.020} & \bftable 0.014 & 1.930 \\
Simple Bells (Add) & 3.203 & 3.143 & 3.171 & 2.459 & 3.275 & 3.119 & 3.483 & 3.353 & 3.328 & 4.069 & 3.689 & \underline{0.392} & 1.993 & \bftable 0.055 & 0.491 & 0.641 & 0.613 & \bftable \underline{0.035} & 4.951 \\
Simple Bells (Mult.) & 0.220 & 0.409 & 0.312 & 0.858 & 0.908 & 0.457 & 0.473 & 0.494 & 0.497 & 0.847 & 0.691 & 0.177 & 1.698 & \bftable \underline{0.004} & 0.212 & 0.206 & \underline{0.025} & \bftable 0.019 & 1.882 \\
\bottomrule
\end{tabular}

    }
\end{table}
\begin{table}[H]
    \centering
    \caption{MASE scores per model in 16 \textbf{complex} synthetic datasets. Models achieving the \underline{\textbf{first}}, \textbf{second}, and \underline{third} best scores have been highlighted. Scores reported are averaged over three random seeds.}
    \label{tab:appendix:ciptfinal_complex_synthetic_ChronosXBaseLarge_aistats_supp_mase}
    \rowcolors{2}{gray!25}{white}
    \renewcommand\theadfont{}
    \resizebox{\textwidth}{!}{%
    \begin{tabular}{lccccccccccccccccccc}
\toprule
 & \multicolumn{3}{c}{Full Finetuned} & \multicolumn{3}{c}{Adapters Only} & \multicolumn{5}{c}{Pretrained Models} & \multicolumn{8}{c}{Baselines} \\
 \cmidrule(lr){2-4} \cmidrule(lr){5-7} \cmidrule(lr){8-12} \cmidrule(lr){13-20}
\rowcolor{white} & \rotatebox{45}{\thead{ChronosX (FF)}} & \rotatebox{45}{\thead{ChronosX (FF) (Base)}} & \rotatebox{45}{\thead{ChronosX (FF) (Large)}} & \rotatebox{45}{\thead{ChronosX}} & \rotatebox{45}{\thead{ChronosX (Base)}} & \rotatebox{45}{\thead{ChronosX (Large)}} & \rotatebox{45}{\thead{Chronos Small}} & \rotatebox{45}{\thead{Chronos Base}} & \rotatebox{45}{\thead{Chronos Large}} & \rotatebox{45}{\thead{Moirai (Large)}} & \rotatebox{45}{\thead{Moirai (Base)}} & \rotatebox{45}{\thead{NHiTS}} & \rotatebox{45}{\thead{AutoARIMA}} & \rotatebox{45}{\thead{TFT}} & \rotatebox{45}{\thead{NBEATSx-I}} & \rotatebox{45}{\thead{NBEATSx-G}} & \rotatebox{45}{\thead{DeepAR}} & \rotatebox{45}{\thead{PatchTSTx}} & \rotatebox{45}{\thead{Seasonal Naive}} \\
\midrule
Diverse ARP (Add) & 1.004 & 0.491 & 0.505 & 1.282 & 0.657 & 0.566 & 0.768 & 0.645 & 0.566 & 2.056 & 1.848 & \underline{0.148} & 2.801 & 0.317 & \bftable 0.110 & 0.242 & 0.419 & \bftable \underline{0.099} & 1.947 \\
Diverse ARP (Mult.) & 0.500 & 0.427 & 0.460 & 0.987 & 0.796 & 0.544 & 0.772 & 0.569 & 0.545 & 1.928 & 1.791 & \underline{0.323} & 3.091 & 0.404 & \bftable 0.314 & 0.374 & 0.370 & \bftable \underline{0.221} & 1.921 \\
Diverse Bells (Add) & 3.392 & 3.596 & 3.664 & 4.066 & 3.880 & 3.810 & 3.975 & 4.111 & 3.960 & 4.634 & 4.713 & \underline{1.512} & 2.234 & \bftable 0.446 & 12.561 & 1.584 & 2.513 & \bftable \underline{0.169} & 5.563 \\
Diverse Bells (Mult.) & 0.925 & \underline{0.911} & 1.136 & 1.959 & 1.786 & 1.443 & 1.620 & 1.419 & 1.478 & 2.834 & 2.755 & 1.000 & 3.453 & \bftable 0.651 & 0.920 & 1.171 & 1.462 & \bftable \underline{0.249} & 2.698 \\
\bottomrule
\end{tabular}

    }
\end{table}

\begin{table}[H]
    \centering
    \caption{WQL scores per model in real datasets with covariates. Models achieving the \underline{\textbf{first}}, \textbf{second}, and \underline{third} best scores have been highlighted. Scores reported are averaged over three random seeds.}
    \label{tab:appendix:ciptfinal_real_ChronosXBaseLarge_aistats_supp_wql}
    \rowcolors{2}{gray!25}{white}
    \renewcommand\theadfont{}
    \resizebox{\textwidth}{!}{    \begin{tabular}{lccccccccccccccccccc}
\toprule
 & \multicolumn{3}{c}{Full Finetuned} & \multicolumn{3}{c}{Adapters Only} & \multicolumn{5}{c}{Pretrained Models} & \multicolumn{8}{c}{Baselines} \\
 \cmidrule(lr){2-4} \cmidrule(lr){5-7} \cmidrule(lr){8-12} \cmidrule(lr){13-20}
\rowcolor{white} & \rotatebox{45}{\thead{ChronosX (FF)}} & \rotatebox{45}{\thead{ChronosX (FF) (Base)}} & \rotatebox{45}{\thead{ChronosX (FF) (Large)}} & \rotatebox{45}{\thead{ChronosX}} & \rotatebox{45}{\thead{ChronosX (Base)}} & \rotatebox{45}{\thead{ChronosX (Large)}} & \rotatebox{45}{\thead{Chronos Small}} & \rotatebox{45}{\thead{Chronos Base}} & \rotatebox{45}{\thead{Chronos Large}} & \rotatebox{45}{\thead{Moirai (Base)}} & \rotatebox{45}{\thead{Moirai (Large)}} & \rotatebox{45}{\thead{AutoARIMA}} & \rotatebox{45}{\thead{TFT}} & \rotatebox{45}{\thead{DeepAR}} & \rotatebox{45}{\thead{PatchTSTx}} & \rotatebox{45}{\thead{NBEATSx-G}} & \rotatebox{45}{\thead{NBEATSx-I}} & \rotatebox{45}{\thead{NHiTS}} & \rotatebox{45}{\thead{Seasonal Naive}} \\
\midrule
ETT (15 Min.) & 0.022 & 0.026 & 0.025 & \bftable 0.016 & 0.020 & 0.019 & 0.016 & 0.017 & 0.018 & 0.019 & 0.040 & 0.020 & 0.020 & 0.072 & 0.035 & \underline{0.016} & 0.027 & \bftable \underline{0.015} & 0.065 \\
ETT (Hourly) & \underline{0.025} & \bftable 0.024 & 0.027 & 0.025 & 0.026 & 0.026 & \bftable \underline{0.022} & 0.025 & 0.026 & 0.037 & 0.076 & 0.045 & 0.045 & 0.026 & 0.036 & 0.065 & 0.063 & 0.035 & 0.101 \\
M5 & \underline{0.538} & \bftable 0.533 & \bftable \underline{0.533} & 0.557 & 0.551 & 0.550 & 0.549 & 0.544 & 0.544 & 0.586 & 0.576 & - & 0.560 & 0.617 & 0.812 & 0.607 & 0.604 & 0.748 & 0.934 \\
Electricity-BE & 0.058 & 0.062 & 0.065 & 0.049 & 0.052 & 0.045 & 0.049 & 0.051 & 0.044 & 0.069 & 0.075 & 0.080 & 0.087 & 0.078 & \underline{0.040} & \bftable \underline{0.036} & 0.054 & \bftable 0.038 & 0.071 \\
\bottomrule
\end{tabular}

    }
\end{table}
\begin{table}[H]
    \centering
    \caption{MASE scores per model in real datasets with covariates. Models achieving the \underline{\textbf{first}}, \textbf{second}, and \underline{third} best scores have been highlighted. Scores reported are averaged over three random seeds.}
    \label{tab:appendix:ciptfinal_real_ChronosXBaseLarge_aistats_supp_mase}
    \rowcolors{2}{gray!25}{white}
    \renewcommand\theadfont{}
    \resizebox{\textwidth}{!}{    \begin{tabular}{lccccccccccccccccccc}
\toprule
 & \multicolumn{3}{c}{Full Finetuned} & \multicolumn{3}{c}{Adapters Only} & \multicolumn{5}{c}{Pretrained Models} & \multicolumn{8}{c}{Baselines} \\
 \cmidrule(lr){2-4} \cmidrule(lr){5-7} \cmidrule(lr){8-12} \cmidrule(lr){13-20}
\rowcolor{white} & \rotatebox{45}{\thead{ChronosX (FF)}} & \rotatebox{45}{\thead{ChronosX (FF) (Base)}} & \rotatebox{45}{\thead{ChronosX (FF) (Large)}} & \rotatebox{45}{\thead{ChronosX}} & \rotatebox{45}{\thead{ChronosX (Base)}} & \rotatebox{45}{\thead{ChronosX (Large)}} & \rotatebox{45}{\thead{Chronos Small}} & \rotatebox{45}{\thead{Chronos Base}} & \rotatebox{45}{\thead{Chronos Large}} & \rotatebox{45}{\thead{Moirai (Base)}} & \rotatebox{45}{\thead{Moirai (Large)}} & \rotatebox{45}{\thead{AutoARIMA}} & \rotatebox{45}{\thead{TFT}} & \rotatebox{45}{\thead{DeepAR}} & \rotatebox{45}{\thead{PatchTSTx}} & \rotatebox{45}{\thead{NBEATSx-G}} & \rotatebox{45}{\thead{NBEATSx-I}} & \rotatebox{45}{\thead{NHiTS}} & \rotatebox{45}{\thead{Seasonal Naive}} \\
\midrule
ETT (15 Min.) & 0.338 & 0.363 & 0.344 & \bftable 0.213 & 0.287 & 0.272 & \underline{0.224} & 0.242 & 0.243 & 0.259 & 0.513 & 0.225 & 0.269 & 0.790 & 0.451 & \bftable \underline{0.208} & 0.404 & 0.239 & 0.677 \\
ETT (Hourly) & 0.354 & \bftable 0.341 & 0.373 & 0.352 & 0.365 & 0.366 & \bftable \underline{0.297} & \underline{0.347} & 0.373 & 0.489 & 0.900 & 0.469 & 0.599 & 0.351 & 0.463 & 0.752 & 0.677 & 0.443 & 0.950 \\
M5 & \underline{0.907} & \bftable \underline{0.902} & \bftable 0.902 & 0.925 & 0.914 & 0.916 & 0.919 & 0.916 & 0.917 & 0.959 & 0.925 & - & 0.911 & 0.938 & 1.038 & 0.946 & 0.953 & 1.326 & 1.268 \\
Electricity-BE & 0.432 & 0.482 & 0.499 & 0.374 & 0.380 & 0.326 & 0.380 & 0.383 & 0.329 & 0.458 & 0.510 & 0.306 & 0.621 & 0.568 & \underline{0.292} & \bftable \underline{0.276} & 0.406 & \bftable 0.276 & 0.405 \\
\bottomrule
\end{tabular}

    }
\end{table}

\newpage\clearpage
\subsection{Ablation: performance with architecture of adapter architecture}
\label{appendix:ablation_adapter_arch}

In this section we present results where \thismodel has different adapter architectures for covariates.

\begin{figure}[H]
    \centering
    \begin{subfigure}{.325\linewidth}
        \includegraphics[width=1\linewidth]{Figures/ciptfinal_simple_synthetic_simpler_blocks_aistats_supp/aggregated_plot_chronos_style_WQL_sd.pdf}
        \caption{WQL on \textbf{Simple} Synthetic Data.}
        \label{fig:ciptfinal_simple_synthetic_simpler_blocks_aistats_supp_wql}
    \end{subfigure}
    \begin{subfigure}{.325\linewidth}
        \includegraphics[width=1\linewidth]{Figures/ciptfinal_complex_synthetic_simpler_blocks_aistats_supp/aggregated_plot_chronos_style_WQL_sd.pdf}
        \caption{WQL on \textbf{Complex} Synthetic Data.}
        \label{fig:ciptfinal_complex_synthetic_simpler_blocks_aistats_supp_wql}
    \end{subfigure}
\begin{subfigure}{.325\linewidth}
    \includegraphics[width=1\linewidth]{Figures/ciptfinal_real_simpler_blocks_aistats_supp/aggregated_plot_chronos_style_WQL_cd.pdf}
    \caption{WQL  \textbf{Real}  Datasets}
    \label{fig:ciptfinal_real_simpler_blocks_aistats_supp_wql}
\end{subfigure}
    \begin{subfigure}{.325\linewidth}
        \includegraphics[width=1\linewidth]{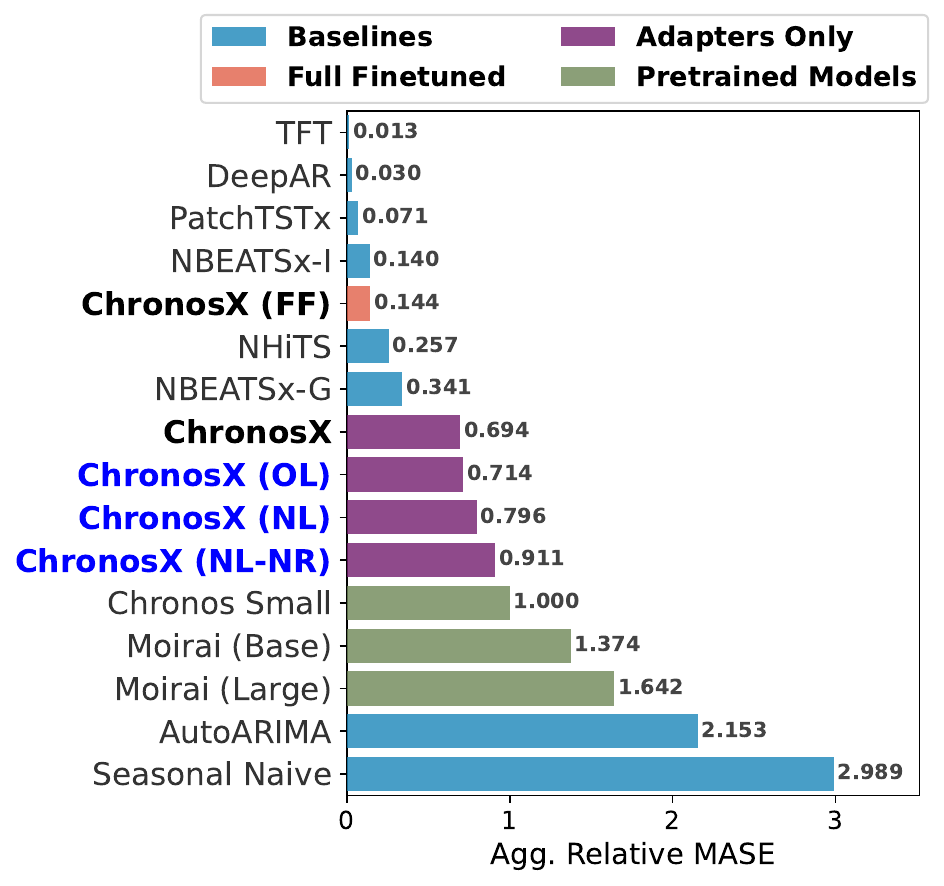}
        \caption{MASE on \textbf{Simple} Synthetic Data.}
        \label{fig:ciptfinal_simple_synthetic_simpler_blocks_aistats_supp_mase}
    \end{subfigure}
    \begin{subfigure}{.325\linewidth}
        \includegraphics[width=1\linewidth]{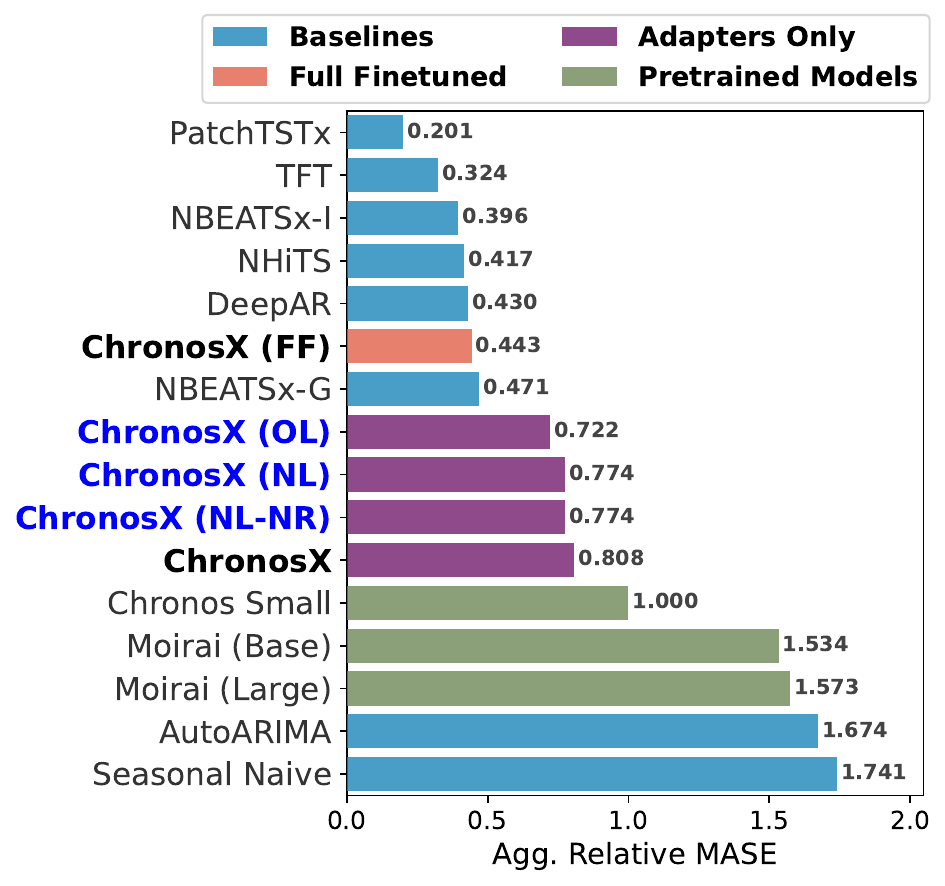}
        \caption{MASE on \textbf{Complex} Synthetic Data.}
        \label{fig:ciptfinal_complex_synthetic_simpler_blocks_aistats_supp_mase}
    \end{subfigure}
\begin{subfigure}{.325\linewidth}
    \includegraphics[width=1\linewidth]{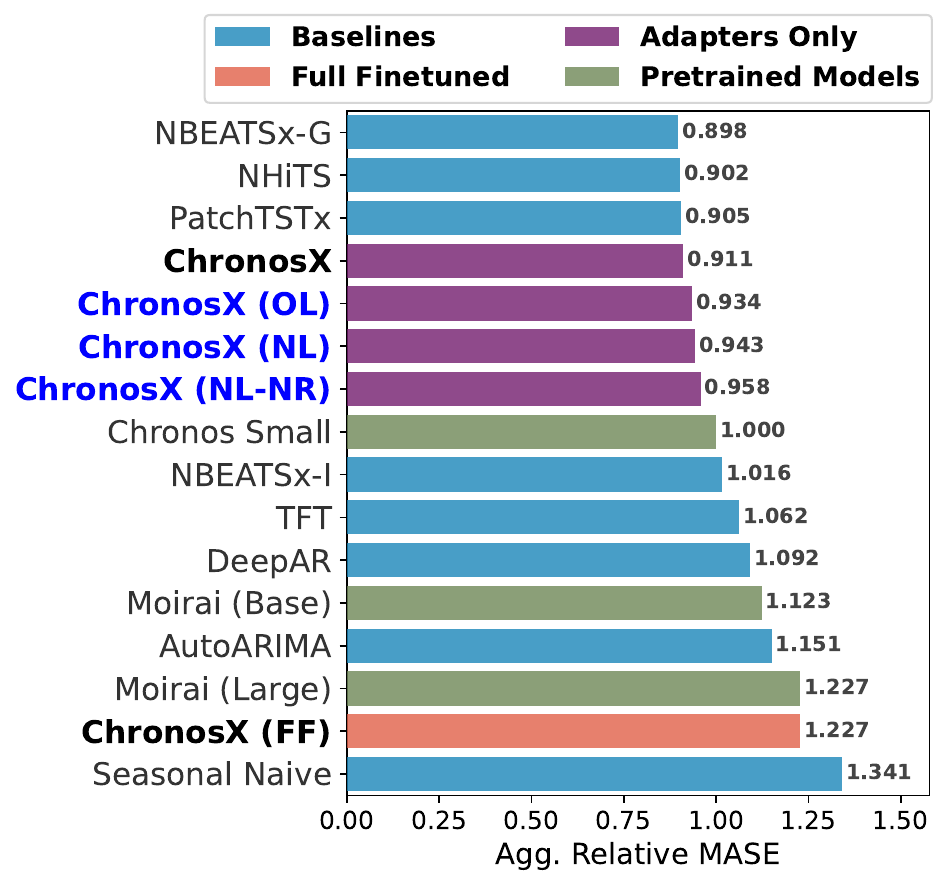}
    \caption{MASE on \textbf{Real}  Datasets}
    \label{fig:ciptfinal_real_simpler_blocks_aistats_supp_mase}
\end{subfigure}
    \caption{
        Ablations of \textbf{ChronosX} with simpler adapter blocks.
    }
    \label{fig:_chronosx_different_adapter_blocks}
\end{figure}

We present the following results:
\begin{itemize}[leftmargin=15pt]
    \item In Fig.~\ref{fig:_chronosx_different_adapter_blocks} we present the Aggregated Relative WQL and MASE on different versions of \thismodel with different adapter blocks.
    \item In~\Cref{tab:appendix:ciptfinal_simple_synthetic_simpler_blocks_aistats_supp_wql} and~\Cref{tab:appendix:ciptfinal_complex_synthetic_simpler_blocks_aistats_supp_wql} we present the  WQL on simple and complex datasets, respectively,
    \item In~\Cref{tab:appendix:ciptfinal_simple_synthetic_simpler_blocks_aistats_supp_mase} and~\Cref{tab:appendix:ciptfinal_complex_synthetic_simpler_blocks_aistats_supp_mase} we present the MASE on simple and complex datasets, respectively.
    \item In~\Cref{tab:appendix:ciptfinal_real_simpler_blocks_aistats_supp_wql} and~\Cref{tab:appendix:ciptfinal_real_simpler_blocks_aistats_supp_mase} we present WQL and MASE on real datasets, respectively.
\end{itemize}

\newpage\clearpage
\begin{table}[H]
    \centering
    \caption{WQL scores per model in 16 \textbf{simple} synthetic datasets. Models achieving the \underline{\textbf{first}}, \textbf{second}, and \underline{third} best scores have been highlighted. Scores reported are averaged over three random seeds.}
    \label{tab:appendix:ciptfinal_simple_synthetic_simpler_blocks_aistats_supp_wql}
    \rowcolors{2}{gray!25}{white}
    \renewcommand\theadfont{}
    \resizebox{\textwidth}{!}{%
\begin{tabular}{lcccccccccccccccc}
\toprule
 & Full Finetuned & \multicolumn{4}{c}{Adapters Only} & \multicolumn{3}{c}{Pretrained Models} & \multicolumn{8}{c}{Baselines} \\
 \cmidrule(lr){2-2} \cmidrule(lr){3-6} \cmidrule(lr){7-9} \cmidrule(lr){10-17}
\rowcolor{white} & \rotatebox{45}{\thead{ChronosX (FF)}} & \rotatebox{45}{\thead{ChronosX}} & \rotatebox{45}{\thead{ChronosX (OL)}} & \rotatebox{45}{\thead{ChronosX (NL)}} & \rotatebox{45}{\thead{ChronosX (NL-NR)}} & \rotatebox{45}{\thead{Chronos Small}} & \rotatebox{45}{\thead{Moirai (Large)}} & \rotatebox{45}{\thead{Moirai (Base)}} & \rotatebox{45}{\thead{NHiTS}} & \rotatebox{45}{\thead{AutoARIMA}} & \rotatebox{45}{\thead{TFT}} & \rotatebox{45}{\thead{NBEATSx-I}} & \rotatebox{45}{\thead{NBEATSx-G}} & \rotatebox{45}{\thead{DeepAR}} & \rotatebox{45}{\thead{PatchTSTx}} & \rotatebox{45}{\thead{Seasonal Naive}} \\
\midrule
Simple ARP (Add) & 0.164 & 0.168 & 0.089 & 0.088 & 0.066 & 0.056 & 0.148 & 0.099 & 0.012 & 0.198 & 0.007 & \bftable 0.005 & 0.015 & \underline{0.007} & \bftable \underline{0.002} & 0.402 \\
Simple ARP (Mult.) & 0.042 & 0.136 & 0.127 & 0.108 & 0.057 & 0.037 & 0.153 & 0.114 & 0.014 & 0.313 & \bftable \underline{0.001} & 0.012 & 0.018 & \underline{0.004} & \bftable 0.003 & 0.572 \\
Simple Bells (Add) & 0.287 & 0.225 & 0.302 & 0.342 & 0.347 & 0.319 & 0.383 & 0.346 & \underline{0.037} & 0.176 & \bftable 0.005 & 0.048 & 0.061 & 0.056 & \bftable \underline{0.003} & 0.582 \\
Simple Bells (Mult.) & 0.041 & 0.176 & 0.165 & 0.151 & 0.133 & 0.089 & 0.156 & 0.136 & 0.033 & 0.273 & \bftable \underline{0.001} & 0.040 & 0.037 & \underline{0.005} & \bftable 0.004 & 0.417 \\
Simple Spikes (Add) & 0.007 & 0.250 & 0.206 & 0.233 & 0.223 & 0.431 & 0.515 & 0.498 & 0.083 & 0.381 & \bftable \underline{0.003} & 0.036 & 0.117 & \bftable 0.003 & \underline{0.004} & 0.978 \\
\bottomrule
\end{tabular}

    }
\end{table}
\begin{table}[H]
    \centering
    \caption{WQL scores per model in 16 \textbf{complex} synthetic datasets. Models achieving the \underline{\textbf{first}}, \textbf{second}, and \underline{third} best scores have been highlighted. Scores reported are averaged over three random seeds.asdf}
    \label{tab:appendix:ciptfinal_complex_synthetic_simpler_blocks_aistats_supp_wql}
    \rowcolors{2}{gray!25}{white}
    \renewcommand\theadfont{}
    \resizebox{\textwidth}{!}{%
    %

    }
\end{table}

\begin{table}[H]
    \centering
    \caption{MASE scores per model in 16 \textbf{simple} synthetic datasets. Models achieving the \underline{\textbf{first}}, \textbf{second}, and \underline{third} best scores have been highlighted. Scores reported are averaged over three random seeds.}
    \label{tab:appendix:ciptfinal_simple_synthetic_simpler_blocks_aistats_supp_mase}
    \rowcolors{2}{gray!25}{white}
    \renewcommand\theadfont{}
    \resizebox{\textwidth}{!}{%
    %

    }
\end{table}
\begin{table}[H]
    \centering
    \caption{MASE scores per model in 16 \textbf{complex} synthetic datasets. Models achieving the \underline{\textbf{first}}, \textbf{second}, and \underline{third} best scores have been highlighted. Scores reported are averaged over three random seeds.}
    \label{tab:appendix:ciptfinal_complex_synthetic_simpler_blocks_aistats_supp_mase}
    \rowcolors{2}{gray!25}{white}
    \renewcommand\theadfont{}
    \resizebox{\textwidth}{!}{%
    %

    }
\end{table}

\begin{table}[H]
    \centering
    \caption{WQL scores per model in real datasets with covariates. Models achieving the \underline{\textbf{first}}, \textbf{second}, and \underline{third} best scores have been highlighted. Scores reported are averaged over three random seeds.}
    \label{tab:appendix:ciptfinal_real_simpler_blocks_aistats_supp_wql}
    \rowcolors{2}{gray!25}{white}
    \renewcommand\theadfont{}
    \resizebox{\textwidth}{!}{    %

    }
\end{table}
\begin{table}[H]
    \centering
    \caption{MASE scores per model in real datasets with covariates. Models achieving the \underline{\textbf{first}}, \textbf{second}, and \underline{third} best scores have been highlighted. Scores reported are averaged over three random seeds.}
    \label{tab:appendix:ciptfinal_real_simpler_blocks_aistats_supp_mase}
    \rowcolors{2}{gray!25}{white}
    \renewcommand\theadfont{}
    \resizebox{\textwidth}{!}{    %

    }
\end{table}

\newpage\clearpage
\subsection{Probabilistic Forecasts of Chronos Variants}
\label{sec::appendix:rq4}

\begin{figure}[h]
    \centering
    \includegraphics[trim={0 0 0 1cm},clip, width=1\linewidth]{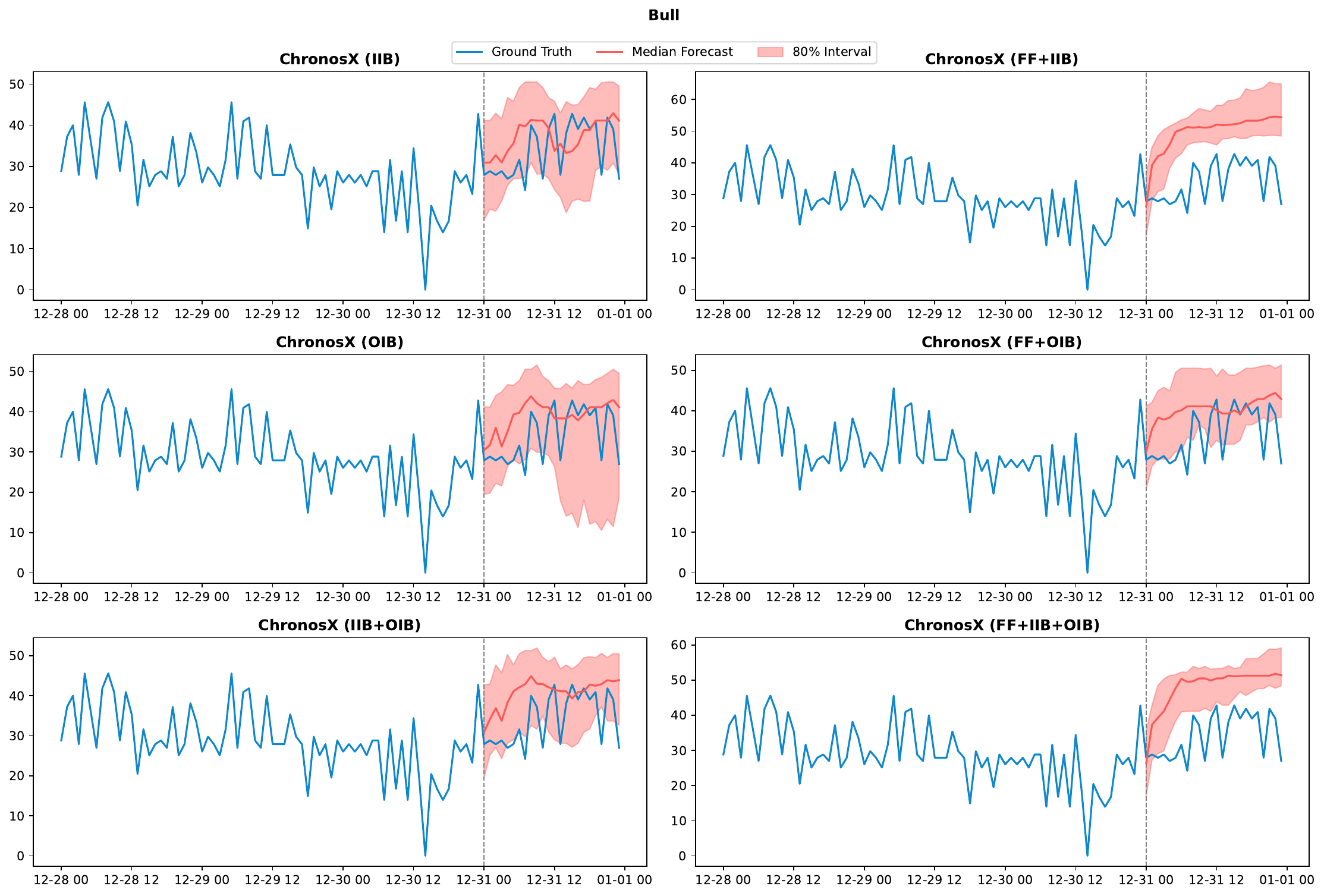}
    \caption{Forecasts on Bull Dataset}
    \label{fig:first_figure_probabilistic}
\end{figure}

\begin{figure}[h]
    \centering
    \includegraphics[trim={0 0 0 1cm},clip, width=1\linewidth]{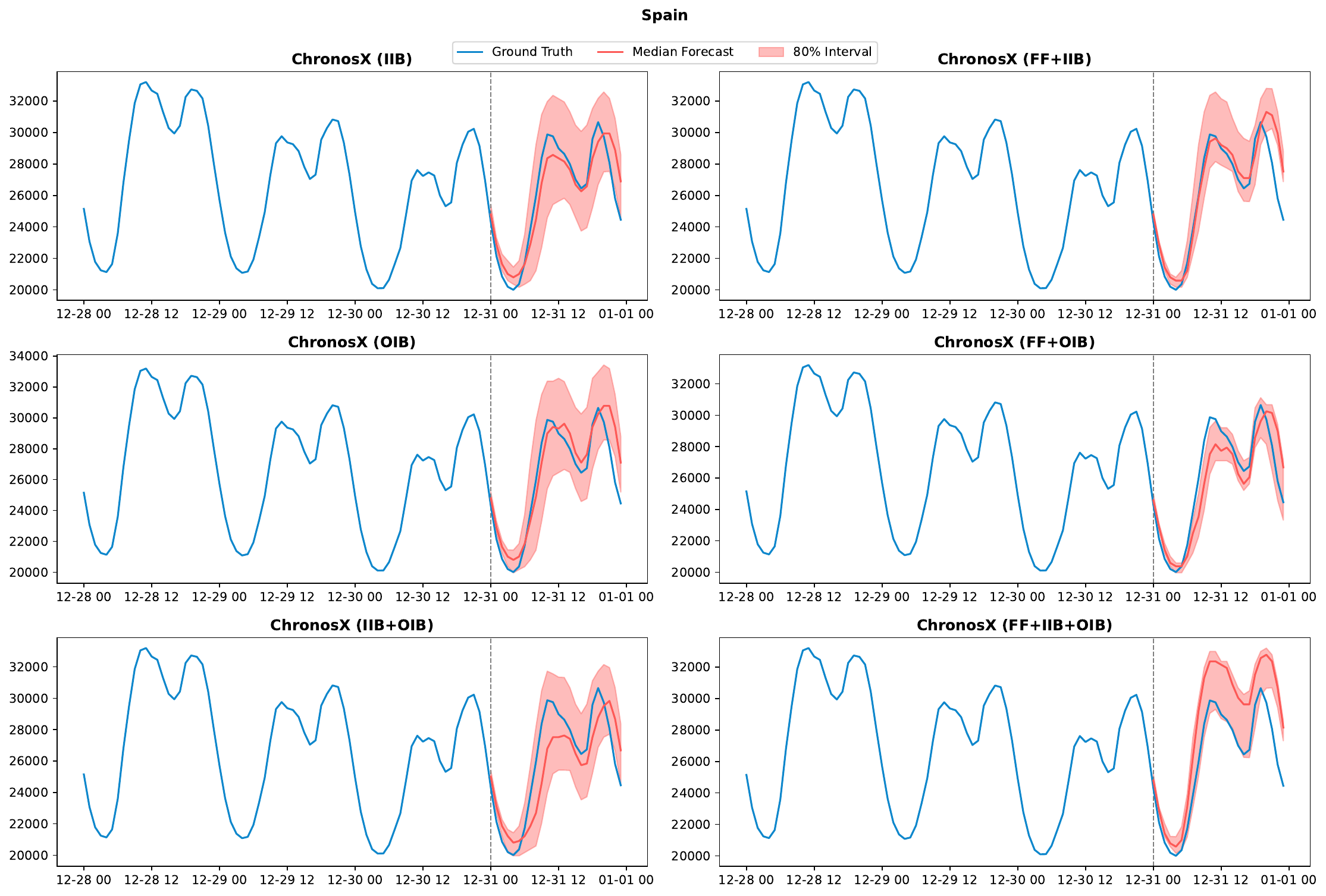}
    \caption{Forecasts on Spanish Dataset}
    \label{fig:last_figure_probabilistic}
\end{figure}

From Figure \ref{fig:first_figure_probabilistic} to \ref{fig:last_figure_probabilistic}, we plot the probabilistic forecast of different Chronos variants.

\begin{figure}[h]
    \centering
    \includegraphics[trim={0 0 0 1cm},clip, width=1\linewidth]{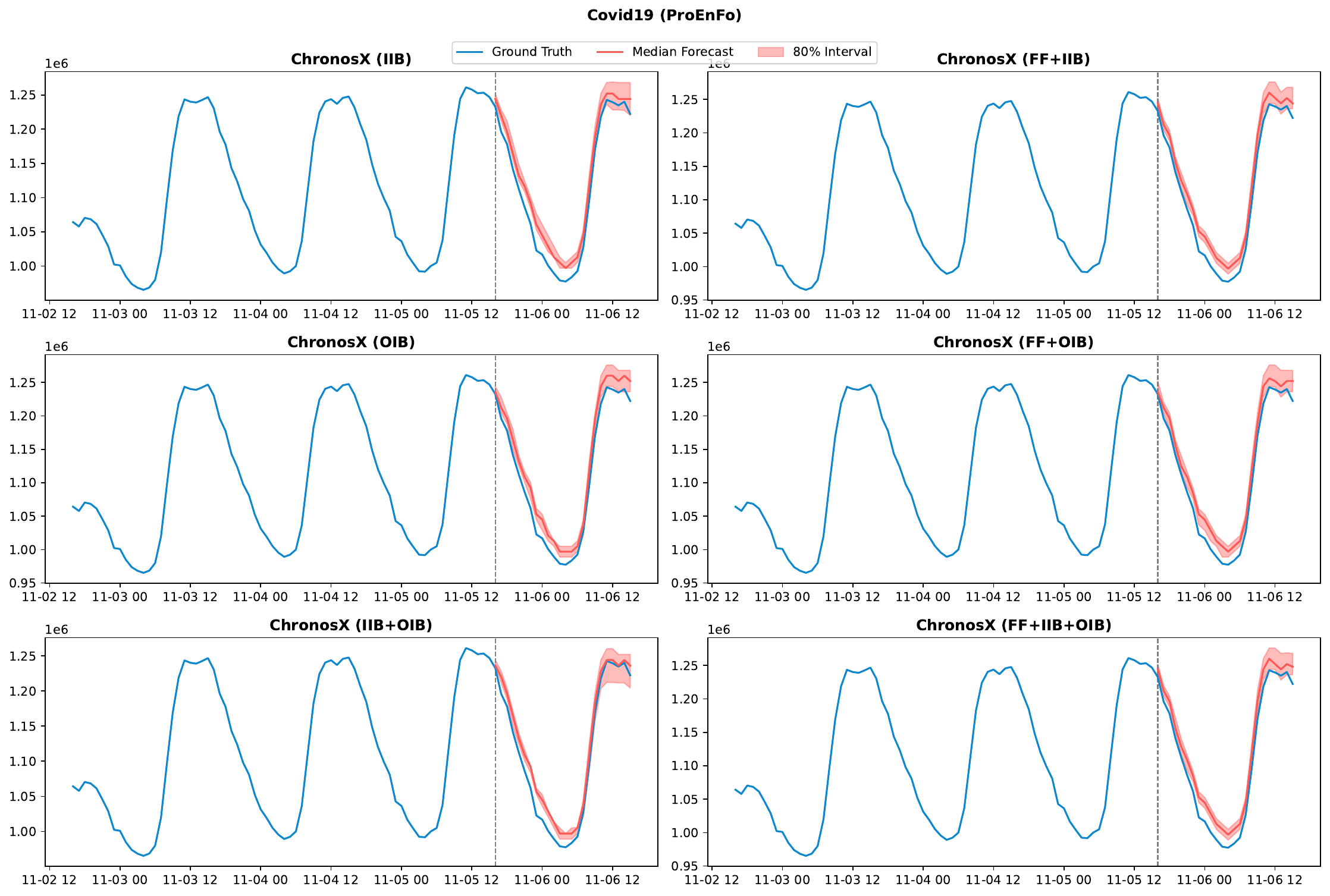}
    \caption{Forecasts on Covid19 Dataset}
\end{figure}
\begin{figure}[h]
    \centering
    \includegraphics[trim={0 0 0 1cm},clip, width=1\linewidth]{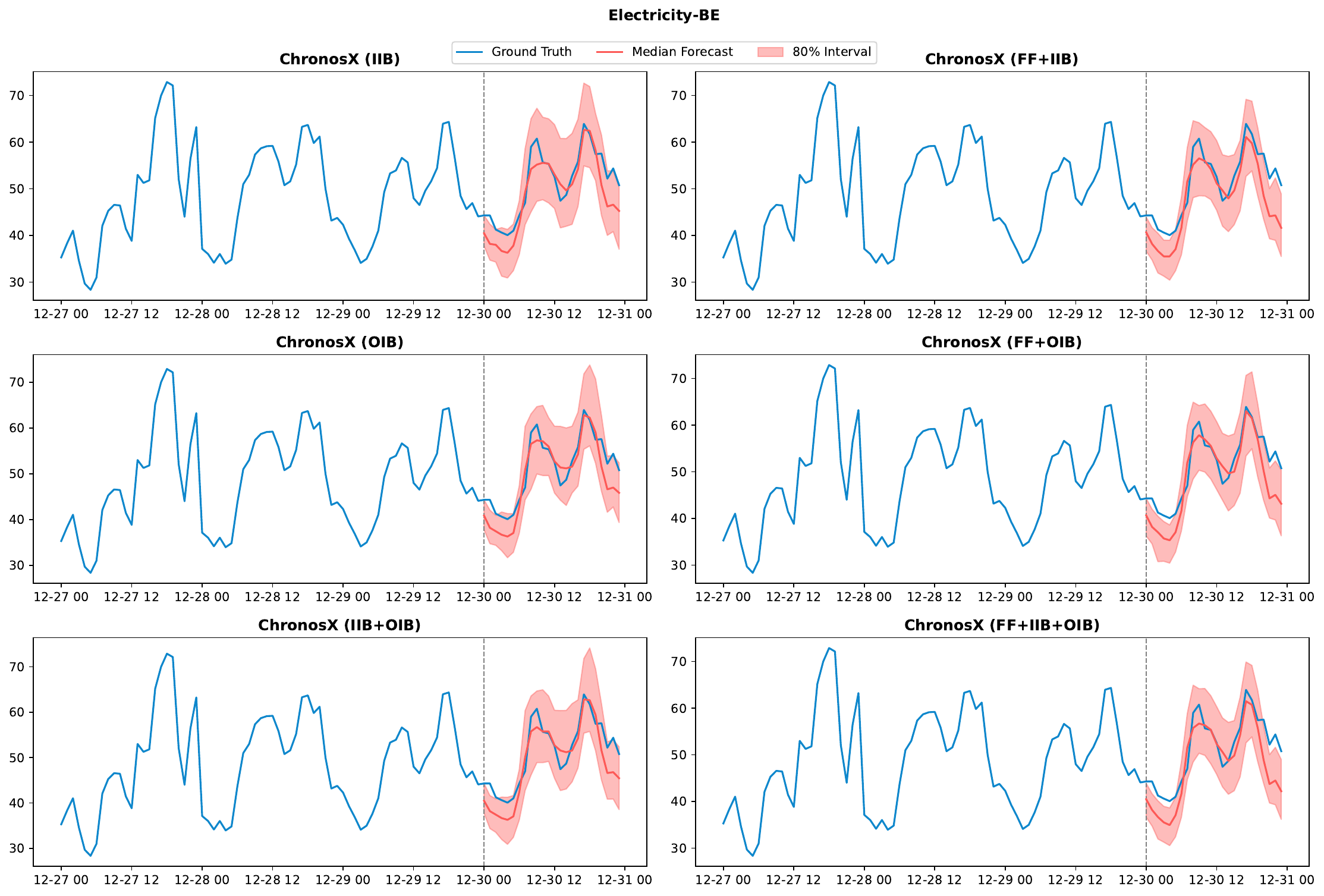}
    \caption{Forecasts on Electricity-BE Dataset}
\end{figure}
\begin{figure}[h]
    \centering
    \includegraphics[trim={0 0 0 1cm},clip, width=1\linewidth]{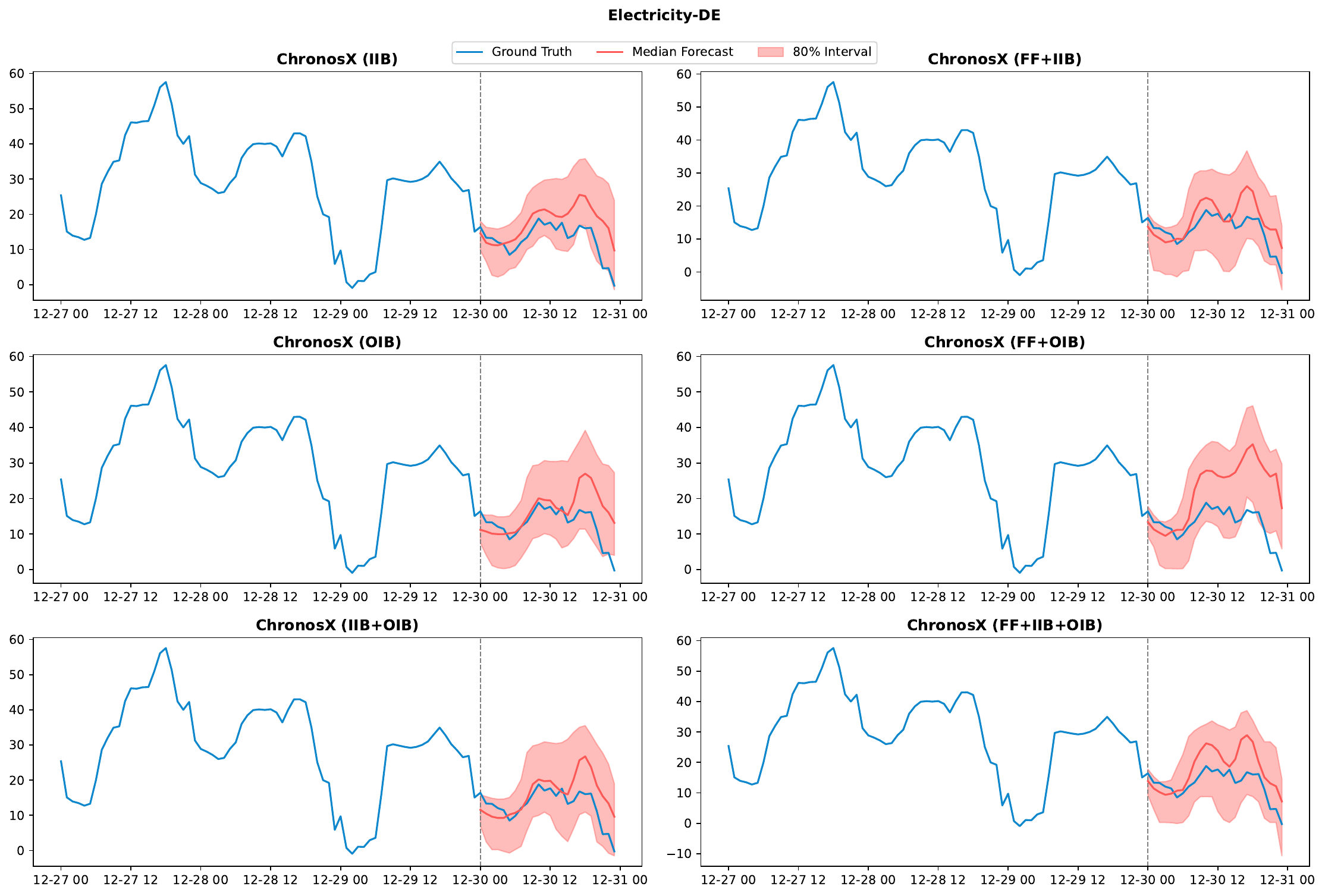}
    \caption{Forecasts on Electricity-DE Dataset}
\end{figure}
\begin{figure}[h]
    \centering
    \includegraphics[trim={0 0 0 1cm},clip, width=1\linewidth]{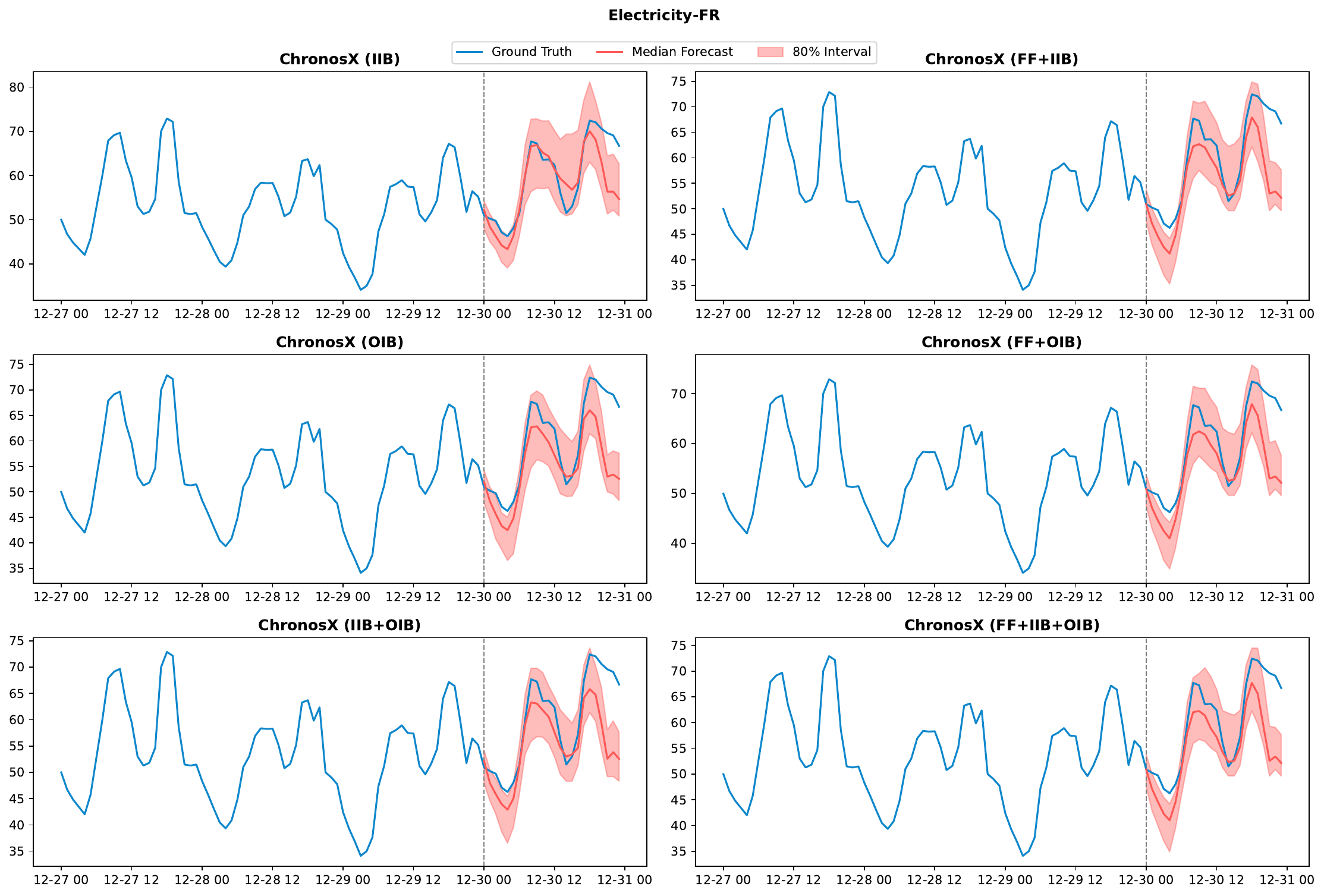}
    \caption{Forecasts on Electricity-FR Dataset}
\end{figure}
\begin{figure}[h]
    \centering
    \includegraphics[trim={0 0 0 1cm},clip, width=1\linewidth]{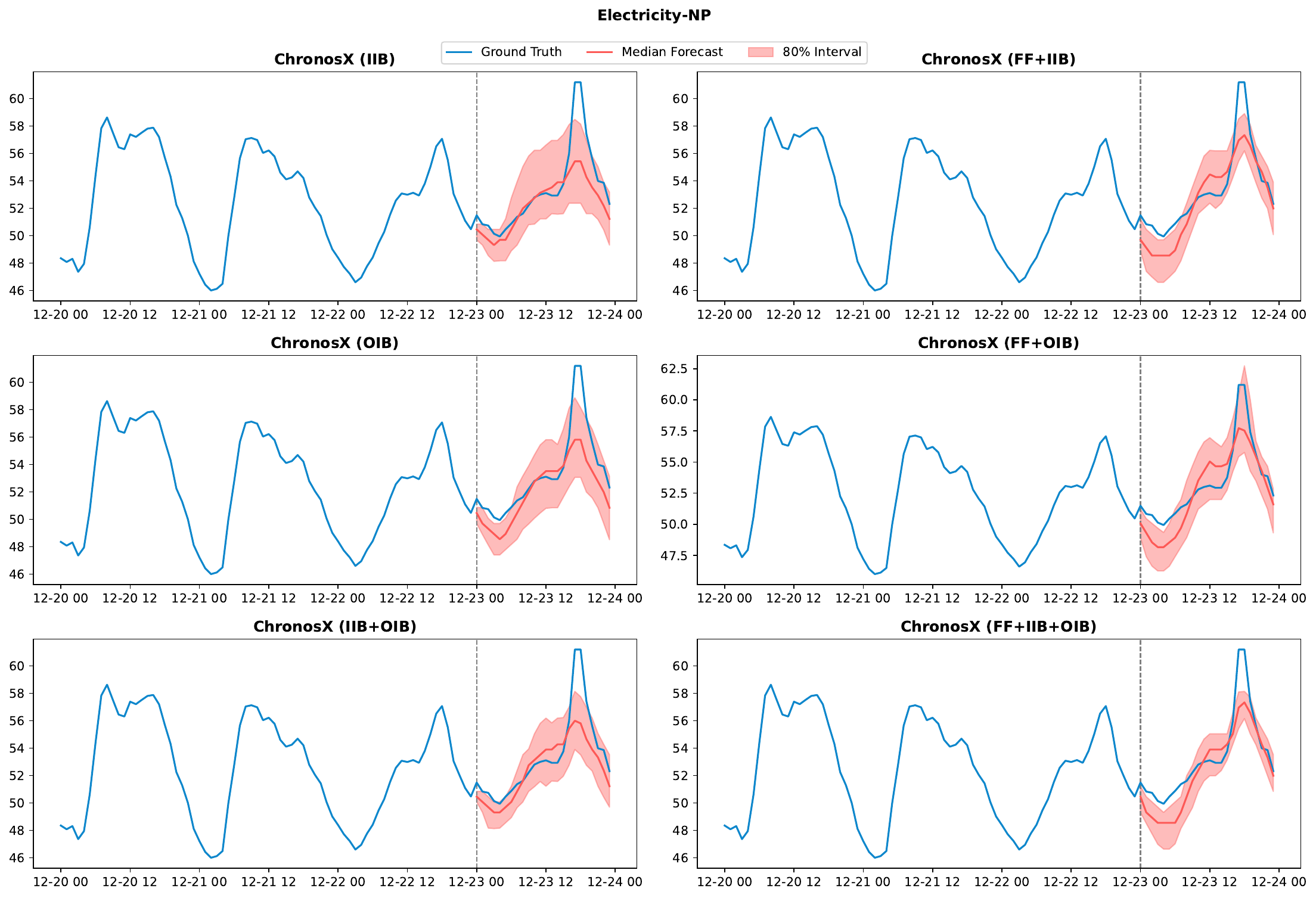}
    \caption{Forecasts on Electricity-NP Dataset}
\end{figure}
\begin{figure}[h]
    \centering
    \includegraphics[trim={0 0 0 1cm},clip, width=1\linewidth]{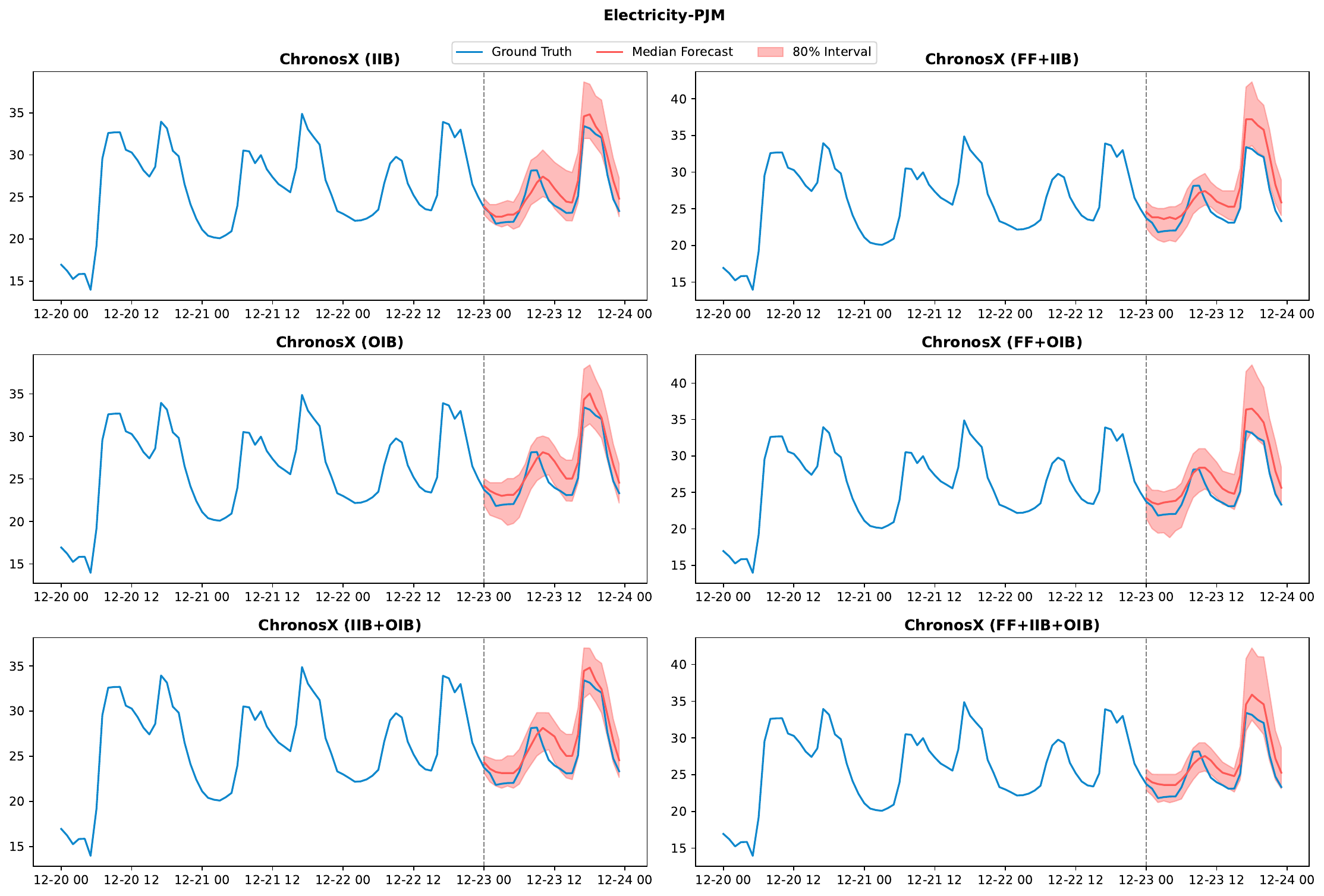}
    \caption{Forecasts on Electricity-PJM Dataset}
\end{figure}
\begin{figure}[h]
    \centering
    \includegraphics[trim={0 0 0 1cm},clip, width=1\linewidth]{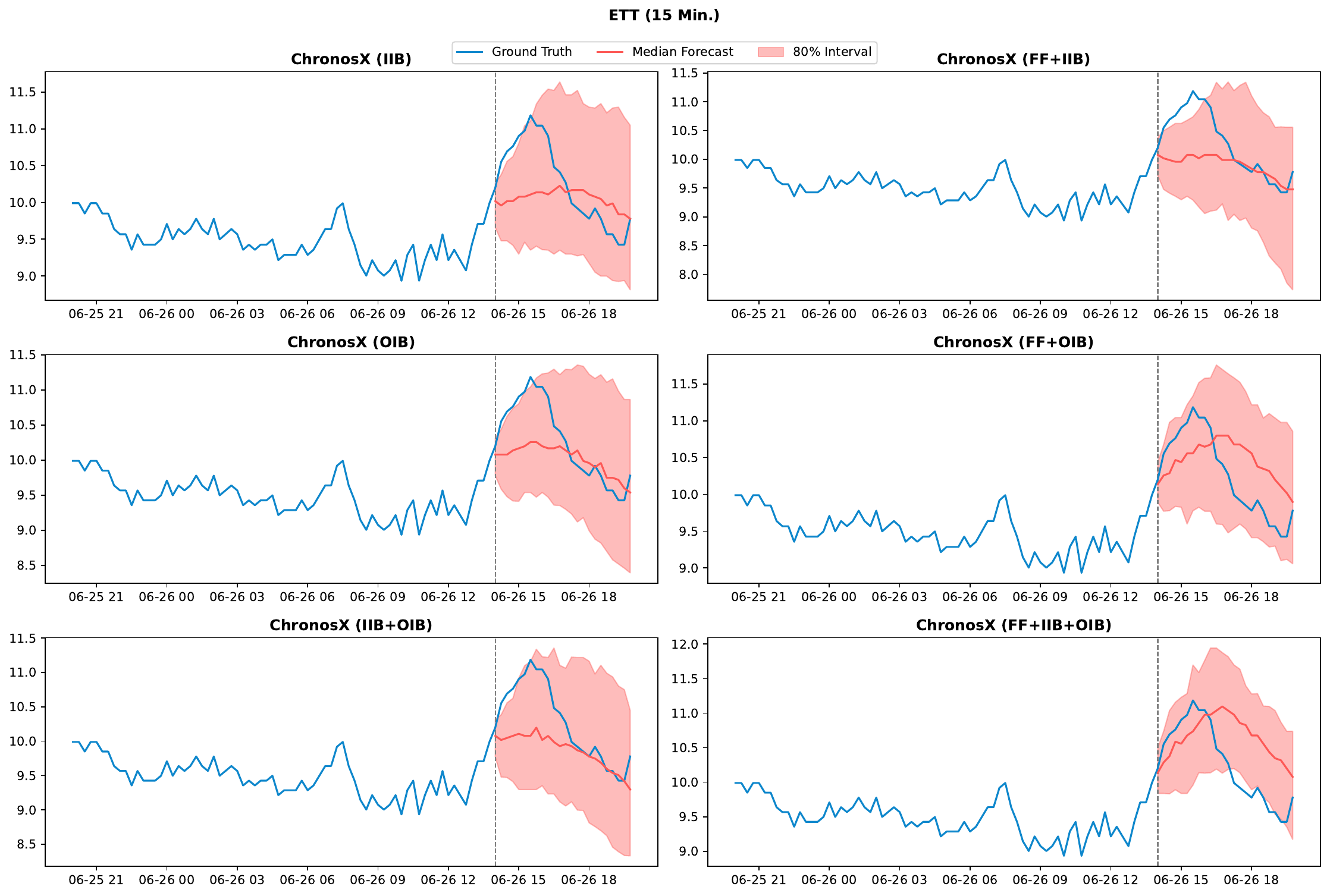}
    \caption{Forecasts on ETT (15 Min.) Dataset}
\end{figure}
\begin{figure}[h]
    \centering
    \includegraphics[trim={0 0 0 1cm},clip, width=1\linewidth]{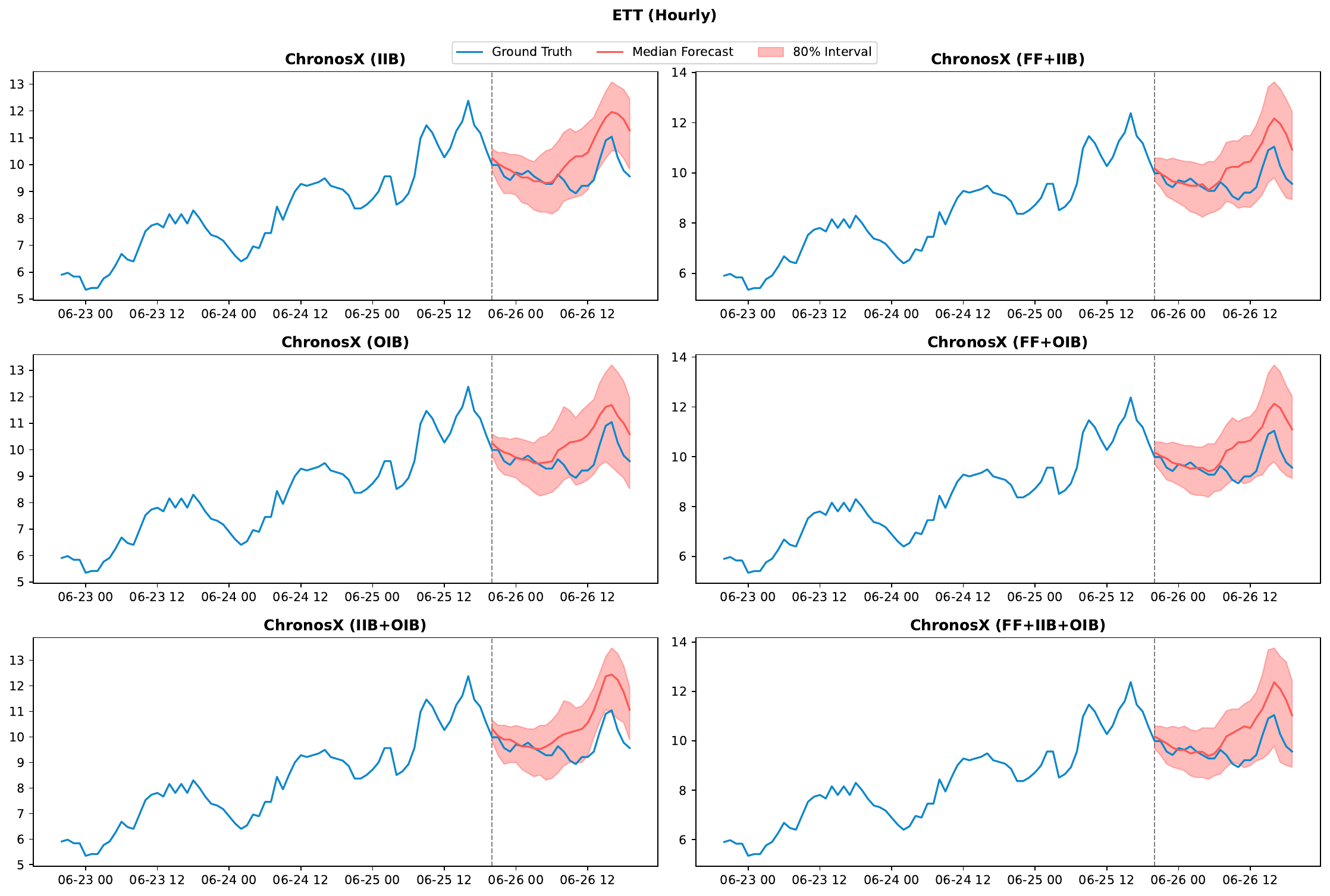}
    \caption{Forecasts on ETT (Hourly) Dataset}
\end{figure}
\begin{figure}[h]
    \centering
    \includegraphics[trim={0 0 0 1cm},clip, width=1\linewidth]{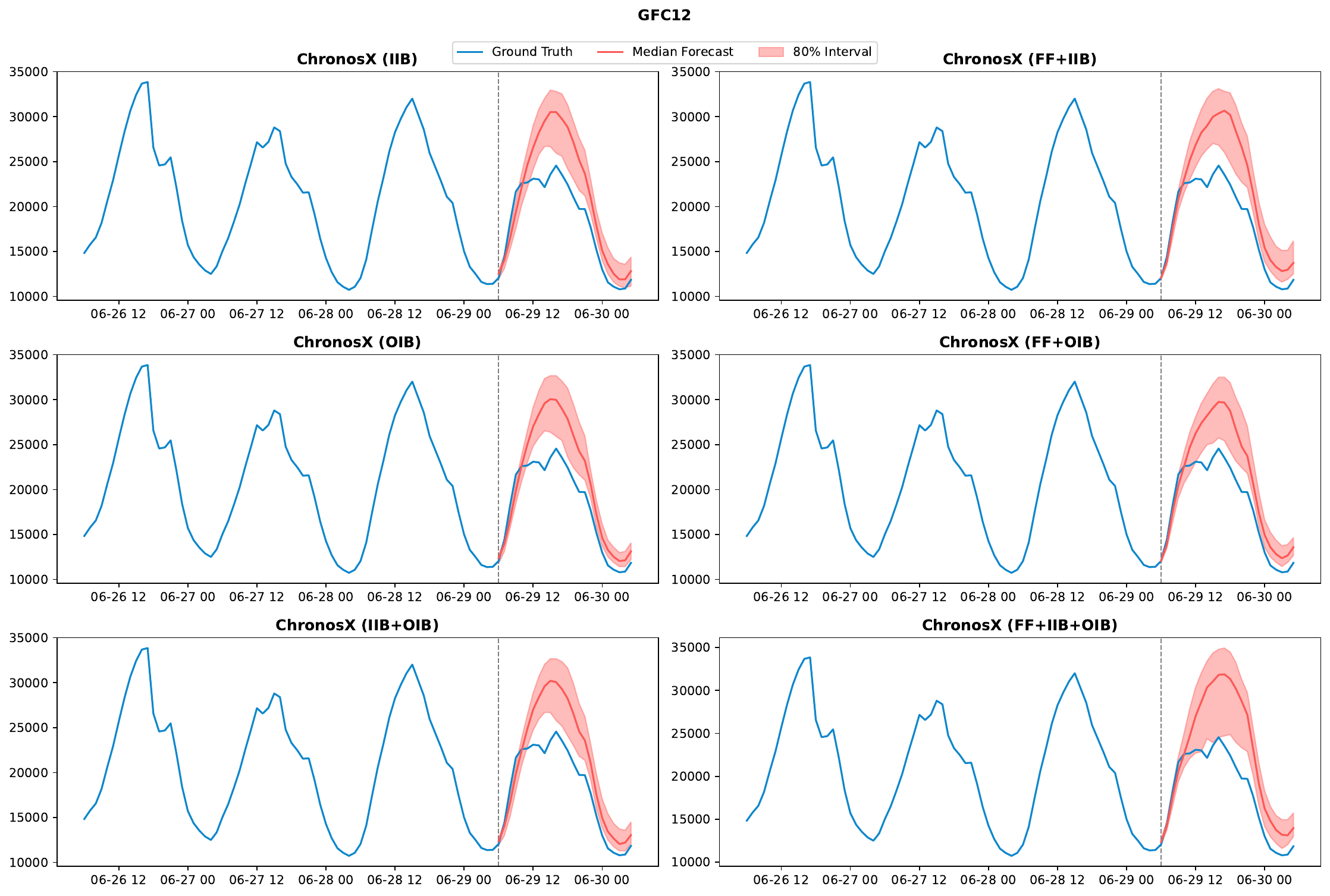}
    \caption{Forecasts on GEF12 Dataset}
\end{figure}
\begin{figure}[h]
    \centering
    \includegraphics[trim={0 0 0 1cm},clip, width=1\linewidth]{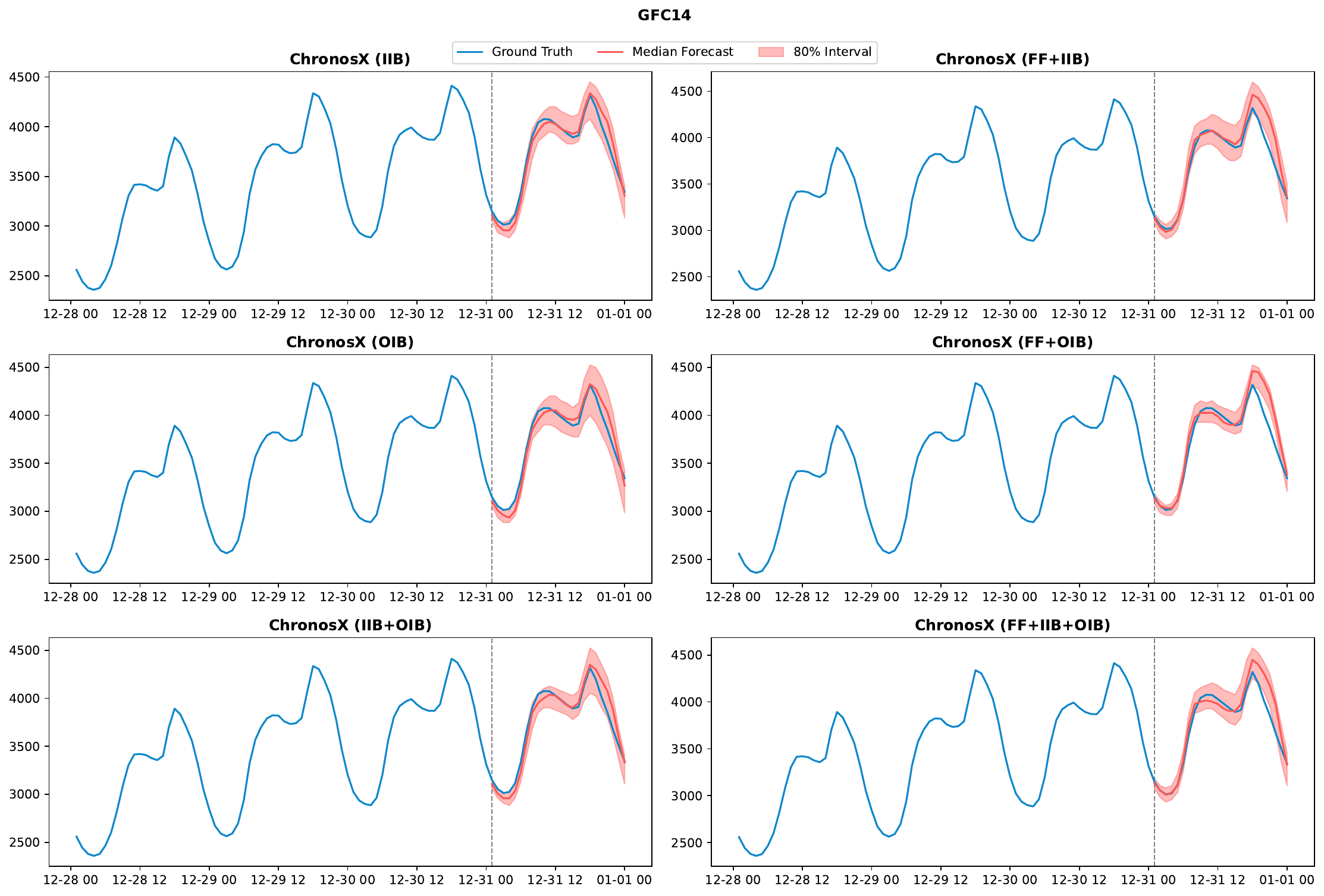}
    \caption{Forecasts on GEF14 Dataset}
\end{figure}
\begin{figure}[h]
    \centering
    \includegraphics[trim={0 0 0 1cm},clip, width=1\linewidth]{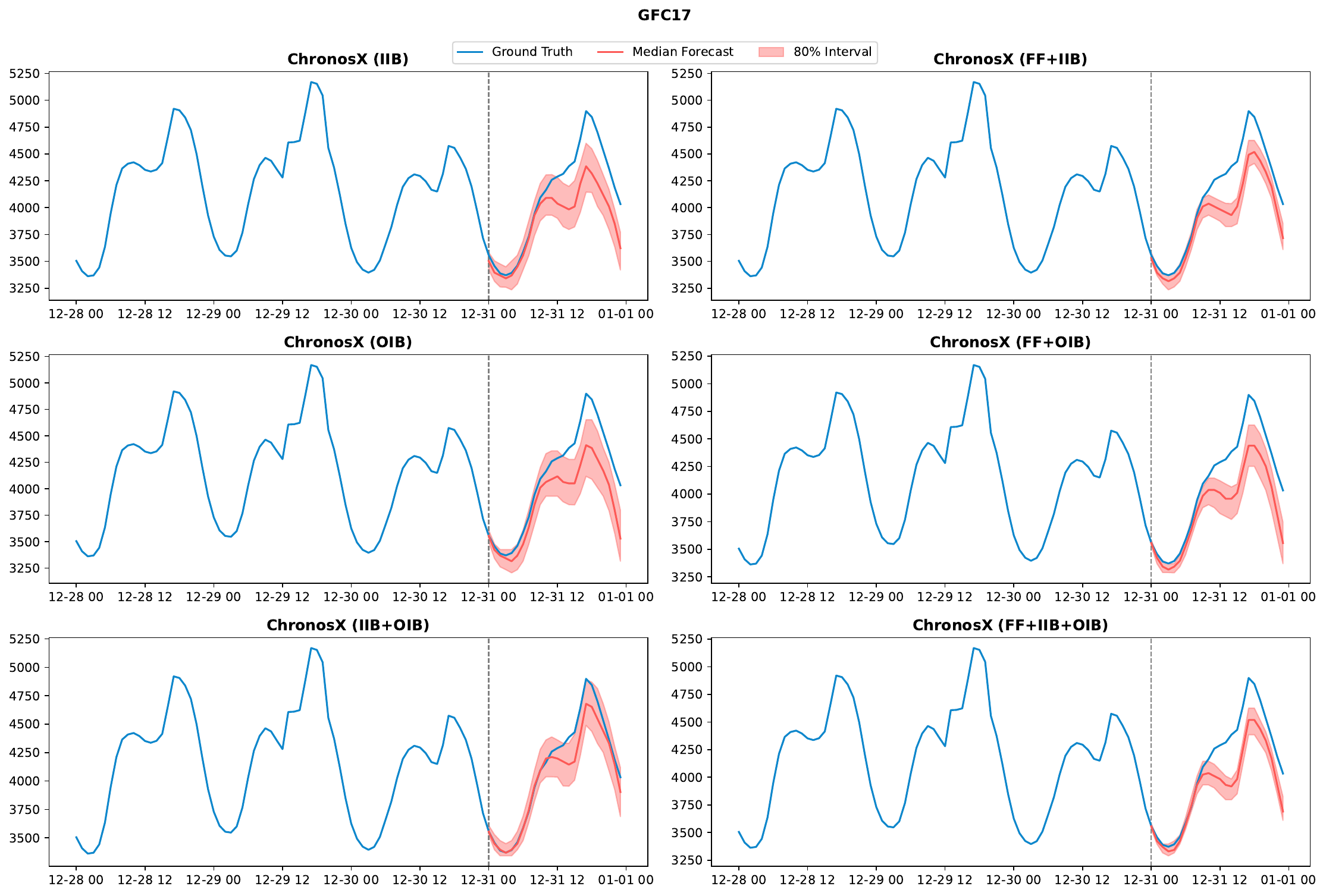}
    \caption{Forecasts on GEF17 Dataset}
\end{figure}
\begin{figure}[h]
    \centering
    \includegraphics[trim={0 0 0 1cm},clip, width=1\linewidth]{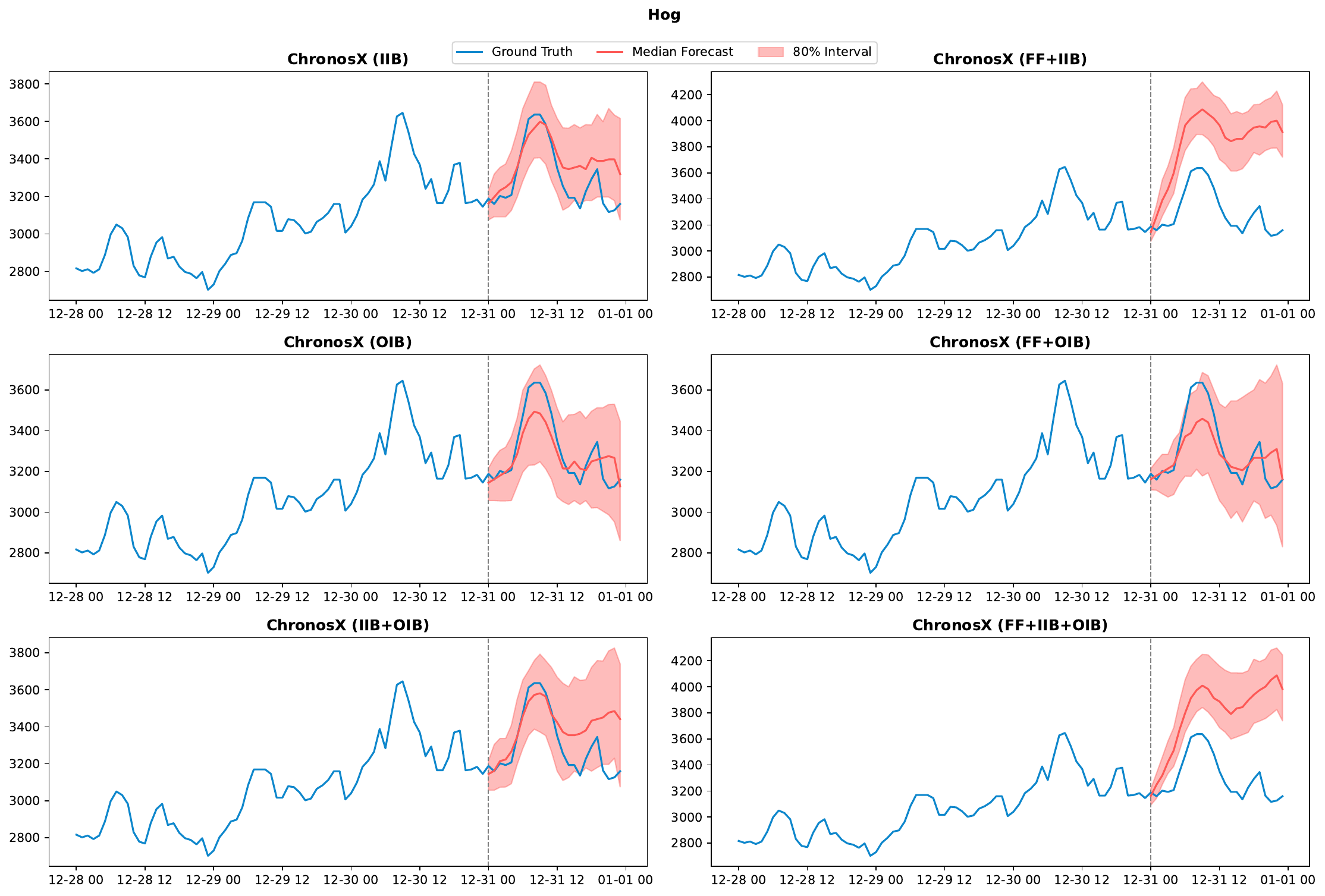}
    \caption{Forecasts on Hog Dataset}
\end{figure}
\begin{figure}[h]
    \centering
    \includegraphics[trim={0 0 0 1cm},clip, width=1\linewidth]{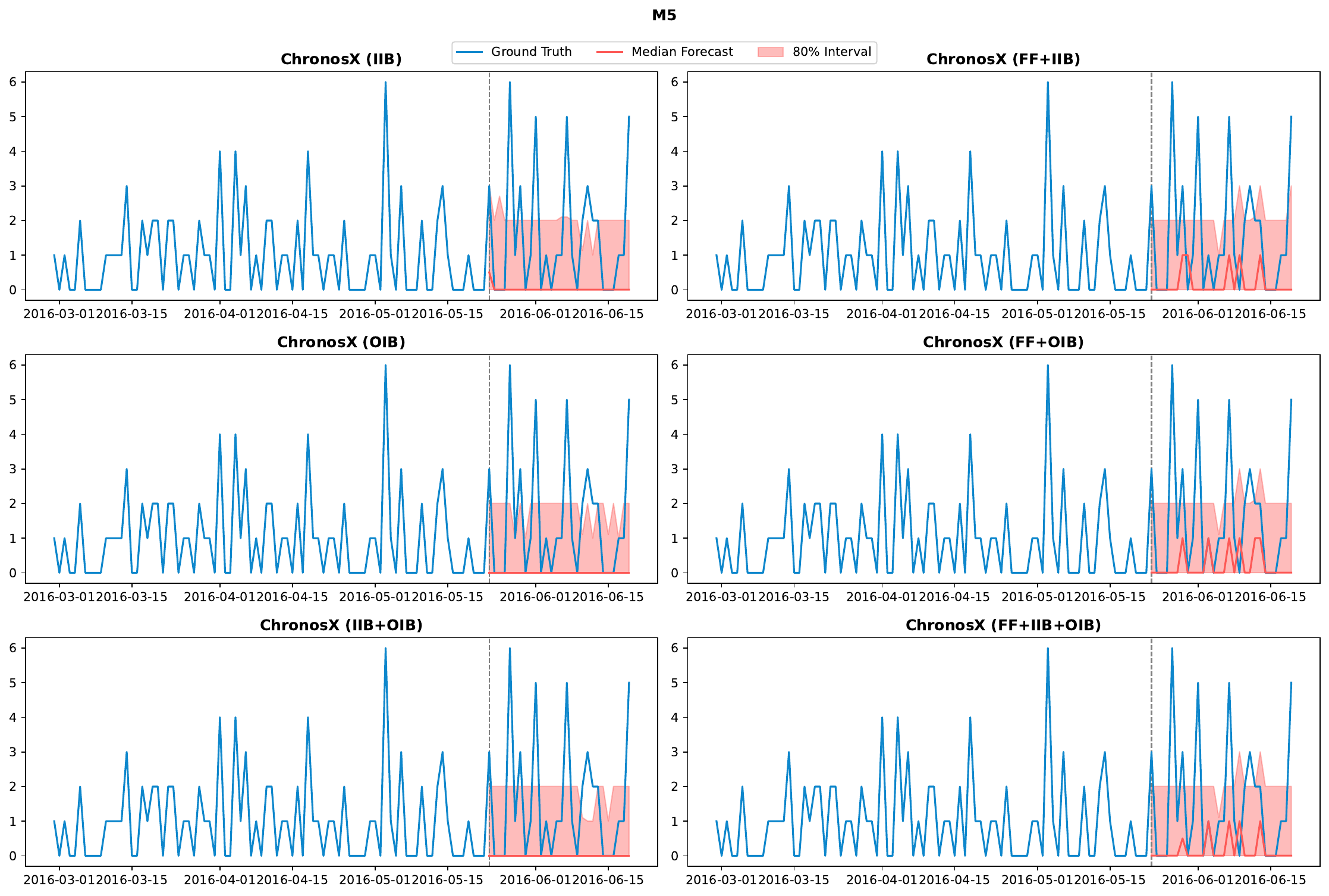}
    \caption{Forecasts on M5 Dataset}
\end{figure}
\begin{figure}[h]
    \centering
    \includegraphics[trim={0 0 0 1cm},clip, width=1\linewidth]{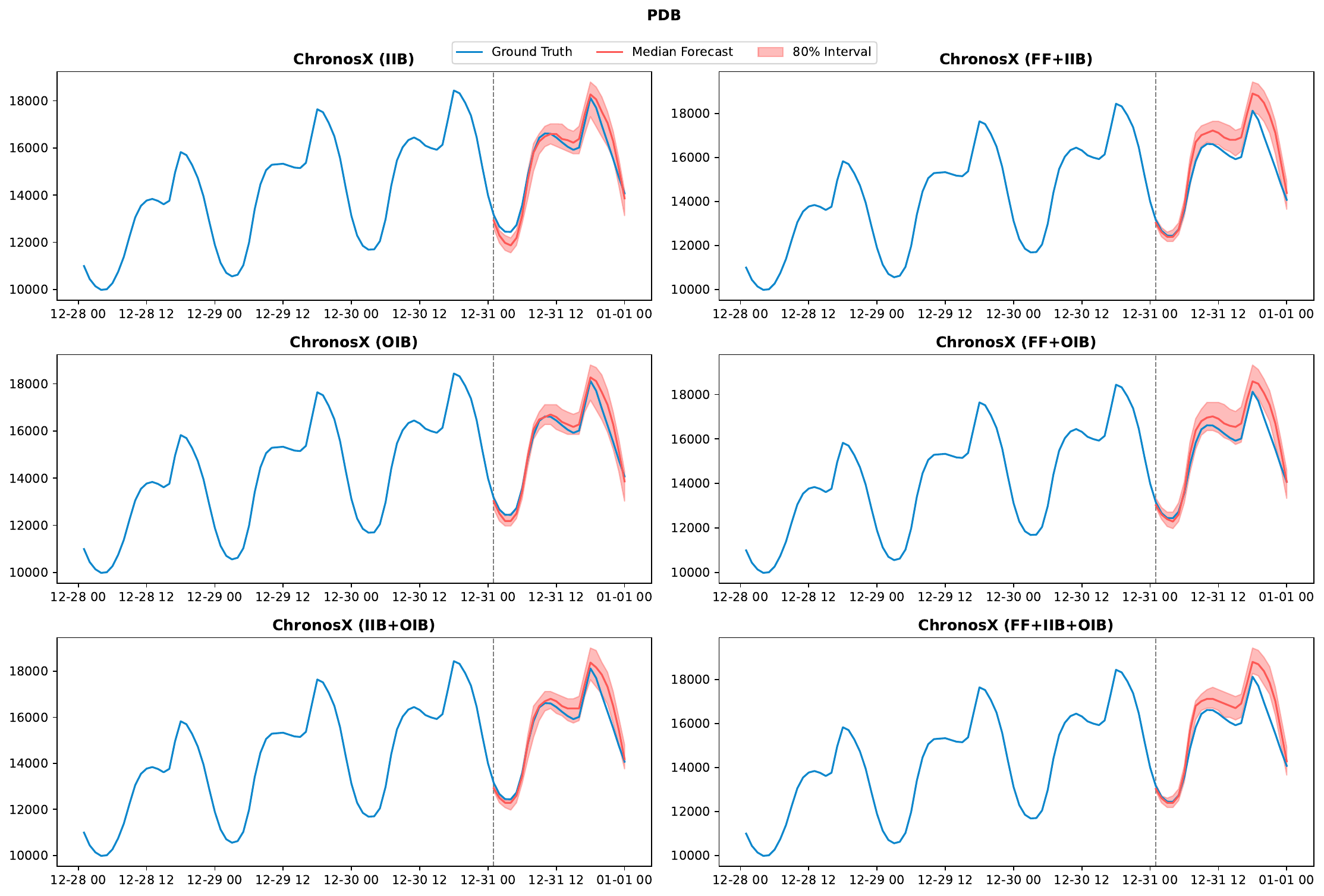}
    \caption{Forecasts on PDB Dataset}
\end{figure}
\begin{figure}[h]
    \centering
    \includegraphics[trim={0 0 0 1cm},clip, width=1\linewidth]{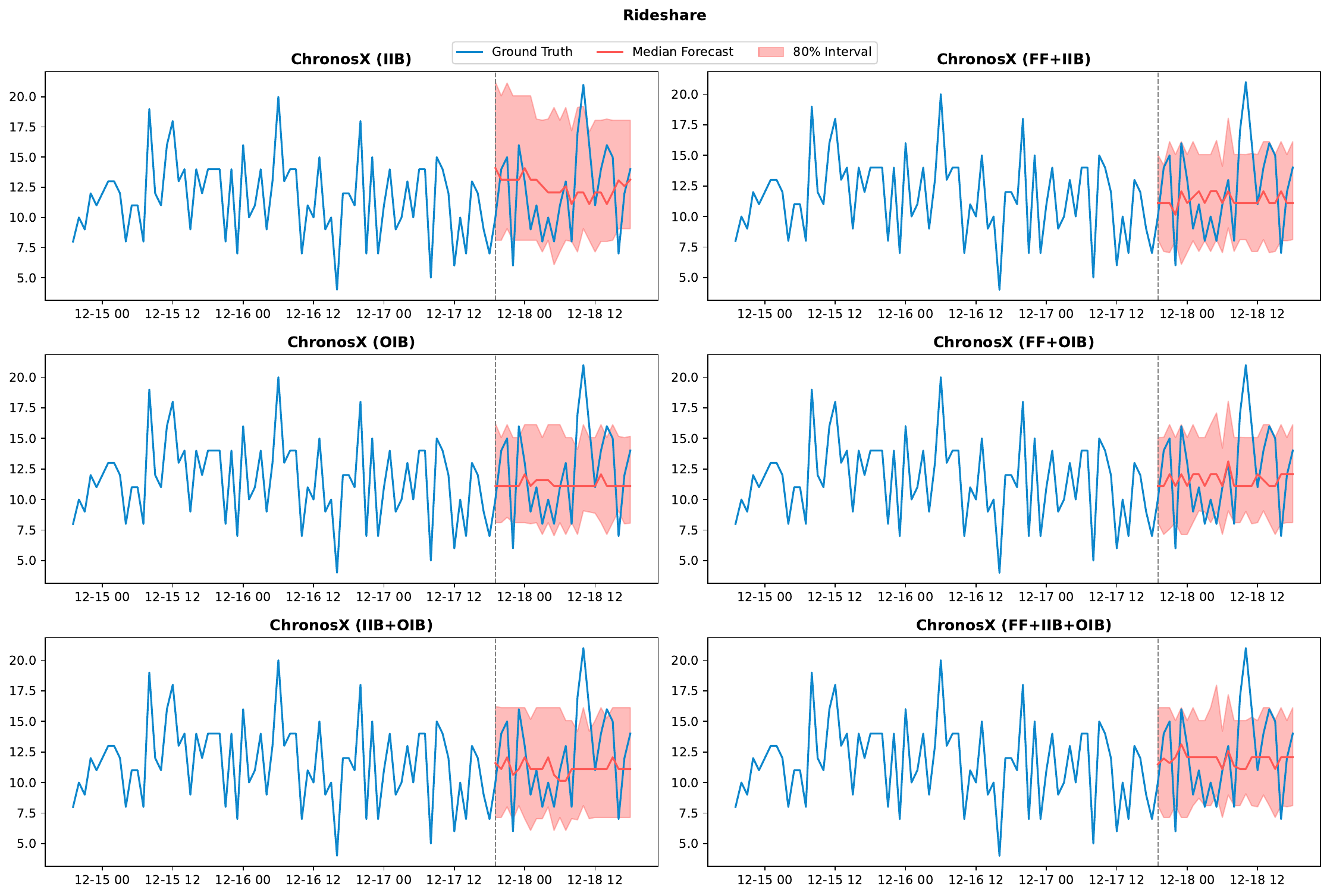}
    \caption{Forecasts on Rideshare Dataset}
\end{figure}
\clearpage\newpage
\clearpage\newpage

\end{document}